%% file: manuscript.tex
\documentclass{article}
\usepackage{custom_twocolumn}

\usepackage[T1]{fontenc}
\usepackage[utf8]{inputenc}
\usepackage[ttscale=.875]{libertine}
\usepackage[english]{babel}
\usepackage{microtype}
\usepackage{csquotes}
\usepackage[dvipsnames]{xcolor}
\usepackage[font=footnotesize]{caption}
\usepackage{hyperref}
\usepackage{algorithm}
\usepackage{algorithmicx}
\usepackage{algpseudocode}
\usepackage{authblk}
\usepackage{booktabs}
\usepackage{graphicx}
\usepackage[export]{adjustbox}
\usepackage{tikz,pgfplots}
\usetikzlibrary{spy,positioning}
\usepgfplotslibrary{groupplots}
\pgfplotsset{compat=1.15}
\usepackage{fancyhdr}
\usepackage[
    backend=biber,
    style=authoryear-comp,
    minbibnames=99,
    maxbibnames=99,
    mincitenames=1,
    maxcitenames=2,
    dashed=false,
    firstinits=true,
    uniquelist=minyear,
    labelyear=true,
]{biblatex}
\DeclareNameAlias{sortname}{last-first}
\hypersetup{
    colorlinks=true,
    linkcolor=NavyBlue,
    filecolor=NavyBlue,
    urlcolor=NavyBlue,
    citecolor=NavyBlue,
}

\definecolor{color0}{rgb}{0.12156862745098,0.466666666666667,0.705882352941177}
\definecolor{color1}{rgb}{1,0.498039215686275,0.0549019607843137}
\definecolor{color2}{rgb}{0.172549019607843,0.627450980392157,0.172549019607843}
\definecolor{color3}{rgb}{0.83921568627451,0.152941176470588,0.156862745098039}
\definecolor{color4}{rgb}{0.580392156862745,0.403921568627451,0.741176470588235}
\definecolor{color5}{rgb}{0.549019607843137,0.337254901960784,0.294117647058824}
\definecolor{color6}{rgb}{0.890196078431372,0.466666666666667,0.76078431372549}
\definecolor{color7}{rgb}{0.737254901960784,0.741176470588235,0.133333333333333}
\definecolor{color8}{rgb}{0.0901960784313725,0.745098039215686,0.811764705882353}
\definecolor{color9}{rgb}{0.498,0.498,0.498}
\definecolor{color10}{rgb}{0.388235294117647,0.47450980392156,0.22352941176470}

\definecolor{tubred}{rgb}{0.6,0.0,0.0}
\definecolor{bmsblue}{rgb}{0.035,0.313,0.62}
\definecolor{darkgreen}{rgb}{0.1, 0.5, 0.1}

\pgfdeclarelayer{fg}
\pgfsetlayers{main,fg}

\newlength\figureheight
\newlength\figurewidth
\input{macros}

\title{Interpretable Neural Networks with Frank-Wolfe: \\ Sparse Relevance Maps and Relevance Orderings}

\author[1]{Jan Macdonald}
\author[2]{Mathieu Besan\c{c}on}
\author[1,2]{Sebastian Pokutta}

\affil[1]{Institut f{\"u}r Mathematik, Technische Universit{\"a}t Berlin}
\affil[2]{Department for AI in Society, Science, and Technology, Zuse Institute Berlin}
\affil[ ]{\texttt{macdonald@math.tu-berlin.de, \{besancon,pokutta\}@zib.de}}

\date{}

\begin{document}

\maketitle

\begin{abstract}
We study the effects of constrained optimization formulations and Frank-Wolfe algorithms for obtaining interpretable neural network predictions.
Reformulating the \emph{Rate-Distortion Explanations (RDE)} method for relevance attribution as a constrained optimization problem provides precise control over the sparsity of relevance maps.
This enables a novel multi-rate as well as a relevance-ordering variant of RDE that both empirically outperform standard RDE and other baseline methods in a well-established comparison test.
We showcase several deterministic and stochastic variants of the Frank-Wolfe algorithm and their effectiveness for RDE.
\end{abstract}

\section{Introduction}
Deep learning methods achieve outstanding results for tasks across various fields, ranging from image analysis \autocite{ksh12,sze+13}, to natural language processing \autocite{cho2014translation,vas+17}, to medical diagnosis \autocite{sws17,mcb+18}. However, they are mostly considered as black-box models. The reasoning of parameter-rich and highly nonlinear neural networks remains generally inaccessible. This is particularly undesirable in sensitive applications, such as medical diagnosis or autonomous driving.

\begin{figure}
    \centering\scriptsize
    \input{./img/classifier-relevance}
    \caption{Relevance attribution methods aim at rendering black-box classifiers more interpretable by providing heatmaps of the input features that contribute most to an individual prediction.}
    \label{fig:relevance}
\end{figure}
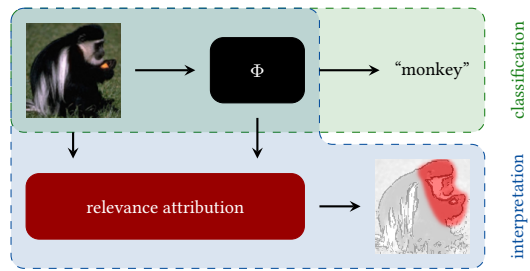

The ability to render these models less opaque by providing human-interpretable explanations of their predictions is essential for a reliable use of neural networks. An important first step is the identification of the most relevant input features for a prediction, as illustrated in Figure~\ref{fig:relevance}.

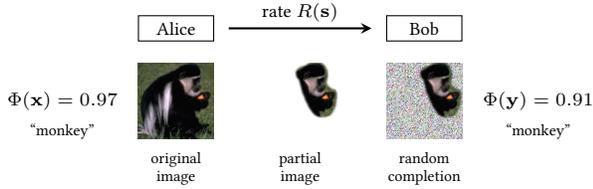
\begin{figure}
    \centering\scriptsize
    \input{./img/rate-distortion}
    \caption{Illustration of the rate-distortion viewpoint of relevance attribution. In a hypothetical scenario, Alice and Bob are given
access to a neural network classifier. Alice also has an image classified as a ``monkey'' and wants to convey this to Bob. He does not have the image and Alice is only allowed to send him a limited number of pixels. Bob will fill the remaining image with random values before classification. The best option is to transmit pixels that were most relevant for the prediction ``monkey'' of the original image. Figure adapted from \textcite{macdonald2020rde}.}
    \label{fig:rate_distortion}
\end{figure}

This has recently been formalized as an optimization problem in a rate-distortion framework \autocite{macdonald2019rde}, cf.\ Figure~\ref{fig:rate_distortion}.
We give a brief informal introduction here. A more detailed description is given in Section~\ref{sec:rde}.
The rate-distortion approach aims at balancing the expected change in the model prediction when modifying the non-relevant input features (distortion $D$) against the number of features that are considered relevant (rate $R$). 
We propose to restate the original Lagrangian formulation
\begin{equation}\label{eq:informal_lagrange}
    \min_{\bfs} \ D(\bfs) + \lambda \cdot R(\bfs),
\end{equation}
with regularization parameter $\lambda>0$ as a rate-constrained formulation
\begin{equation}\label{eq:informal_rc}
    \min_{\bfs\in\CC} \ D(\bfs),
\end{equation}
with a feasible region of the form $\CC=\{\bfs\,:\, R(\bfs)\leq k\}$ for some $k\in\N$.
This has the advantage of providing finer control over the trade-off between the rate and the distortion, by precisely prescribing the number $k$ of relevant features.

Variants of Gradient Descent (GD) are by far the most popular optimization methods when working with neural networks and have previously been used to solve the Lagrangian formulation \eqref{eq:informal_lagrange}.\footnote{More generally, proximal methods can be considered for non-smooth problems, e.g., including an $\ell_1$-norm sparsity penalty as the rate function, cf. Section~\ref{sec:rde}.} Incorporating the constraint $\bfs\in\CC$ of \eqref{eq:informal_rc} could also be achieved through Projected Gradient Descent (PGD). This requires a projection step
\begin{equation*}
    \bfs_{t+1} = \proj_\CC\left(\bfs_t - \eta_t \nabla D(\bfs_t)\right)
\end{equation*}
with step size $\eta_t > 0$ for each update in order to maintain feasible iterates.

Depending on the feasible region $\CC$ such projections can be costly. An alternative projection-free first-order method is the Frank-Wolfe (FW) algorithm that relies on a (often computationally cheaper) Linear Minimization Oracle (LMO)
\begin{equation}\label{eq:lmo}
    \bfv_t = \argmin_{\bfv \in \CC} \left\langle \nabla D(\bfs_t), \bfv \right\rangle,
\end{equation}
and then moves in direction $\bfv_t$ via the update
\begin{equation*}
    \bfs_{t+1} = \bfs_t + \eta_t(\bfv_t-\bfs_t),
\end{equation*}
with step size $\eta_t \in [0,1]$. Feasibility is maintained for convex regions $\CC$ since the new iterate is a convex combination of two feasible points $\bfs_t$ and $\bfv_t$.

In this work, we examine the effectiveness of using FW (and variants thereof) for solving rate-constrained problems of the form \eqref{eq:informal_rc} with the goal of obtaining interpretable neural networks.

\paragraph{Related Work}
A formal definition of the relevance of individual input features towards a prediction was proposed by \textcite{waeldchen2021jair} for classifiers on discrete domains and by \textcite{macdonald2019rde} for classifiers on continuous domains. It was shown that it is generally a hard computational problem to determine small sets of relevant features in the discrete setting \autocite{waeldchen2021jair} and that this hardness carries over to the continuous setting (more specifically to neural networks) \autocite{macdonald2020rde}. We consider a problem-relaxation and heuristic solution strategy, named \emph{Rate-Distortion Explanations} (RDE), introduced by \textcite{macdonald2019rde}. This results in the Lagrangian formulation \eqref{eq:informal_lagrange} and is described in more detail in Section~\ref{sec:rde}.

Similarly, \textcite{fong2017interpretable} consider different types of perturbations of the non-relevant features. This also results in an optimization problem in the spirit of \eqref{eq:informal_lagrange} that could be restated as a constrained problem. However, for brevity, we restrict our analysis of the efficacy of FW algorithms to the setting of RDE. Other explanation methods that are not based on optimization are discussed in Section~\ref{sec:xai}.

It is worth mentioning, that the objective function in \eqref{eq:informal_rc} will be non-convex. Although the Frank-Wolfe algorithm was originally intended for convex problems, it has also been studied and applied as a local method for problems with non-convex objectives over convex regions \autocite{lacostejulien2016convergence,reddi2016stochastic}. In particular, FW has recently received attention also in the context of neural networks, e.g., by \textcite{pokutta2020deepfw,berrada2021deepfw}.

\paragraph{Contributions}
This work extends and improves several aspects of the original RDE approach, which has been established as a promising and theoretically sound method for relevance attribution. We propose a rate-constrained formulation of RDE. Solving it with Frank-Wolfe algorithms yields sparse explanations and provides precise control over the size of the set of relevant features. This allows us to introduce a novel multi-rate variant of RDE.
Further, in addition to sparsity constraints FW enables us to explore other feasible regions. We propose a novel method aiming at ordering input features according to their relevance for the prediction instead of a partition into relevant and irrelevant features. To this end, we optimize over the Birkhoff polytope (the convex relaxation of the set of permutation matrices).
Our empirical study confirms the efficacy and flexibility of Frank-Wolfe algorithms for obtaining interpretable neural network predictions. We show for two exemplary image classification benchmark datasets that FW is able to determine relevant input features. Furthermore, the multi-rate as well as the relevance ordering variant of RDE both outperform standard RDE on a well-established comparison test.

\paragraph{Outline}
We review important concepts of interpretable neural networks in Section~\ref{sec:xai}. A formal definition of RDE is given in Section \ref{sec:rde}. The idea of ordering input features according to their relevance is presented in Section~\ref{sec:orderings}.
Section~\ref{sec:fw_algos} and Appendix~\ref{apx:technical} provide details regarding variants of the Frank-Wolfe algorithms, feasible regions $\CC$, and their linear minimization oracles. The description of our empirical study and its results are found in Section~\ref{sec:results} and Appendix~\ref{apx:results}.

% ----- ----- ----- ----- -----
\section{Interpretable Neural Networks}\label{sec:xai}
In an abstract sense, \emph{interpretability} refers to the ability to \emph{explain} the predictions made by a deep learning model (or more generally: a machine learning model) to humans in an \emph{understandable} way \autocite{doshivelez2017rigorous}. Here, ``explain'' and ``understandable'' are rather vague terms and can mean different things depending on the context and application. In an effort to address this task from any possible angle, there has been a surge of research related to explainable artificial intelligence (XAI) and various attempts to categorize interpretability methods, see e.g., the recent surveys by \textcite{fan2021survey,chakraborty2017survey,gilpin2018survey,zhang2021survey}.

\Textcite{zhang2021survey} propose a taxonomy to distinguish methods according to three characteristics: i) \emph{passive} methods give post hoc explanations for already trained neural networks while \emph{active} methods require changing the network architectures already during training, ii) the \emph{type of explanation} (in increasing order of explanatory power) ranges from extracting prototypical examples, to feature attribution, to extracting hidden semantics, and finally to extracting specific logical decision rules, iii) \emph{local} methods explain network predictions for individual data samples while \emph{global} methods aim at explaining networks as a whole.

We focus on passive and local feature attribution, which is among the most widely used forms of explanations. In this case, the goal is to assign scores to each input feature of a data sample indicating its \emph{relevance} for a prediction, cf.\ Figure~\ref{fig:relevance}. Feature attribution methods are particularly popular in the context of image classification, where the scores are visualized as so-called \emph{heatmaps} or \emph{relevance maps}. Besides optimization based approaches such as RDE, various other relevance attribution methods for neural networks have been proposed, e.g., gradient based methods such as \emph{Sensitivity Analysis} \autocite{simonyan2013deep} and \emph{SmoothGrad} \autocite{smilkov2017smoothgrad}, backward propagation methods such as \emph{Guided Backprop} \autocite{springenberg2015guidedbackprop}, \emph{Layerwise Relevance Propagation} (LRP) \autocite{bach2015lrp}, and \emph{Deep Taylor Decompositions} \autocite{montavon2018deeptaylor}, surrogate model methods such as \emph{Local Interpretable Model-Agnostic Explanations} (LIME) \autocite{marcotr2016lime}, and game theoretic approaches such as \emph{Shapley Additive Explanations} (SHAP) \autocite{lundberg2017shap}. We use these methods as comparison baselines in our numerical evaluations.

\subsection{Sparse Relevance Maps}\label{sec:rde}
Intuitively, the relevant input features of a neural network classifier\footnote{In multi-class problems the classifier would usually be a map $\widetilde{\Phi}\colon\R^n\to \R^\text{\#classes}$ instead of $\Phi\colon\R^n\to\R$. However, as common for local explanations, we only consider the restriction of $\widetilde{\Phi}$ to the component corresponding to the class predicted for the data sample $\bfx$.} $\Phi\colon\R^n\to\R$ for an input sample $\bfx\in\R^n$ are determined by the following desiderata: changing the non-relevant features of $\bfx$ while keeping its relevant features fixed should not result in a large change in the classifier prediction $\Phi(\bfx)$. On the other hand, we do not want to mark all features as relevant but only those that necessarily have to be kept fixed to leave the prediction unchanged.

Encoding the relevance scores in a vector $\bfs\in[0,1]^n$, where $0$ means least relevant and $1$ means most relevant, this can be formulated as a rate-distortion tradeoff with a distortion function
\begin{equation}\label{eq:s_distortion}
    D(\bfs) = \E_{\bfn\sim\CV}[\left(\Phi(\bfx) - \Phi(\bfs\odot\bfx+(\bfone-\bfs)\odot\bfn)\right)^2],
\end{equation}
measuring the expected quadratic change in the classifier prediction when randomizing parts of $\bfx$ determined by $\bfs$, and a rate function $R(\bfs)=\|\bfs\|_1$, measuring the size\footnote{The $\ell_1$-norm can be seen as a convex surrogate for the cardinality of the support of $\bfs$, i.e., the size of the \emph{relevant set} of input components (those with positive relevance scores).} of $\bfs$ \autocite{macdonald2019rde,macdonald2020rde}.
Here, $\odot$ denotes the Hadamard product and $\CV$ is some chosen probability distribution on $\R^n$ that is used to modify the features of $\bfx$ that are not fixed by $\bfs$. Typical choices for $\CV$ include Gaussians \autocite{fong2017interpretable}, uniform distributions \autocite{samek2017pixelflip}, constant baseline signals \autocite{fong2017interpretable}, or local distributions around $\bfx$ \autocite{ribeiro2018anchors}.
In this work, we will consider Gaussian distributions, since more data-adaptive choices of $\CV$ can even be detrimental for uncovering the reasoning of a classifier \autocite{janzing20cond_vs_marg,macdonald2020rde}.

The goal is to find the smallest $\bfs$ (in terms of the rate) achieving a certain distortion or vice-versa the minimal distortion possible for a fixed size of $\bfs$. While the former results in a computationally challenging optimization problem with a non-convex constraint (distortion-constraint), the latter problem has a convex feasible region and it is---as we demonstrate in this work---computationally feasible to compute local minima.

The original Rate-Distortion Explanation (RDE) approach addresses the tradeoff between rate and distortion as a box-constrained optimization problem via a Lagrangian formulation
\begin{alignat*}{2}
     &\min_{\bfs}  \ &&D(\bfs) + \lambda \|\bfs\|_1 \tag{L-RDE}\label{eq:l_rde}\\
     &\subjectto \ &&\bfs\in [0,1]^n,
\intertext{
with a regularization parameter $\lambda>0$. Instead, we propose to use the rate-constrained formulation
}
     &\min_{\bfs}  \ &&D(\bfs) \tag{RC-RDE}\label{eq:rc_rde}\\
     &\subjectto \ &&\|\bfs\|_1 \leq k, \\
     &&&\bfs\in [0,1]^n,
\end{alignat*}
with a maximal rate $k\in[n-1]=\{1,\dots,n-1\}$.\footnote{The case $k=n$ always has the trivial solution $\bfs=\bfone$.} This allows for a precise control of the size of $\bfs$ or in other words for the \emph{sparsity} of the relevance map. Controlling the sparsity makes it easier to exactly pinpoint the most relevant input features of $\bfx$, which would otherwise require more manual tuning of the regularization parameter $\lambda$.

Computing the expectation value in \eqref{eq:s_distortion} exactly is generally not possible.
As in the original RDE approach, we exploit the layered structure of neural networks and rely on an approximation scheme based on \emph{assumed density filtering} \autocite{minka2001adf} for the evaluation of the non-convex function $D(\bfs)$. The Gaussian distribution $\CV$ is estimated from the training data and the gradient $\nabla D(\bfs)$ is obtained through automatic differentiation. Using this first-order information, we can then solve \eqref{eq:rc_rde} with (variants of) the Frank-Wolfe algorithm \autocite{frank1956fw} or PGD, see Section~\ref{sec:fw_algos}.

\subsection{Relevance Orderings}\label{sec:orderings}
The exact values of a relevance map are often not individually meaningful (in the sense that knowing the value $s_i$ for the $i$-th variable alone is not helpful). It is rather the ordering (by relevance) of the variables induced by $\bfs$ that is of interest, i.e., the relations $s_i < s_j$ or $s_j < s_i$ between different variables. In fact, an established evaluation method for the comparison of different relevance maps, the \emph{pixel-flipping} test \autocite{samek2017pixelflip}, and a variant thereof proposed by \textcite{macdonald2019rde} is based on this idea. It proceeds as follows: i) order variables by their relevance ii) keep increasingly large parts of the input $\bfx$ fixed and randomize the remaining variables iii) observe the change in the classifier prediction (distortion) during this process. A good ordering will lead to a quick decrease of the distortion as truly important input features are fixed first. This precisely corresponds to the rate-distortion tradeoff described above. In fact, \eqref{eq:l_rde} and \eqref{eq:rc_rde} aim at optimizing the resulting rate-distortion curve for a single rate determined implicitly by $\lambda$ or explicitly by $k$ respectively, cf.\ Figure~\ref{fig:mnist_curves_rc}.

Instead of first obtaining relevance scores and afterwards retrieving a relevance ordering from them, one could find an optimal ordering directly by solving
\begin{alignat*}{2}
    &\min_{\bfPi}  \ &&\frac{1}{n-1}\sum_{k\in[n-1]}D(\bfPi\bfp_k) \\
    &\subjectto \ &&\bfPi\in S_n,
\end{alignat*}
where $S_n$ denotes the set of $(n\times n)$ permutation matrices and $\bfp_k = \sum_{j=1}^{k} \bfe_j$ is the vector of $k$-ones and $(n-k)$-zeros. Hence, $D(\bfPi\bfp_k)$ corresponds to the distortion of fixing the $k$ most relevant features (according to $\bfPi$) and the objective aims at minimizing the average distortion across all rates $k\in[n-1]$ simultaneously. We relax this combinatorial problem ($S_n$ is discrete) to
\begin{alignat}{2}
    &\min_{\bfPi}  \ &&\frac{1}{n-1}\sum_{k\in [n-1]} D(\bfPi\bfp_{k}) \tag{Ord-RDE}\label{eq:ord_rde}\\
    &\subjectto \ &&\bfPi\in B_n, \nonumber
\end{alignat}
where $S_n$ is replaced with its convex hull, i.e., the Birkhoff polytope $B_n = \conv(S_n)$ of doubly stochastic $(n\times n)$-matrices.
This can be solved with a (batched) stochastic version of the Frank-Wolfe algorithm \autocite{hazan2016sfw} or as before with non-stochastic versions of Frank-Wolfe if $n$ is small enough so that evaluating the complete sum in \eqref{eq:ord_rde} is not too expensive. There is no exact projection method specific to the Birkhoff polytope, see Appendix~\ref{apx:technical}, hence we do not consider PGD for solving \eqref{eq:ord_rde}.

A solution to \eqref{eq:ord_rde} can be seen as a greedy-approximation to \eqref{eq:rc_rde} across all rates $k\in[n-1]$ simultaneously in the following sense:
a solution $\bfPi^\text{opt}$ is a convex combination of permutation matrices (the vertices of $B_n$). It can be used to obtain mappings for specific rates via $\bfPi^\text{opt}\bfp_k$, which we interpret as a convex combination of the respective $k$ most relevant components according to each permutation contributing to a convex decomposition of $\bfPi^\text{opt}$. From now on we refer to $\bfPi^\text{opt}\bfp_k$ with $k\in[n-1]$ as the single-rate mappings associated to $\bfPi^\text{opt}$.

One should note that, in contrast to \eqref{eq:rc_rde}, a straightforward approach to solving \eqref{eq:ord_rde} is not feasible for large-scale problems: optimizing over matrices in $\R^{n\times n}$ instead of vectors in $\R^n$ results in increased computational costs, both in terms of memory requirements and runtime (see also descriptions of the LMOs in Appendix~\ref{apx:technical}). However, this might in part be remedied by a clever and more memory-efficient representation of iterates.\footnote{The $t$-th FW iterate $\bfPi_t$ is a convex combination of an active set of at most $t$ permutations (corresponding to vertices of $B_n$). Storing these together with the convex weights allows to effectively recover the iterate but reduces the memory requirement from $\CO(n^2)$ to $\CO(tn)$ (assuming $t<n$). For algorithms keeping track of an active set anyways, such as Away-Step Frank-Wolfe, see Appendix~\ref{apx:technical}, there is no computational overhead. Similarly, a sparse matrix representation reduces the memory requirement to $\CO(\#\text{non-zero components})$. However, the bottleneck of our current implementation is the LMO evaluation. The gradients ($\bfG_t$ in Algorithm~\ref{alg:sfw}) will still be dense matrices. In fact, they are sums of rank-1 matrices. We leave a study of the practical benefits of exploiting this structure for the LMO evaluation to future research.}

Another possibility to overcome this limitation is to emulate a similar multi-rate strategy that relies solely on the rate-constrained formulation: we can separately solve \eqref{eq:rc_rde} for all $k\in\CK$ for some $\CK\subseteq [n-1]$ and combine these solutions, e.g.\ by averaging, to obtain relevance scores
\begin{equation}\tag{MR-RDE}\label{eq:mr_rde}
\bfs = \frac{1}{|\CK|}\sum_{k\in\CK} \left\{
\begin{alignedat}{2}
    & \argmin  \ && D(\bfs) \\
    &\subjectto \ &&\|\bfs\|_1 \leq k, \\
    &&&\bfs\in [0,1]^n,
\end{alignedat}
\right\}
\end{equation}
that take multiple rates into account. Consequently, this effectively yields an induced ordering of the variables according to their relevance across all rates (if a variable is contained in the solutions for many of the rates then it contributes more to the averaging of relevance maps over $\CK$ and should be considered more relevant than a variable that is only contained in few solutions, hence it should come earlier in the ordering).

In summary, we can interpret the different RDE variants in terms of their objective regarding the pixel-flipping evaluation: solving a single \eqref{eq:rc_rde} problem aims at optimizing the rate-distortion curve, i.e., achieving low distortion, at a single rate. This might lead to suboptimal results at other rates, see Figure~\ref{fig:mnist_curves_rc}.
In contrast, \eqref{eq:ord_rde} aims at directly optimizing the relevance ordering and thus optimizes the distortion for all rates simultaneously on average.
Finally, \eqref{eq:mr_rde} combines single-rate solutions spread across the range of considered rates and thus approximately also aims at optimizing the distortion everywhere by minimization at well-chosen attachment points.
This is only possible because the Frank-Wolfe methods allow us to efficiently solve the rate-constrained problem with precise control over the sparsity of the computed relevance maps.

% ----- ----- ----- ----- -----
\section{Frank-Wolfe Algorithms}\label{sec:fw_algos}

The Frank-Wolfe algorithm \autocite{frank1956fw} or conditional gradient method \autocite{levitin1966cg} is a projection-free first-order algorithm for constrained optimization over a convex compact feasible set.
Since its first appearance, several algorithmic variations have been developed that enhance the performance of the original algorithm, while maintaining many of its advantages.
In our experiments, we consider vanilla Frank-Wolfe (FW), Away-Step Frank-Wolfe (AFW), Lazified Conditional Gradients (LCG), Lazified Away-Step Frank-Wolfe (LAFW), Stochastic Frank-Wolfe (SFW). For comparison we also consider Projected Gradient Descent (PGD). We give a brief overview of FW and SFW here and defer a description of AFW, LCG, and LAFW as well as of the feasible regions $\CC$ and associated LMOs and projections to Appendix~\ref{apx:technical}. The interested reader is referred to a more detailed presentation of the algorithm variants and their implementations\footnote{We use the \texttt{FrankWolfe.jl} Julia package available at \url{https://github.com/ZIB-IOL/FrankWolfe.jl} for our experiments. See Appendix~\ref{apx:technical} for more details regarding our computational setup.} \autocite{besancon2021fwjl}.

%TODO ref to new repo/zenodo/doi

\paragraph{Vanilla Frank-Wolfe (FW)} In its basic version, see Algorithm~\ref{alg:fw}, a linear approximation to the objective function at the current iterate (obtained from first-order information) is minimized over the feasible region.
Such a linear minimization oracle (LMO) is often computationally less costly than a corresponding projection (which amounts to solving a quadratic problem).
The next FW iterate is obtained as a convex combination of the solution to the LMO and the current iterate. For convex regions $\CC$, this guarantees that the algorithm produces iterates that remain feasible throughout all iterations.
This basic variant has the lowest memory requirements among all deterministic variants since it only requires keeping track of the current iterate.
Hence, it is well suited for large-scale problems. However, other variants can achieve improvements in terms of convergence speed (iteration count and time for more specialized setups), see Appendix~\ref{apx:technical}.

\begin{algorithm}
\caption{Frank-Wolfe for \eqref{eq:rc_rde}}
\label{alg:fw}
\begin{algorithmic}[1]
    \Input initial guess $\bfs_0\in\CC=\{\bfs\in[0,1]^n\,:\,\|\bfs\|_1\leq k\}$, number of steps $T$, step sizes $\eta_t\in[0,1]$
    \Output a stationary point $\bfs^\text{opt}$ of \eqref{eq:rc_rde}
    \Statex
    \For{$t\gets 1$ \textbf{to} $T$}
        \State $\bfv_t \gets \argmin_{\bfv\in\CC} \left\langle \nabla D(\bfs_{t-1}), \bfv \right\rangle$
        \State $\bfs_t \gets \bfs_{t-1} + \eta_t(\bfv_t-\bfs_{t-1})$
    \EndFor
    \State \Return $\bfs_T$
\end{algorithmic}
\end{algorithm}

\paragraph{Stochastic Frank-Wolfe (SFW)} In some cases, evaluating the full objective function and its gradients is expensive, but cheaper unbiased estimators are available.
The typical example is that of objective functions that are sums of a large number of terms, such as in the \eqref{eq:ord_rde} formulation. An estimator is given by evaluating the sum only over a randomly chosen subset of terms.
The stochastic version of Frank-Wolfe \autocite{hazan2016sfw} developed in Algorithm~\ref{alg:sfw} uses a gradient estimate instead of the exact gradient in combination with a momentum term \autocite{mokhtari2020sfw} to build the linear approximation to the objective in each iteration. Then the LMO is evaluated and a step is taken exactly as in the vanilla FW algorithm. Different variants of SFW have recently been studied also in the non-convex setting \autocite{yurtsever2019sfw,shen2019sfw,hassani2020stochastic,negiar2020stochastic}.

\begin{algorithm}
\caption{Stochastic Frank-Wolfe for \eqref{eq:ord_rde}}
\label{alg:sfw}
\begin{algorithmic}[1]
    \Input initial guess $\bfPi_0\in B_n$, number of steps $T$, step sizes $\eta_t\in[0,1]$, batch sizes $b_t\in[n-1]$, momentum factors $\rho_t\in[0,1]$
    \Output a stationary point $\bfPi^\text{opt}$ of \eqref{eq:ord_rde}
    \Statex
    \State $\bfM_0 \gets \bfzero_{n\times n}$
    \For{$t\gets 1$ \textbf{to} $T$}
        \State sample $k_1,\dots,k_{b_t}$ i.i.d.\ uniformly from $[n-1]$
        \State $\bfG_t \gets \frac{1}{b_t}\sum_{j=1}^{b_t} \nabla D(\bfPi_{t-1}\bfp_{k_j}) \bfp_{k_j}^\top$
        \State $\bfM_t \gets \rho_t\bfM_{t-1} + (1-\rho_t)\bfG_t$
        \State $\bfV_t \gets \argmin_{\bfV\in B_n} \left\langle  \bfM_t, \bfV \right\rangle$
        \State $\bfPi_t \gets \bfPi_{t-1} + \eta_t(\bfV_t-\bfPi_{t-1})$
    \EndFor
    \State \Return $\bfPi_T$
\end{algorithmic}
\end{algorithm}

\paragraph{Parameter Choices} The basic step size rule
\begin{equation}\label{eq:step_basic}
    \eta_t = \frac{1}{\sqrt{t+1}}
\end{equation}
can be used for non-convex objectives \autocite{reddi2016stochastic,combettes2021fwml}.
An adaptive step-size choice similar to one proposed by \textcite{carderera2021steps} is
\begin{equation}\label{eq:step_monotone}
    \eta_t = \frac{2^{-r_t}}{\sqrt{t+1}}
\end{equation}
where $r_t\in\N$ is found by repeated increments starting from $r_{t-1}$ until primal progress is made.
This ensures monotonicity in the objective, which is not necessarily the case for the basic rule \eqref{eq:step_basic}.
In our experiments, we use the monotone rule \eqref{eq:step_monotone} for FW, AFW, LCG, and LAFW and a corresponding (rescaled) variant also for the PGD comparison. Enforcing monotonicity does not make sense in the stochastic setting and we use the basic rule \eqref{eq:step_basic} for SFW. We test multiple configurations of SFW with constant or increasing batch sizes and momentum factors as proposed by \textcite{hazan2016sfw} and \textcite{mokhtari2020sfw} respectively. The full results can be found in Appendix~\ref{apx:results}. In the next section, we only show the best performing configuration with constant batch size $b_t=40$ and no momentum, i.e., $\rho_t=0$.
In all cases, we terminate after a maximal number of $T=2000$ iterations or if the dual gap $\langle\bfs_t-\bfv_t,\nabla D(\bfs_t)\rangle$ (respectively $\langle\bfPi_t-\bfV_t,\bfM_t\rangle$ for SFW) drops below the prescribed threshold $\varepsilon=10^{-7}$.

% ----- ----- ----- ----- -----
\section{Computational Results}\label{sec:results}
We generate relevance mappings for neural network classifiers trained on two image classification benchmark tasks. The first consists of greyscale images of handwritten digits from the MNIST dataset \autocite{lecun1998mnist} and the second consists of color images from the STL-10 dataset \autocite{coates2011stl}. We use the relevance ordering-based comparison test, as described in Section~\ref{sec:orderings}, for a quantitative evaluation of the relevance maps in addition to a purely visual qualitative evaluation. 
We show results for \eqref{eq:rc_rde}, \eqref{eq:mr_rde}, and \eqref{eq:ord_rde} as well as \eqref{eq:l_rde} and several established relevance mapping methods as comparison baselines.\footnote{Our code is publicly available at \url{https://github.com/ZIB-IOL/fw-rde} \autocite{macdonald_jan_2021_5718781}. All \eqref{eq:l_rde} and comparison method results were provided by \textcite{macdonald2019rde}. Their code is publicly available at \url{https://github.com/jmaces/rde}. Unlike RDE, some comparison methods give relevance scores in the range $[-1,1]$ instead of $[0,1]$ (indicating contributions \emph{against} and \emph{towards} a prediction).}

\paragraph{MNIST Experiment} For direct comparability, we use the same convolutional neural network of \textcite{macdonald2019rde} (with three convolutional layers each followed by average-pooling and finally two fully-connected layers and softmax output) that was trained end-to-end up to test accuracy of 0.99. The relevance mappings are calculated for the pre-softmax score of the class with the highest activation.

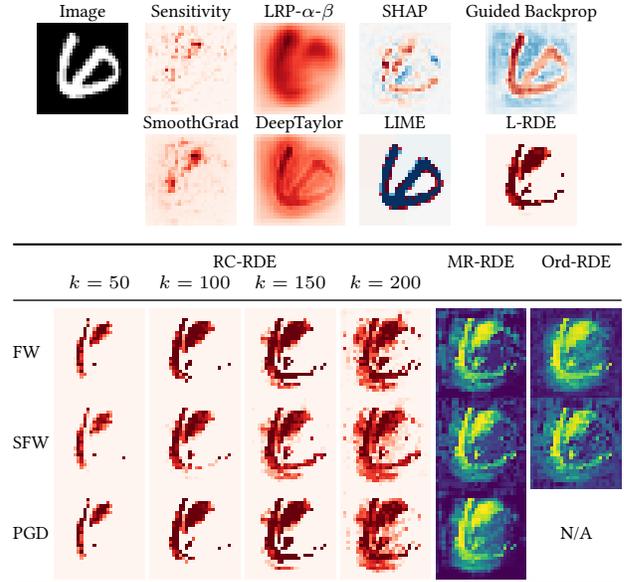
\begin{figure}
    \centering\scriptsize
    \input{./img/mnist/mnist_examples}

    \caption{Relevance mappings for an MNIST image classified as \emph{digit six} by the network. The colormap indicates positive relevance as red and negative relevance as blue. Multi-rate solutions are shown in a different colormap to highlight the fact, that they are not to be viewed as sparse relevance maps but as component orderings from least relevant (blue) to most relevant (yellow). Results for the other FW variants are shown in Figure~\ref{fig:mnist_examples_extra} in the appendix.}
    \label{fig:mnist_examples}
\end{figure}

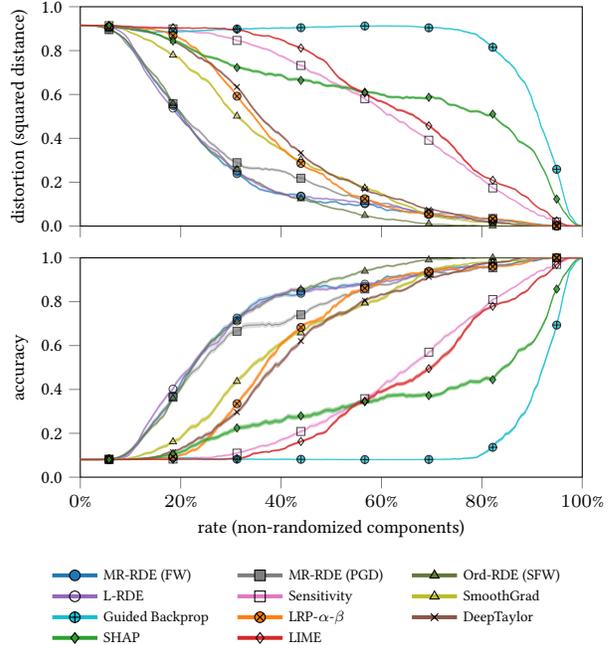
\begin{figure}
    \centering\scriptsize
    \input{./img/mnist/mnist_curves}
    \caption{Relevance ordering test results for MNIST for all considered methods. An average result over 50 images from the test set (5 images per class) and 512 noise input samples per image is shown (shaded regions mark $\pm$ standard deviation). A comparison of different FW variants for the RDE method is shown in Figure~\ref{fig:mnist_curves_extra} in the appendix.}
    \label{fig:mnist_curves}
\end{figure}

\begin{figure}
    \centering\scriptsize
    \input{./img/mnist/mnist_curves_rc}
    \caption{Relevance ordering test results for MNIST and \eqref{eq:rc_rde} at various rates. Vertical lines show the rates $k$ at which the mappings were optimized. The combined \eqref{eq:mr_rde} solution approximates a lower envelope of the individual curves. An average result over 50 images from the test set (5 images per class) and 512 noise input samples per image is shown (shaded regions mark $\pm$ standard deviation).}
    \label{fig:mnist_curves_rc}
\end{figure}
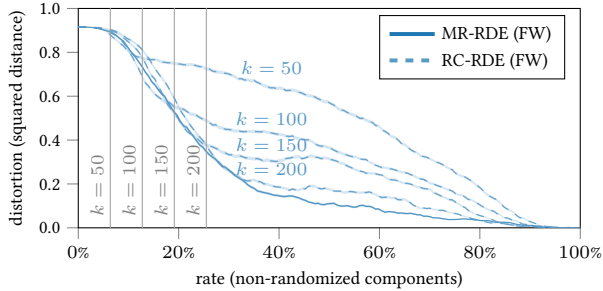

The mappings for an example image of a digit six are shown in Figure~\ref{fig:mnist_examples}. It shows \eqref{eq:rc_rde} mappings for FW and PGD at different selected rates $k$, as well as respective \eqref{eq:mr_rde} mappings with $\CK=\{50,\allowbreak 100,\allowbreak 150,\allowbreak 200,\allowbreak 250,\allowbreak 300,\allowbreak 350,\allowbreak 400\}$ and an \eqref{eq:ord_rde} mapping.\footnote{Images are $28\times 28$ greyscale, hence $n=28\cdot 28=784$.} The middle row shows the corresponding single-rate mappings $\bfPi^\text{opt}\bfp_k$ associated to the SFW solution of \eqref{eq:ord_rde} as well as a corresponding multi-rate mapping $\frac{1}{n-1}\sum_{k\in [n-1]}\bfPi^\text{opt}\bfp_k$. Mappings for the other FW variants are shown in Figure~\ref{fig:mnist_examples_extra} in the appendix. All RDE variants generate similar results and highlight an area at the top as relevant, that distinguishes a six from the digits zero and eight. All FW methods and PGD are robust across varying rates, in the sense that solutions for larger rates add additional features to the relevant set without significantly modifying the features that were already considered relevant at smaller rates. The \eqref{eq:l_rde} solution is most similar to the \eqref{eq:rc_rde} solutions at rate $k=100$.

The quantitative effect of solving \eqref{eq:rc_rde} for different rates is illustrated in Figure~\ref{fig:mnist_curves_rc}.
For the sake of clarity, we only show the relevance ordering test results for FW. The results for AFW, LCG, LAFW, and PGD are comparable.
The \eqref{eq:rc_rde} mappings achieve a low distortion at the rates for which they were optimized but are suboptimal at other rates. The combined \eqref{eq:mr_rde} solution approximates the lower envelope of the individual curves and performs best overall.

Figure~\ref{fig:mnist_curves} shows a comparison of the relevance ordering test results for two different performance measures (distortion at the top, classification accuracy at the bottom). Figure~\ref{fig:mnist_curves_extra} in the appendix shows the corresponding results for the other FW variants. All RDE methods result in a fast drop in the distortion (respectively a fast rise in the accuracy), indicating that the relevant features were correctly identified. They clearly outperform the other relevance attribution methods. The FW solutions perform slightly better than the PGD solutions. As expected, the \eqref{eq:ord_rde} solution performs best overall.

\paragraph{STL-10 Experiment} We consider the same VGG-16 network \autocite{sim2014vgg} pretrained on the Imagenet dataset and refined on STL-10 to a final test accuracy of 0.935 as used by \textcite{macdonald2019rde}. The relevance mappings are again calculated for the pre-softmax score of the class with the highest activation.

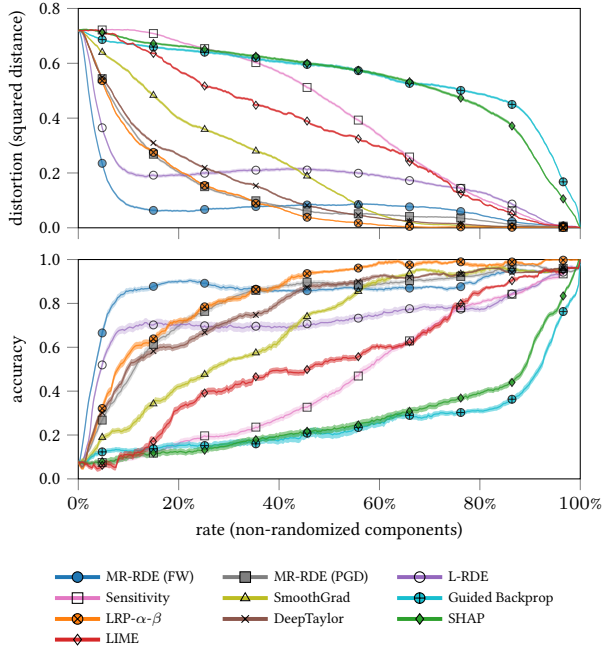
\begin{figure}
    \centering\scriptsize
    \input{./img/stl10/stl10_curves}
    \caption{Relevance ordering test results for STL-10. An average result over 50 images from the test set (5 images per class) and 64 noise input samples per image is shown (shaded regions mark $\pm$ standard deviation). A comparison of different FW variants for the RDE method is shown in Figure~\ref{fig:stl10_curves_extra}.}
    \label{fig:stl10_curves}
\end{figure}

\begin{figure}
    \centering\scriptsize
    \input{./img/stl10/stl10_examples3_short}

    \caption{Relevance mappings for an STL-10 image classified as \emph{monkey} by the network. The colormap indicates positive relevance as red and negative relevance as blue. Multi-rate solutions are shown in a different colormap to highlight the fact, that they are not to be viewed as sparse relevance maps but as component orderings from least relevant (blue) to most relevant (yellow). Results for the other FW variants are shown in Figure~\ref{fig:stl10_examples3and4} (left) in the appendix.}
    \label{fig:stl10_examples3_short}
\end{figure}
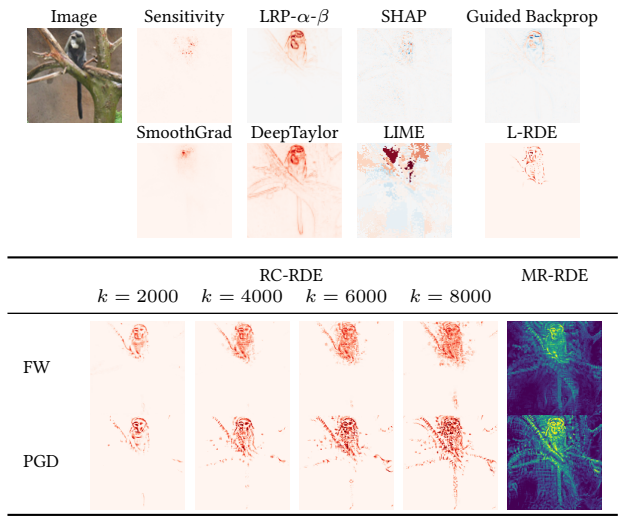

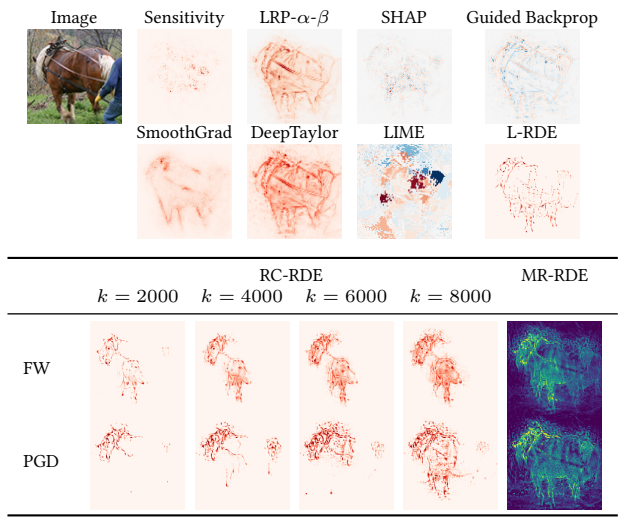
\begin{figure}
    \centering\scriptsize
    \input{./img/stl10/stl10_examples4_short}

    \caption{Relevance mappings for an STL-10 image classified as \emph{horse} by the network. The colormap indicates positive relevance as red and negative relevance as blue. Multi-rate solutions are shown in a different colormap to highlight the fact, that they are not to be viewed as sparse relevance maps but as component orderings from least relevant (blue) to most relevant (yellow). Results for the other FW variants are shown in Figure~\ref{fig:stl10_examples3and4} (right) in the appendix.}
    \label{fig:stl10_examples4_short}
\end{figure}

The mappings for two example images of a monkey and a horse are shown in Figures~\ref{fig:stl10_examples3_short} and~\ref{fig:stl10_examples4_short}. They show \eqref{eq:rc_rde} mappings for FW and PGD at different selected rates $k$, as well as respective \eqref{eq:mr_rde} mappings with $\CK=\{2000,\allowbreak 4000,\allowbreak 6000,\allowbreak 8000,\allowbreak 10000,\allowbreak 14000,\allowbreak 18000,\allowbreak 22000,\allowbreak 26000,\allowbreak 30000,\allowbreak 34000,\allowbreak 38000,\allowbreak 42000,\allowbreak 46000,\allowbreak 50000\}$.\footnote{Images are resized to $224\times 224$ with three color channels, hence $n=3\cdot 224\cdot 224=150528$. Mappings are visualized as a single channel heatmap that averages relevance scores across color channels.} Corresponding mappings for the other FW variants are shown in Figure~\ref{fig:stl10_examples3and4} in the appendix.
All RDE methods generate similar results and highlight parts of the face and body of the monkey as relevant at small rates. Increasingly large parts of the body and tail of the monkey are added to the relevant set at higher rates. Similarly, parts of the head and front legs of the horse are marked relevant first and larger parts of its body get added at higher rates. As before, the results are robust across varying rates. Additional images are shown in Appendix~\ref{apx:extra}.
Compared to MNIST, the difference between the FW and PGD solutions and between the Lagrangian and the rate-constrained formulations are more pronounced for STL-10. The sparsity of the \eqref{eq:l_rde} solution is most comparable to the \eqref{eq:rc_rde} solutions at rate $k=2000$. However, it is less concentrated at a specific part of the body of the animals, especially for the horse image.

Figure~\ref{fig:stl10_curves} shows the relevance ordering test results for two different performance measures (distortion at the top, classification accuracy at the bottom).
Figure~\ref{fig:stl10_curves_extra} in the appendix shows the corresponding results for the other FW variants.
Due to the large size of problem instances, we do not show \eqref{eq:ord_rde} solutions in this experiment.
All RDE methods result in a fast drop in the distortion (respectively a fast rise in the accuracy), indicating that the relevant features were correctly identified. Again, they clearly outperform the other relevance attribution methods, especially at the lower rates. The non-monotone behavior of the curves for rates between $30\%$ and $70\%$ can be explained by the fact that the maximal considered rate in $\CK$ corresponds to about $33\%$ of the total number of components. Unlike for MNIST, there is a considerable difference between the FW and PGD solutions for RDE and between \eqref{eq:l_rde} and \eqref{eq:mr_rde}. The latter outperforms the original RDE version across all rates. FW outperforms PGD for rates up to $40\%$. Results are comparable for higher rates.

% ----- ----- ----- ----- -----
\section{Discussion}\label{sec:discussion}
We have proposed and analyzed a rate-constrained and a relevance-ordering variation of Rate-Distortion Explanations (RDE) for interpretable neural networks. Solutions can be efficiently obtained using Frank-Wolfe algorithms. While optimization based relevance attribution methods, such as RDE, are computationally more demanding than single-pass methods, such as, e.g., LRP and Guided Backprop, we believe that this does not hinder their practical use. On the contrary, meaningful interpretations and excellent performance in quantitative evaluations are of more interest in critical applications than mere runtimes. Further, we observe that already very few Franke-Wolfe iterations are sufficient for obtaining accurate RDE mappings, cf.\ Figure~\ref{fig:mnist_iter} in the appendix. These can be computed within seconds, which renders RDE feasible for all applications that do not require real-time interpretations.

\subsubsection*{Acknowledgements}
{\small Research reported in this paper was partially supported through the Research Campus Modal funded by the German Federal Ministry of Education and Research (fund numbers 05M14ZAM, 05M20ZBM) and the Deutsche Forschungsgemeinschaft (DFG) through the DFG Cluster of Excellence MATH+.}

\subsubsection*{References}

\printbibliography[heading=none]

% ----- ----- ----- ----- -----
\newpage
\appendix
\onecolumn

% ----- ----- ----- ----- -----
\section{Technical Considerations}\label{apx:technical}

In this section, we give an overview of three deterministic variants of the Frank-Wolfe algorithm that offer potential improvements over vanilla Frank-Wolfe in terms of iteration count or runtime in certain specialized settings. Further, we specify all feasible regions considered in our experiments and present the corresponding linear minimization oracles for FW and projections for PGD.

\subsection{Variations of the Frank-Wolfe Algorithm}
Recall that Frank-Wolfe algorithms, at their core, solve a linear minimization oracle
\begin{equation}
    \bfv_t = \argmin_{\bfv \in \CC} \left\langle \nabla D(\bfs_t), \bfv \right\rangle, \tag{\ref{eq:lmo}, restated}
\end{equation}
over the feasible region $\CC$ and then move in direction $\bfv_t$ via the update
\begin{equation*}
    \bfs_{t+1} = \bfs_t + \eta_t(\bfv_t-\bfs_t),
\end{equation*}
that is a convex combination of $\bfv_t$ and the iterate $\bfs_{t}$.

Complementing the descriptions of the Vanilla Frank-Wolfe (FW) and Stochastic Frank-Wolfe (SFW) algorithms in Section~\ref{sec:fw_algos} we also consider the following three deterministic Frank-Wolfe variants in our experiments.

\paragraph{Away-Step Frank-Wolfe (AFW)} While the vanilla Frank-Wolfe algorithm can only move \emph{towards} an extreme point of the feasible set (solution of the LMO at the current iterate),
the Away-Step Frank-Wolfe algorithm \autocite{wolfe1970afw,guelat1986afw,lacostejulien2015afw} is allowed to move \emph{away} from some extreme points.
More specifically, it maintains an \emph{active set} of extreme points used in the previous iterations as well as a convex decomposition of the current iterate in terms of the active set. At each iteration either a standard FW step towards a new extreme point or a step away from an extreme point in the active set is taken, whichever promises a better decrease in the objective function. This can result in faster convergence (in terms of iteration count and time) but requires additional memory to store the active set.

\paragraph{Lazified Conditional Gradients (LCG)} In some cases the evaluation of the LMO might be costly (even if it is still cheaper than a corresponding projection).
In such a setting, the idea of \emph{lazy} FW steps can help to avoid unnecessary evaluations of the LMO. Instead of exactly solving the LMO subproblem, an approximate solution that guarantees enough progress is used \autocite{braun2019lcg}. In other words, the LMO can be replaced by a weak separation oracle \autocite{braun2019lcg}, i.e., an oracle returning an extreme point with sufficient decrease of the linear objective or a certificate that such a point does not exist.
More precisely, the algorithm maintains a cache of previous extreme points and at each iteration searches the cache for a direction that provides sufficient progress. If this is not possible, a new extreme point is obtained via the LMO and added to the cache. The lazification can result in increased performance (due to fewer LMO evaluations) but requires additional memory to store the cache of previous extreme points.

\paragraph{Lazified Away-Step Frank-Wolfe (LAFW)} LAFW uses the same idea of a weak separation oracle as in LCG. The search for an appropriate direction providing sufficient progress is carried out over the active set of AFW.

\subsection{Feasible Regions and Linear Minimization Oracles}
Two different feasible sets are of importance in this work. For both,
we give a brief definition and description of the corresponding linear minimization oracle below.

\paragraph{$k$-sparse polytope} For $k\in[n]$, the \emph{$k$-sparse polytope of radius $\tau>0$} is the intersection of the closed $\ell_1$-ball $B_1(\tau k)$ of radius $\tau k$ and the
closed $\ell_\infty$-ball (hypercube) $B_\infty(\tau)$ of radius $\tau$. It is the convex hull of vectors in $\R^n$ with exactly $k$ non-zero entries, each taking the values $\tau$ or $-\tau$.

\smallskip

\noindent
\emph{LMO:} A valid solution $\bfv$ to \eqref{eq:lmo} for the $k$-sparse polytope is given by the vector with exactly $k$ non-zero entries at the components where $|\nabla D(\bfs)|$ takes its $k$ largest values. If $v_j$ is such a non-zero entry then it is equal to $-\tau\sign((\nabla D(\bfs))_j)$. The complexity of finding this solution is $\CO(n \log k)$.

\smallskip

\noindent
\emph{Projection:} \Textcite{gupta2010l1boxprojection} propose an $\CO(n)$ algorithm for projections onto $\ell_1$-balls with box-constraints, extending the work of \textcite{duchi2008l1projection} on efficient projections onto $\ell_1$-balls. It is based on linear time median finding, see e.g. \autocite{clrs2009algorithms}. A slightly simplified $\CO(n\log n)$ variant based on sorting is used in our implementation.
\medskip

\noindent
More relevant to us is the following variation:
\paragraph{Non-negative $k$-sparse polytope:} For $k\in[n]$ and $\tau>0$ the \emph{non-negative $k$-sparse polytope of radius $\tau$} is defined as the intersection of the $k$-sparse polytope of radius $\tau$ with the non-negative orthant $\R^n_{\geq 0}$.

\smallskip

\noindent
\emph{LMO:} A valid solution $\bfv$ to \eqref{eq:lmo} for the non-negative $k$-sparse polytope is given by the vector with at most $k$ non-zero entries at the components where $\nabla D(\bfs)$ is negative and takes its $k$ smallest values (thus largest in magnitude as above). If $\nabla D(\bfs)$ has fewer than $k$ negative entries, then $\bfv$ has fewer than $k$ non-zero entries. If $v_j$ is a non-zero entry then it is equal to $\tau$. As above the complexity of finding this solution is $\CO(n \log k)$.

\smallskip

\noindent
\emph{Projection:} The same algorithm for projections onto $\ell_1$-balls with box-constraints as above can be used.

\medskip

\noindent
The feasible region $\CC = \{\bfs\in[0,1]^n\,:\,\|\bfs\|_1\leq k\}$ for the \eqref{eq:rc_rde} problem is the non-negative $k$-sparse polytope of radius $\tau=1$.
\paragraph{Birkhoff polytope} The \emph{Birkhoff polytope} $B_n$ is the set of doubly-stochastic $(n\times n)$-matrices. It is the convex hull of the set of $(n\times n)$-permutation matrices.

\smallskip

\noindent
\emph{LMO:} The Birkhoff polytope arises in matching and ranking problems. Linear minimization over $B_n$ results in a linear program, which can be solved with $\CO(n^3)$ complexity using the Hungarian method \autocite{kuhn1955hungarian} implemented in the \texttt{Hungarian.jl} package.
Linear minimization can also be performed using the standard or network simplex algorithms, opening the possibility for optimized and potentially parallelizable implementations.
In our experiments we found that the LMO was nonetheless more efficient, both in terms of runtime and memory footprint, when using the Hungarian algorithm compared to off-the-shelf simplex solvers.

\smallskip

\noindent
\emph{Projection:} To the best of our knowledge, there is no exact projection method specific to the Birkhoff polytope. An approximate method based on the Douglas-Ratchford splitting algorithm was proposed by \textcite{combettes2021lmo_proj}. Its complexity to achieve $\epsilon$-convergence is $\CO(n^2 c^2 / \epsilon^2)$ where $c$ is not known a-priory and depends on the distance of the initial guess for the algorithm and a fixed point of the proximal operator evaluated in each iteration.

\medskip

\noindent
The set $B_n$ is used as the feasible region for the \eqref{eq:ord_rde} problem.

\subsection{Computational Setup}
We give a brief specification of our computational setup. All experiments were run on computer cluster nodes with a Nvidia Quadro RTX 6000 GPU and an AMD EPYC 7262 or AMD EPYC 7302P CPU. We use \texttt{Julia (v1.6.1)} and the \texttt{FrankWolfe.jl (v0.1.8)} package for all variants of FW algorithms. The \texttt{Python (v3.7.8)} packages \texttt{Tensorflow (v1.15.0)} and \texttt{Keras (v2.2.4)} are used for the neural network classifiers and computation of gradients through automatic differentiation. The interactions between \texttt{Julia} and \texttt{Python} are handled through the \texttt{PyCall.jl (v1.92.3)} package.

% ----- ----- ----- ----- -----
\section{Additional Computational Results}\label{apx:results}
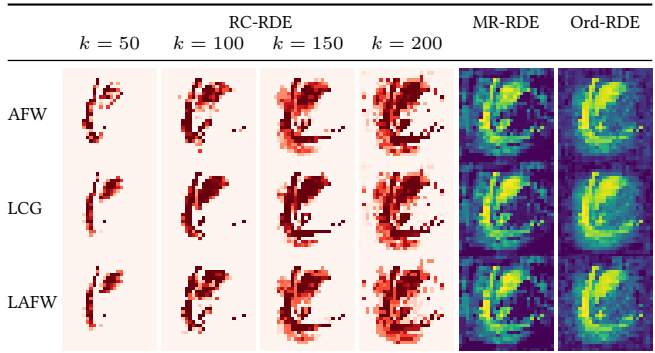
\begin{figure}
    \centering\scriptsize
    \input{./img/mnist/mnist_examples_additional_fw}

    \caption{Additional relevance mappings for the MNIST image from Figure~\ref{fig:mnist_examples} classified as \emph{digit six} by the network. Multi-rate solutions are shown in a different colormap to highlight the fact, that they are not to be viewed as sparse relevance maps but as component orderings from least relevant (blue) to most relevant (yellow).}
    \label{fig:mnist_examples_extra}
\end{figure}

\begin{figure}
    \centering\scriptsize
    \input{./img/stl10/stl10_curves_extra}
    \caption{Relevance ordering test results for STL-10 for RDE with different FW variants and PGD. An average result over 50 images from the test set (5 images per class) and 64 noise input samples per image is shown (shaded regions mark $\pm$ standard deviation). This complements Figure~\ref{fig:stl10_curves}.}
    \label{fig:stl10_curves_extra}
\end{figure}
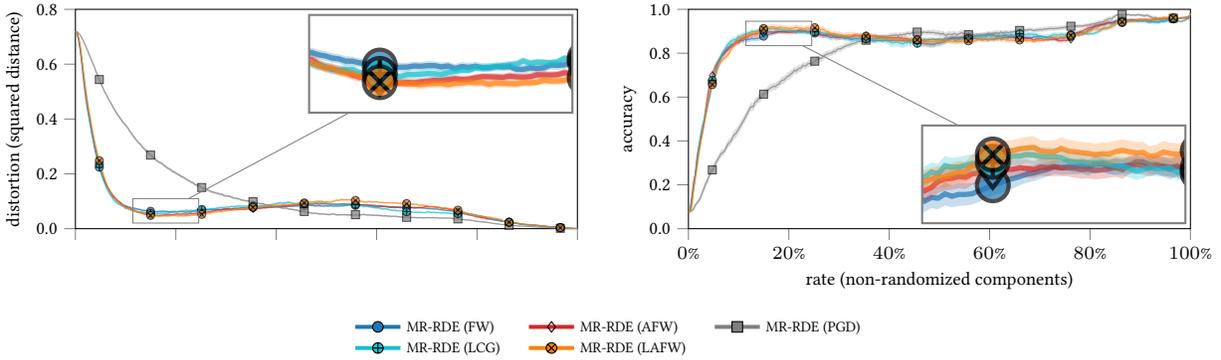

Figure~\ref{fig:mnist_examples_extra} shows a comparison of results obtained using the different FW variants for solving \eqref{eq:rc_rde} and \eqref{eq:mr_rde} for the MNIST experiment example image from Figure~\ref{fig:mnist_examples}. All FW variants yield similar results and are robust across varying rates, in the sense that solutions for larger rates add additional features to the relevant set without significantly modifying the features that were already considered relevant at smaller rates. Figures~\ref{fig:mnist_curves_extra} and~\ref{fig:stl10_curves_extra} complement Figures~\ref{fig:mnist_curves} and~\ref{fig:stl10_curves} and show shows the relevance ordering test results for all FW variants for the MNIST and STL-10 experiment respectively.

A comparison of runtimes for the MNIST experiment and the corresponding numbers of iterations that are taken until the termination criterion $\langle\bfs_t-\bfv_t,\nabla D(\bfs_t)\rangle < \varepsilon=10^{-7}$ is reached for the different FW variants is shown in Figure~\ref{fig:mnist_iter}(a). We observe that AFW converges fastest for small rates $k$, while all variants perform similarly at higher rates (with slight advantages of FW and LCG over AFW and LAFW).
This is due to a reduced number of iterations of AFW and LAFW at small rates. On the other hand, the increased runtime of the active-set methods AFW and LAFW can be explained by the fact that each vertex in the active set is a sparse vector with more and more non-zero components as the rate increases. Hence, active set operations require more arithmetic operations at higher rates.
All methods reach the maximum number of  $T=2000$ iterations before satisfying the termination criterion at high rates. This can be explained by the fact that the termination threshold $\varepsilon=10^{-7}$ is chosen quite conservatively to ensure convergence of all methods on all instances. However, we observe that all methods typically converge much faster to a satisfactory solution. Figure~\ref{fig:mnist_iter}(b) shows \eqref{eq:rc_rde} solutions at a single exemplary rate $k=150$ for the MNIST image from Figure~\ref{fig:mnist_examples} after $T=50$, $100$, $150$, and $200$ iterations. The solutions are visually indistinguishable for all methods, confirming that they were already mostly converged after only $50$ iterations. We do not show a comparable analysis of runtimes and iteration counts for the STL-10 experiment. Due to higher computational costs, the STL-10 mappings were computed in parallel on different machines. Hence, no runtimes that are directly comparable between the different methods are available.

\begin{figure}
    \centering\scriptsize
    \begin{tabular}{@{}c@{\;\;}c@{}}
        \input{./img/mnist/mnist_eval} &
        \input{./img/mnist/mnist_examples_iterations} \\
        (a) & (b)
    \end{tabular}
    \caption{(a) Average runtimes (top) and number of iterations until convergence (bottom) of the considered FW variants for \eqref{eq:rc_rde} on MNIST at different rates. An average result over 50 images from the test set is shown (shaded regions mark $\pm$ standard deviation). (b) Relevance mappings obtained at rate $k=150$ after different maximal numbers of iterations. Results for the same MNIST image from Figure~\ref{fig:mnist_examples} classified as \emph{digit six} by the network are shown. All methods were converged effectively already after only $T=50$ iterations.}
    \label{fig:mnist_iter}
\end{figure}
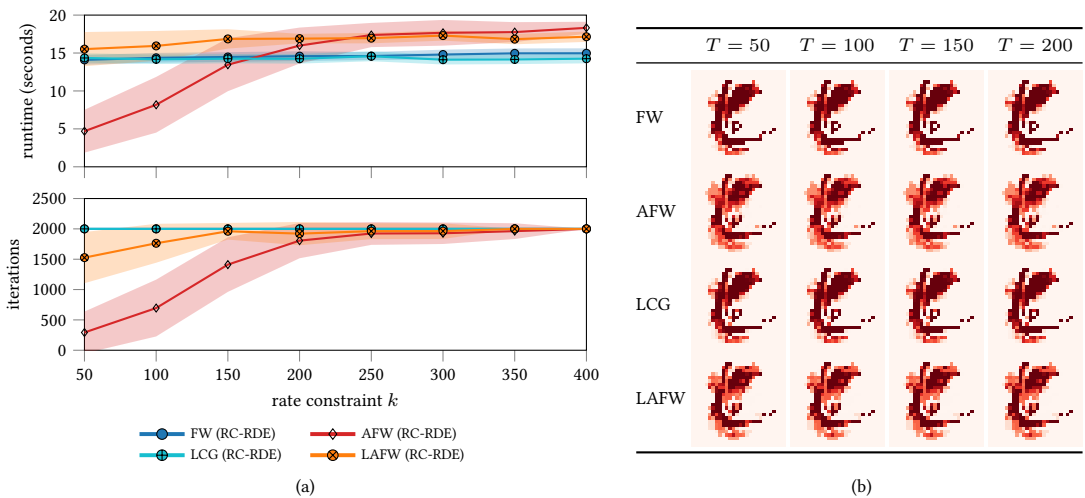

\begin{figure}
    \centering\scriptsize
    \input{./img/mnist/mnist_curves_extra}
    \caption{Relevance ordering test results for MNIST for RDE with different FW variants and PGD. An average result over 50 images from the test set (5 images per class) and 512 noise input samples per image is shown (shaded regions mark $\pm$ standard deviation). This complements Figure~\ref{fig:mnist_curves}.}
    \label{fig:mnist_curves_extra}
\end{figure}
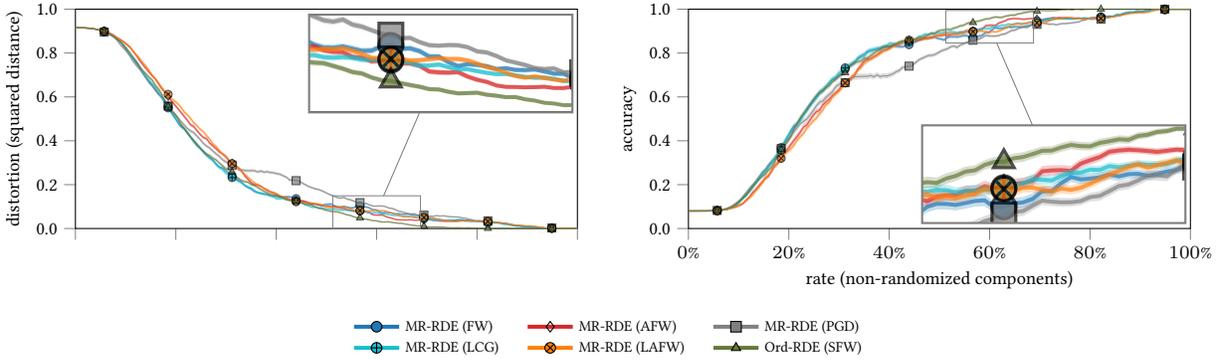

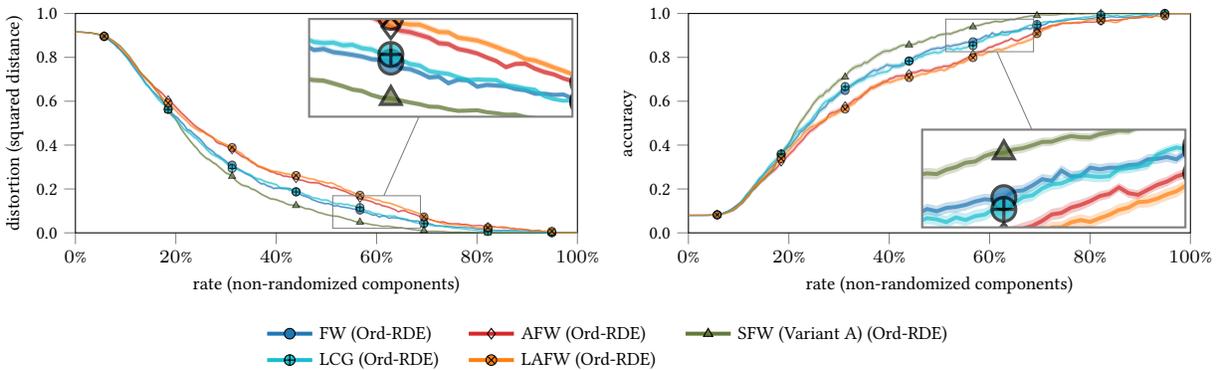
\begin{figure}
    \centering\scriptsize
    \input{./img/mnist/mnist_curves_order}
    \caption{Relevance ordering test results for \eqref{eq:ord_rde} on the MNIST dataset for all considered FW variants. An average result over 50 images from the test set (5 images per class) and 512 noise input samples per image is shown (shaded regions mark $\pm$ standard deviation). This complements Figure~\ref{fig:mnist_curves}, which contains the same SFW result.}
    \label{fig:mnist_curves_order}
\end{figure}

\begin{figure}
    \centering\scriptsize
    \input{./img/mnist/mnist_curves_sfw}
    \caption{Relevance ordering test results for \eqref{eq:ord_rde} on MNIST for different variants of SFW. An average result over 50 images from the test set (5 images per class) and 512 noise input samples per image is shown (shaded regions mark $\pm$ standard deviation). This complements Figure~\ref{fig:mnist_curves}, which contains the same SFW results with constant batch size and no momentum.}
    \label{fig:mnist_curves_sfw}
\end{figure}
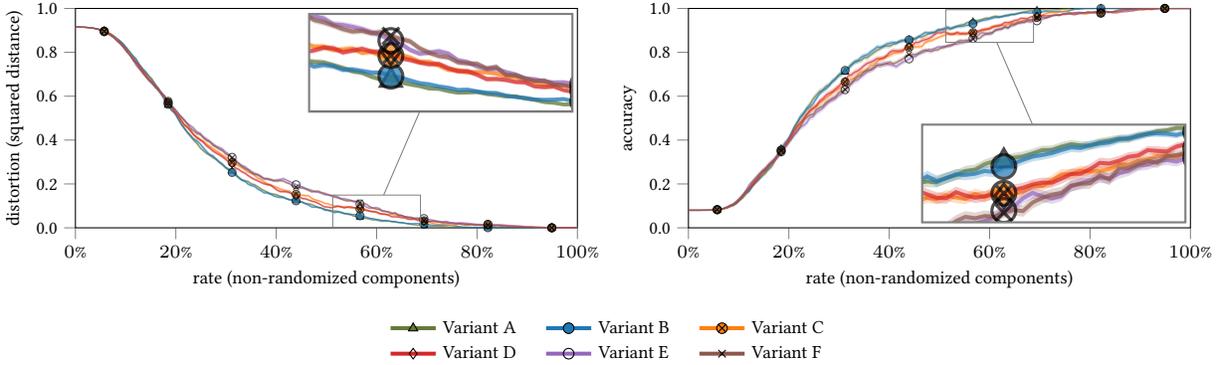

The relevance-ordering problem \eqref{eq:ord_rde} can be solved with a stochastic Frank-Wolfe algorithm (as shown in Section~\ref{sec:results}) or with deterministic Frank-Wolfe algorithms if the number $n$ of terms in the objective function is not too large. For the MNIST dataset, both approaches are feasible. We compare the relevance-ordering comparison test results of all FW variants in Figure~\ref{fig:mnist_curves_order}. The SFW result is the same as in Figure~\ref{fig:mnist_curves}. We observe that all variants perform well and similarly for very low and high rates. For rates between $20\%$ and $80\%$ of the total number of components SFW has an advantage over FW and LCG which in turn perform slightly better than AFW and LAFW.

Further, we compare different hyperparameter configurations for SFW regarding the batch sizes $b_t$ and momentum factors $\rho_t$. \Textcite{hazan2016sfw} suggest a linearly increasing\footnote{We limit the batch size growth up to a maximal size $b_\text{max}=100$ in our experiments.} batch size and \textcite{mokhtari2020sfw} propose a momentum factor $\rho_t=1-4 / (8+t)^{\frac{2}{3}}$ approaching one. We consider the following combinations:
\begin{center}
    \scriptsize
    \begin{tabular}{lcccccc}
        \toprule
        & Variant A & Variant B & Variant C & Variant D & Variant E & Variant F \\
        \midrule
        momentum & $\rho_t=0$ & $\rho_t=0$ & $\rho_t=\frac{1}{2}$ & $\rho_t=\frac{1}{2}$ & $\rho_t=1-\frac{4}{(8+t)^{\frac{2}{3}}}$ & $\rho_t=1-\frac{4}{(8+t)^{\frac{2}{3}}}$ \\
        batch size & $b_t=40$ & $b_t = \min\left\{40+t,100\right\}$ & $b_t=40$ & $b_t = \min\left\{40+t,100\right\}$ & $b_t=40$ & $b_t = \min\left\{40+t,100\right\}$ \\
        \bottomrule
    \end{tabular}
\end{center}
The relevance-ordering comparison test results are shown in Figure~\ref{fig:mnist_curves_sfw}. For reference Variant A corresponds to the SFW result also shown in Figure~\ref{fig:mnist_curves} and~\ref{fig:mnist_curves_order}. We observe, that all variants perform well and similarly across all rates. The batch size has negligible effect in this experiment, while momentum yields no advantage. In particular, both configurations without any momentum perform best. Hence, we show only Variant A (no momentum, constant batch size) in all other experiments.

% ----- ----- ----- ----- -----
\section{Additional STL-10 Relevance Maps}\label{apx:extra}
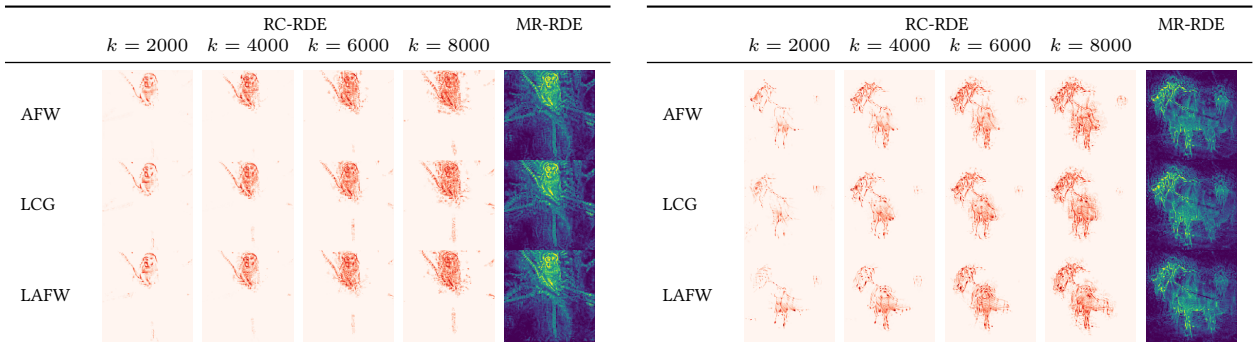
\begin{figure}
    \centering\scriptsize
    \begin{tabular}{cc}
        \input{./img/stl10/stl10_examples3} &
        \input{./img/stl10/stl10_examples4}

    \end{tabular}
    \caption{Relevance mappings for STL-10 images classified as \emph{monkey} and \emph{horse} by the network. This complements Figures~\ref{fig:stl10_examples3_short} and~\ref{fig:stl10_examples4_short}.}
    \label{fig:stl10_examples3and4}
\end{figure}

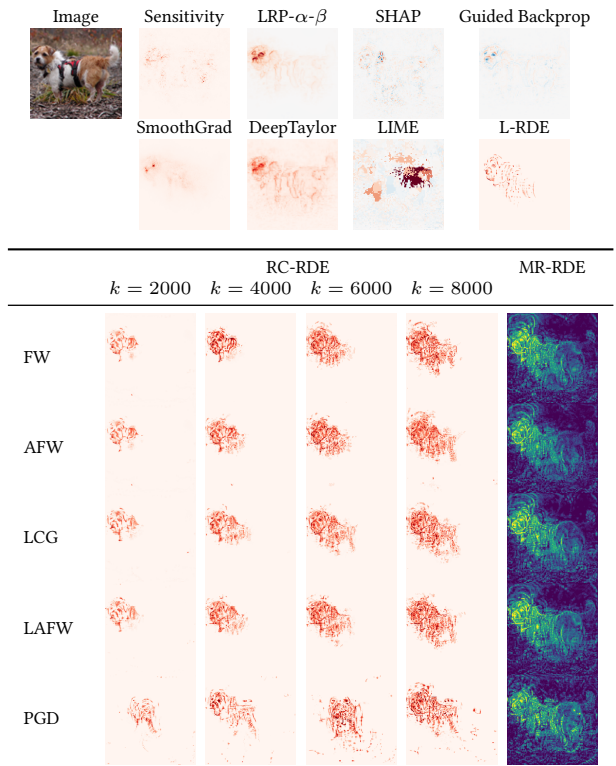
\begin{figure}[!p]
    \centering\scriptsize
    \input{./img/stl10/stl10_examples1}

    \caption{Relevance mappings for an STL-10 image classified as \emph{dog} by the network.}
    \label{fig:stl10_examples1}
\end{figure}

\begin{figure}[!p]
    \centering\scriptsize
    \input{./img/stl10/stl10_examples7}

    \caption{Relevance mappings for an STL-10 image classified as \emph{deer} by the network.}
    \label{fig:stl10_examples7}
\end{figure}
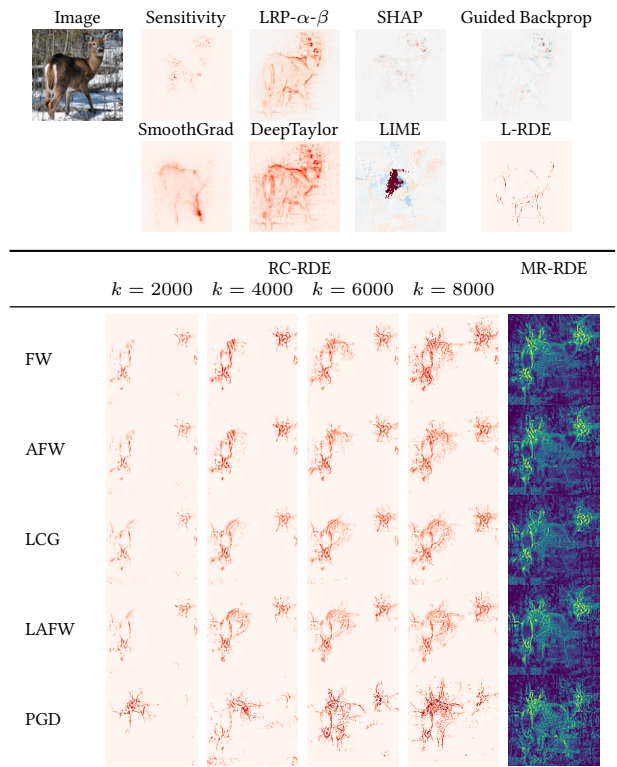

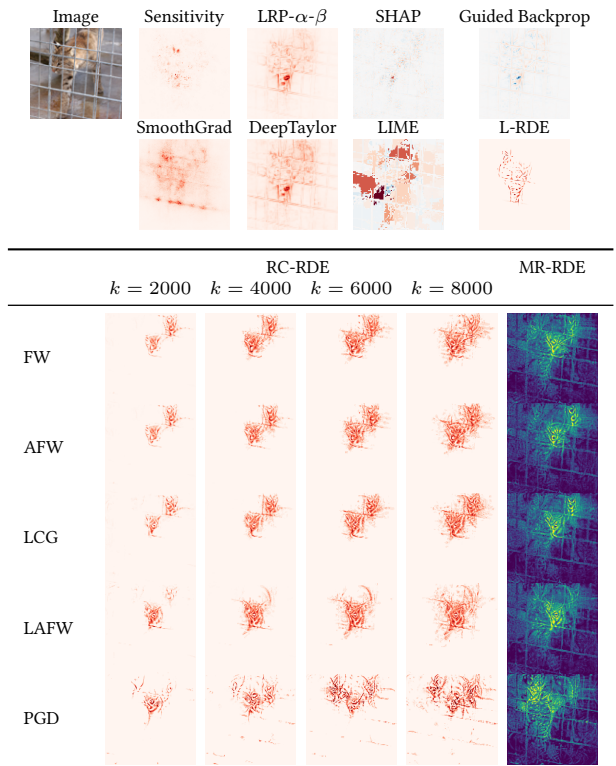
\begin{figure}[!p]
    \centering\scriptsize
    \input{./img/stl10/stl10_examples6}

    \caption{Relevance mappings for an STL-10 image classified as \emph{cat} by the network.}
    \label{fig:stl10_examples6}
\end{figure}

\begin{figure}[!p]
    \centering\scriptsize
    \input{./img/stl10/stl10_examples10}

    \caption{Relevance mappings for an STL-10 image classified as \emph{airplane} by the network.}
    \label{fig:stl10_examples10}
\end{figure}
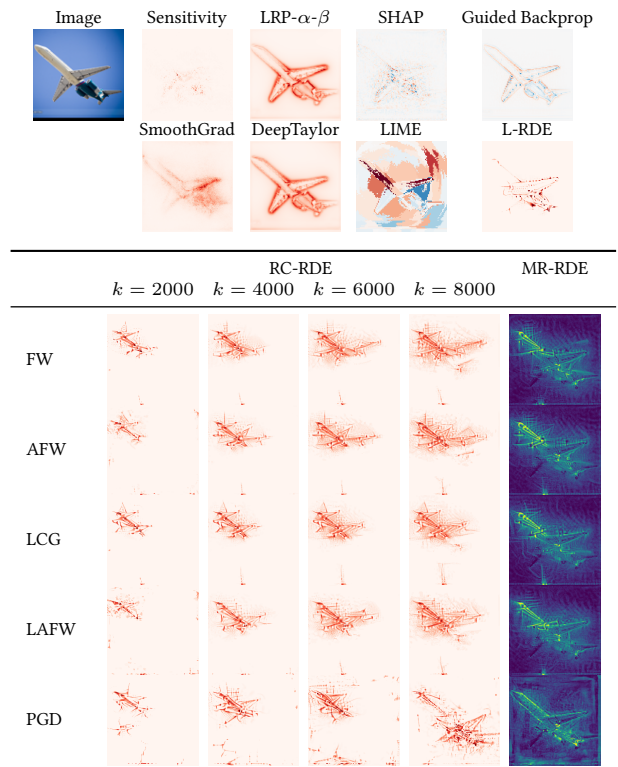

We complement Figures~\ref{fig:stl10_examples3_short} and~\ref{fig:stl10_examples4_short} by Figures~\ref{fig:stl10_examples3and4}--\ref{fig:stl10_examples10} showing additional examples of relevance maps for images of different classes from the STL-10 dataset as well as additional results for all FW variants for the images from Figures~\ref{fig:stl10_examples3_short} and~\ref{fig:stl10_examples4_short}.
The colormap indicates positive relevance as red and negative relevance as blue.
All multi-rate solutions are shown in a different colormap to highlight the fact, that they are not to be viewed as sparse relevance maps but as component orderings from least relevant (blue) to most relevant (yellow). All \eqref{eq:l_rde} solutions and results for the other comparison methods are provided by \textcite{macdonald2019rde}.

\twocolumn

\begin{figure}[!p]
    \centering\scriptsize
    \input{./img/stl10/stl10_examples2}

    \caption{Relevance mappings for an STL-10 image classified as \emph{bird} by the network.}
    \label{fig:stl10_examples2}
\end{figure}

\begin{figure}[!p]
    \centering\scriptsize
    \input{./img/stl10/stl10_examples5}

    \caption{Relevance mappings for an STL-10 image classified as \emph{ship} by the network.}
    \label{fig:stl10_examples5}
\end{figure}

\begin{figure}[!p]
    \centering\scriptsize
    \input{./img/stl10/stl10_examples8}

    \caption{Relevance mappings for an STL-10 image classified as \emph{deer} by the network.}
    \label{fig:stl10_examples8}
\end{figure}

\begin{figure}[!p]
    \centering\scriptsize
    \input{./img/stl10/stl10_examples9}

    \caption{Relevance mappings for an STL-10 image classified as \emph{horse} by the network.}
    \label{fig:stl10_examples9}
\end{figure}

\end{document}

%% file: img/classifier-relevance.tex
\begin{tikzpicture}
  \begin{pgfonlayer}{fg}
      \node[inner sep=0cm, outer sep=.2cm] (img) at (-4,0) {\includegraphics[width=1.25cm]{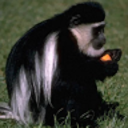}};
      \node[right=0.8cm of img, rounded corners=5pt, outer sep=.2cm, draw, rectangle, fill=black, text=white, minimum width=1.25cm, inner sep=.35cm] (phi) {$\Phi$};
      \node[right=0.8cm of phi, inner sep=0cm, outer sep=.2cm] (out) {``monkey''};

      \draw[->, >=stealth, thick] (img) -- (phi);
      \draw[->, >=stealth, thick] (phi) -- (out);

      \node[below=.35cm of img.south west, anchor=north west, rounded corners=5pt, outer sep=.2cm, draw=tubred, fill=tubred, text=white, rectangle, minimum width=3.7cm, inner sep=.35cm] (map) {relevance attribution};
      \node[inner sep=0cm, outer sep=.2cm, anchor=east] (rel) at (out.east|-map.east) {\includegraphics[width=1.25cm]{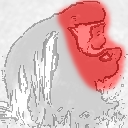}};

      \draw[->, >=stealth, thick] (img.south) -- (img.south|-map.north);
      \draw[->, >=stealth, thick] (phi.south) -- (phi.south|-map.north);
      \draw[->, >=stealth, thick] (map) -- (rel);

      \node[right=.15cm of out, inner sep=0cm, outer sep=.2cm, darkgreen, rotate=90, anchor=north] {classification};
      \node[right=.15cm of rel, inner sep=0cm, outer sep=.2cm, bmsblue, rotate=90, anchor=90] {interpretation};

  \end{pgfonlayer}

  \draw[dashed, rounded corners=5pt, darkgreen, fill=darkgreen, fill opacity=.15] (img.north west) -- (img.north west-|out.north east) -- (img.south west-|out.south east) -- (img.south west) -- cycle;
  \draw[dashed, rounded corners=5pt, bmsblue,  fill=bmsblue, fill opacity=.15] (img.north west) -- (img.north west-|phi.north east) -- (phi.south east|-rel.north west) -- (rel.north east) -- (rel.south east) -- (img.north west|-rel.south east) -- cycle;

\end{tikzpicture}

%% file: img/rate-distortion.tex
\begin{tikzpicture}

  % Alice
  \node[draw, minimum width=1cm] (A) at (-1.65, 0) {Alice};

  % Bob
  \node[draw, minimum width=1cm] (B) at (1.65, 0) {Bob};

  \draw[->, >=stealth,  thick] ($(A) + (0.7,0.0)$) -- ($(B) + (-0.7,0.0)$) node[midway, above] {rate $R(\bfs)$};

  \node[below=0.2cm of A] (img) {\includegraphics[width=1cm]{./img/stl10_training_384.png}};
  \node[below=0.2cm of B] (nimg) {\includegraphics[width=1cm]{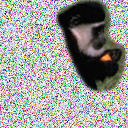}};
  \node at ($(img)!0.5!(nimg)$) (simg) {\includegraphics[width=1cm]{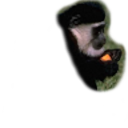}};

  \node[below=0.0cm of img] {\parbox{1.5cm}{\tiny\centering original \\ image}};
  \node[below=0.0cm of simg] {\parbox{1.5cm}{\tiny\centering partial \\ image}};
  \node[below=0.0cm of nimg] {\parbox{1.5cm}{\tiny\centering random \\ completion}};

  \node[left=0.1cm of img] (phix) {$\Phi(\bfx) = 0.97$};
  \node[below=0.0cm of phix] {\tiny ``monkey''};
  \node[right=0.1cm of nimg] (phiy) {$\Phi(\bfy) = 0.91$};
  \node[below=0.0cm of phiy] {\tiny ``monkey''};

\end{tikzpicture}

%% file: img/mnist/mnist_examples.tex
\begin{tabular}{c@{\;\;\;}c@{\;\;\;}c@{\;\;\;}c@{\;\;\;}c}
    Image & Sensitivity & LRP-$\alpha$-$\beta$ & SHAP & Guided Backprop \\
    \includegraphics[width=1.20cm, valign=c]{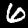} &
    \includegraphics[width=1.20cm, valign=c]{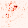} &
    \includegraphics[width=1.20cm, valign=c]{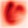} &
    \includegraphics[width=1.20cm, valign=c]{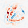} &
    \includegraphics[width=1.20cm, valign=c]{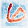} \\
    & SmoothGrad & DeepTaylor & LIME & L-RDE \\
    &
    \includegraphics[width=1.20cm, valign=c]{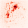} &
    \includegraphics[width=1.20cm, valign=c]{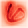} &
    \includegraphics[width=1.20cm, valign=c]{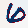} &
    \includegraphics[width=1.20cm, valign=c]{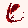}
\end{tabular}
\\[2ex]
\begin{tabular}{@{}l@{\;}c@{\;}c@{\;}c@{\;}c@{\;}c@{\;}c@{}}
    \toprule
    & \multicolumn{4}{c}{RC-RDE} & MR-RDE & Ord-RDE \\
    & $k=50$ & $k=100$ & $k=150$ & $k=200$ & & \\
    \midrule
    FW &
    \includegraphics[width=1.20cm, valign=c]{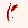} &
    \includegraphics[width=1.20cm, valign=c]{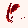} &
    \includegraphics[width=1.20cm, valign=c]{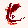} &
    \includegraphics[width=1.20cm, valign=c]{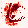} &
    \includegraphics[width=1.20cm, valign=c]{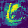} &
    \includegraphics[width=1.20cm, valign=c]{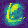} \\

    SFW &
    \includegraphics[width=1.20cm, valign=c]{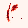} &
    \includegraphics[width=1.20cm, valign=c]{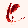} &
    \includegraphics[width=1.20cm, valign=c]{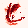} &
    \includegraphics[width=1.20cm, valign=c]{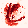} &
    \includegraphics[width=1.20cm, valign=c]{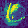} &
    \includegraphics[width=1.20cm, valign=c]{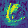} \\

    PGD &
    \includegraphics[width=1.20cm, valign=c]{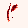} &
    \includegraphics[width=1.20cm, valign=c]{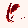} &
    \includegraphics[width=1.20cm, valign=c]{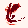} &
    \includegraphics[width=1.20cm, valign=c]{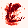} &
    \includegraphics[width=1.20cm, valign=c]{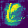} &
    N/A \\
    \bottomrule
\end{tabular}

%% file: img/mnist/mnist_curves.tex
\begin{tikzpicture}[spy using outlines={rectangle, gray, magnification=3, connect spies}]
\begin{groupplot}[group style={group size=1 by 2, vertical sep=4ex}]

    % mnist rate distortion
    \nextgroupplot[
        height=.75\figureheight,
        legend cell align={left},
        legend entries={
            {MR-RDE (FW)},
            {MR-RDE (PGD)},
            {Ord-RDE (SFW)},
            {L-RDE},
            {Sensitivity},
            {SmoothGrad},
            {Guided Backprop},
            {LRP-$\alpha$-$\beta$},
            {DeepTaylor},
            {SHAP},
            {LIME}
        },
        legend columns=3,
        legend style={anchor=north west, draw=none, line width=1.5, nodes={scale=0.8},
            /tikz/every even column/.append style={column sep=0.3cm}},
        legend to name=named-mnist,
        tick align=outside,
        tick pos=left,
        width=\figurewidth,
        x grid style={white!69.01960784313725!black},
        xticklabels=\empty,
        xmin=0, xmax=100,
        y grid style={white!69.01960784313725!black},
        ylabel={distortion (squared distance)},
        ymin=0.0, ymax=1.0,
        ytick={0,0.2,0.4,0.6,0.8,1},
        yticklabels={0.0,0.2,0.4,0.6,0.8,1.0}
    ]
    \addlegendimage{color0, mark=*, mark options={draw=black, fill=color0, thin}}
    \addlegendimage{color9, mark=square*, mark options={draw=black, fill=color9, thin}}
    \addlegendimage{color10, mark=triangle*, mark options={draw=black, fill=color10, thin}}
    \addlegendimage{color4, mark=o, mark options={draw=black, fill=color4, thin}}
    \addlegendimage{color6, mark=square,mark options={draw=black, fill=color6, thin}}
    \addlegendimage{color7, mark=triangle, mark options={draw=black, fill=color7, thin}}
    \addlegendimage{color8, mark=oplus*, mark options={draw=black, fill=color8, thin}}
    \addlegendimage{color1, mark=otimes*, mark options={draw=black, fill=color1, thin}}
    \addlegendimage{color5, mark=x, mark options={draw=black, fill=color5, thin}}
    \addlegendimage{color2, mark=diamond*, mark options={draw=black, fill=color2, thin}}
    \addlegendimage{color3, mark=diamond, mark options={draw=black, fill=color3, thin}}

    % mr-rde fw
    \input{./img/mnist/mnist_rate_distortion_fw_multi}
    % mr-rde pgd
    \input{./img/mnist/mnist_rate_distortion_pgd_multi}
    % ord-rde sfw
    \input{./img/mnist/mnist_rate_distortion_sfw_order_var_a}
    % l-rde
    \input{./img/mnist/mnist_rate_distortion_lagrange}
    % sensitivity
    \input{./img/mnist/mnist_rate_distortion_sensitivity}
    % smoothgrad
    \input{./img/mnist/mnist_rate_distortion_smoothgrad}
    % guided backprop
    \input{./img/mnist/mnist_rate_distortion_guidedbackprop}
    % lrp
    \input{./img/mnist/mnist_rate_distortion_lrp}
    % deep taylor
    \input{./img/mnist/mnist_rate_distortion_deeptaylor}
    % shap
    \input{./img/mnist/mnist_rate_distortion_shap}
    % lime
    \input{./img/mnist/mnist_rate_distortion_lime}

    %\coordinate (spyon1) at (axis cs:60,0.075);
    %\coordinate (spyat1) at (axis cs:99,0.975);
    %\spy [width=3.5cm, height=1.3cm] on (spyon1) in node[fill=white,anchor=north east] at (spyat1);

    % mnist rate accuracy
    \nextgroupplot[
        height=.75\figureheight,
        tick align=outside,
        tick pos=left,
        width=\figurewidth,
        xlabel={rate (non-randomized components)},
        x grid style={white!69.01960784313725!black},
        xticklabel={\pgfmathparse{\tick}\pgfmathprintnumber{\pgfmathresult}\%},
        xmin=0, xmax=100,
        y grid style={white!69.01960784313725!black},
        ylabel={accuracy},
        ymin=0.0, ymax=1.0,
        ytick={0,0.2,0.4,0.6,0.8,1},
        yticklabels={0.0,0.2,0.4,0.6,0.8,1.0}
    ]
    % mr-rde fw
    \input{./img/mnist/mnist_rate_accuracy_fw_multi}
    % mr-rde pgd
    \input{./img/mnist/mnist_rate_accuracy_pgd_multi}
    % ord-rde sfw 
    \input{./img/mnist/mnist_rate_accuracy_sfw_order_var_a}
    % l-rde
    \input{./img/mnist/mnist_rate_accuracy_lagrange}
    % sensitivity
    \input{./img/mnist/mnist_rate_accuracy_sensitivity}
    % smoothgrad
    \input{./img/mnist/mnist_rate_accuracy_smoothgrad}
    % guided backprop
    \input{./img/mnist/mnist_rate_accuracy_guidedbackprop}
    % lrp
    \input{./img/mnist/mnist_rate_accuracy_lrp}
    % deep taylor
    \input{./img/mnist/mnist_rate_accuracy_deeptaylor}
    % shap
    \input{./img/mnist/mnist_rate_accuracy_shap}
    % lime
    \input{./img/mnist/mnist_rate_accuracy_lime}
    
    %\coordinate (spyon2) at (axis cs:60,0.92);
    %\coordinate (spyat2) at (axis cs:99,0.025);
    %\spy [width=3.5cm, height=1.3cm] on (spyon2) in node[fill=white,anchor=south east] at (spyat2);
   
\end{groupplot}
\end{tikzpicture}

\begin{center}
  \ref*{named-mnist}
\end{center}
\vspace{-1.5em}

%% file: img/mnist/mnist_curves_rc.tex
\begin{tikzpicture}
\begin{groupplot}[group style={group size=1 by 1, horizontal sep=0em}]

    % mnist rate distortion
    \nextgroupplot[
        height=.75\figureheight,
        legend cell align={left},
        legend entries={
            {MR-RDE (FW)},
            {RC-RDE (FW)}
        },
        legend pos={north east},
        legend image post style={line width =1.5},
        tick align=outside,
        tick pos=left,
        width=\figurewidth,
        xlabel={rate (non-randomized components)},
        x grid style={white!69.01960784313725!black},
        xticklabel={\pgfmathparse{\tick}\pgfmathprintnumber{\pgfmathresult}\%},
        xmin=0, xmax=100,
        y grid style={white!69.01960784313725!black},
        ylabel={distortion (squared distance)},
        ymin=0.0, ymax=1.0,
        ytick={0,0.2,0.4,0.6,0.8,1},
        yticklabels={0.0,0.2,0.4,0.6,0.8,1.0}
    ]
    \addlegendimage{no markers, color0}
    \addlegendimage{no markers, dashed, color0!75}

    % rc rde fw
    \input{./img/mnist/mnist_rate_distortion_fw_multi_no_marker}
    \input{./img/mnist/mnist_rate_distortion_fw_50}
    \input{./img/mnist/mnist_rate_distortion_fw_100}
    \input{./img/mnist/mnist_rate_distortion_fw_150}
    \input{./img/mnist/mnist_rate_distortion_fw_200}

    \addplot [gray!75, forget plot]
    table {%
    6.377551 -0.05
    6.377551 1.05
    };
    \addplot [gray!75, forget plot]
    table {%
    12.75510 -0.05
    12.75510 1.05
    };
    \addplot [gray!75, forget plot]
    table {%
    19.13265 -0.05
    19.13265 1.05
    };
    \addplot [gray!75, forget plot]
    table {%
    25.51020 -0.05
    25.51020 1.05
    };

    \node[gray, left, anchor=south east] at (6.377551, 0.025) {\rotatebox{90}{$k=50$}};
    \node[gray, left, anchor=south east] at (12.75510, 0.025) {\rotatebox{90}{$k=100$}};
    \node[gray, left, anchor=south east] at (19.13265, 0.025) {\rotatebox{90}{$k=150$}};
    \node[gray, left, anchor=south east] at (25.51020, 0.025) {\rotatebox{90}{$k=200$}};

    \node[color0!75, above] at (38.5, 0.67) {$k=50$};
    \node[color0!75, above] at (38.5, 0.44) {$k=100$};
    \node[color0!75, above] at (38.5, 0.32) {$k=150$};
    \node[color0!75, above] at (38.5, 0.21) {$k=200$};

\end{groupplot}
\end{tikzpicture}

%% file: img/stl10/stl10_curves.tex
\begin{tikzpicture}[spy using outlines={rectangle, gray, magnification=4, connect spies}]
\begin{groupplot}[group style={group size=1 by 2, vertical sep=4ex}]

    % stl10 rate distortion
    \nextgroupplot[
        height=.75\figureheight,
        legend cell align={left},
        legend entries={
            {MR-RDE (FW)},
            {MR-RDE (PGD)},
            {L-RDE},
            {Sensitivity},
            {SmoothGrad},
            {Guided Backprop},
            {LRP-$\alpha$-$\beta$},
            {DeepTaylor},
            {SHAP},
            {LIME}
        },
        legend columns=3,
        legend style={anchor=north west, draw=none, line width=1.5, nodes={scale=0.8},
            /tikz/every even column/.append style={column sep=0.3cm}},
        legend to name=named-stl10,
        tick align=outside,
        tick pos=left,
        width=\figurewidth,
        x grid style={white!69.01960784313725!black},
        xticklabels=\empty,        
        xmin=0, xmax=100,
        y grid style={white!69.01960784313725!black},
        ylabel={distortion (squared distance)},
        ymin=0.0, ymax=0.8,
        ytick={0,0.2,0.4,0.6,0.8,1},
        yticklabels={0.0,0.2,0.4,0.6,0.8,1.0}
    ]
    \addlegendimage{color0, mark=*, mark options={draw=black, fill=color0, thin}}
    \addlegendimage{color9, mark=square*, mark options={draw=black, fill=color9, thin}}
    \addlegendimage{color4, mark=o, mark options={draw=black, fill=color4, thin}}
    \addlegendimage{color6, mark=square,mark options={draw=black, fill=color6, thin}}
    \addlegendimage{color7, mark=triangle, mark options={draw=black, fill=color7, thin}}
    \addlegendimage{color8, mark=oplus*, mark options={draw=black, fill=color8, thin}}
    \addlegendimage{color1, mark=otimes*, mark options={draw=black, fill=color1, thin}}
    \addlegendimage{color5, mark=x, mark options={draw=black, fill=color5, thin}}
    \addlegendimage{color2, mark=diamond*, mark options={draw=black, fill=color2, thin}}
    \addlegendimage{color3, mark=diamond, mark options={draw=black, fill=color3, thin}}

   % mr-rde fw
    \input{./img/stl10/stl10_rate_distortion_fw_multi}
    % mr-rde pgd
    \input{./img/stl10/stl10_rate_distortion_pgd_multi}
    % l-rde
    \input{./img/stl10/stl10_rate_distortion_lagrange}
    % sensitivity
    \input{./img/stl10/stl10_rate_distortion_sensitivity}
    % smoothgrad
    \input{./img/stl10/stl10_rate_distortion_smoothgrad}
    % guided backprop
    \input{./img/stl10/stl10_rate_distortion_guidedbackprop}
    % lrp
    \input{./img/stl10/stl10_rate_distortion_lrp}
    % deep taylor
    \input{./img/stl10/stl10_rate_distortion_deeptaylor}
    % shap
    \input{./img/stl10/stl10_rate_distortion_shap}
    % lime
    \input{./img/stl10/stl10_rate_distortion_lime}

    %\coordinate (spyon1) at (axis cs:18,0.065);
    %\coordinate (spyat1) at (axis cs:99,0.7795);
    %\spy [width=3.5cm, height=1.3cm] on (spyon1) in node[fill=white,anchor=north east] at (spyat1);

    % stl10 rate accuracy
    \nextgroupplot[
        height=.75\figureheight,
        tick align=outside,
        tick pos=left,
        width=\figurewidth,
        xlabel={rate (non-randomized components)},
        x grid style={white!69.01960784313725!black},
        xticklabel={\pgfmathparse{\tick}\pgfmathprintnumber{\pgfmathresult}\%},
        xmin=0, xmax=100,
        y grid style={white!69.01960784313725!black},
        ylabel={accuracy},
        ymin=0.0, ymax=1.0,
        ytick={0,0.2,0.4,0.6,0.8,1},
        yticklabels={0.0,0.2,0.4,0.6,0.8,1.0}
    ]

    % mr-rde fw
    \input{./img/stl10/stl10_rate_accuracy_fw_multi}
    % mr-rde pgd
    \input{./img/stl10/stl10_rate_accuracy_pgd_multi}
    % l-rde
    \input{./img/stl10/stl10_rate_accuracy_lagrange}
    % sensitivity
    \input{./img/stl10/stl10_rate_accuracy_sensitivity}
    % smoothgrad
    \input{./img/stl10/stl10_rate_accuracy_smoothgrad}
    % guided backprop
    \input{./img/stl10/stl10_rate_accuracy_guidedbackprop}
    % lrp
    \input{./img/stl10/stl10_rate_accuracy_lrp}
    % deep taylor
    \input{./img/stl10/stl10_rate_accuracy_deeptaylor}
    % shap
    \input{./img/stl10/stl10_rate_accuracy_shap}
    % lime
    \input{./img/stl10/stl10_rate_accuracy_lime}
        
    %\coordinate (spyon2) at (axis cs:18,0.89);
    %\coordinate (spyat2) at (axis cs:99, 0.025);
    %\spy [width=3.5cm, height=1.3cm] on (spyon2) in node[fill=white,anchor=south east] at (spyat2);
   
\end{groupplot}
\end{tikzpicture}

\begin{center}
  \ref*{named-stl10}
\end{center}
\vspace{-1.5em}

%% file: img/stl10/stl10_examples3_short.tex
\begin{tabular}{c@{\;\;\;}c@{\;\;\;}c@{\;\;\;}c@{\;\;\;}c}
    Image & Sensitivity & LRP-$\alpha$-$\beta$ & SHAP & Guided Backprop \\
    \includegraphics[width=1.25cm, valign=c]{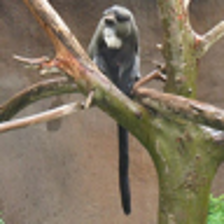} &
    \includegraphics[width=1.25cm, valign=c]{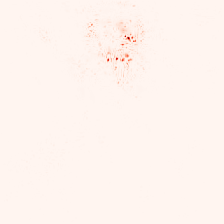} &
    \includegraphics[width=1.25cm, valign=c]{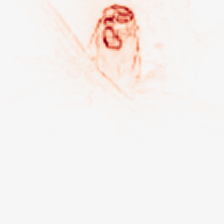} &
    \includegraphics[width=1.25cm, valign=c]{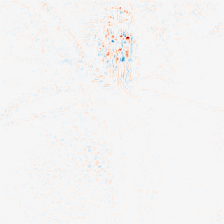} &
    \includegraphics[width=1.25cm, valign=c]{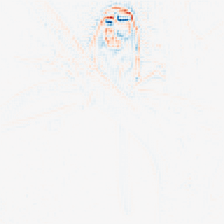} \\
    & SmoothGrad & DeepTaylor & LIME & L-RDE \\
    &
    \includegraphics[width=1.25cm, valign=c]{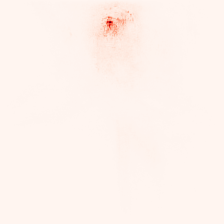} &
    \includegraphics[width=1.25cm, valign=c]{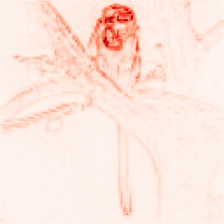} &
    \includegraphics[width=1.25cm, valign=c]{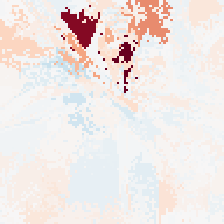} &
    \includegraphics[width=1.25cm, valign=c]{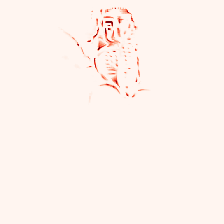}
\end{tabular}
\\[2ex]
\begin{tabular}{lc@{\;\;}c@{\;\;}c@{\;\;}c@{\;\;}c}
    \toprule
    & \multicolumn{4}{c}{RC-RDE} & MR-RDE \\
    & $k=2000$ & $k=4000$ & $k=6000$ & $k=8000$ & \\
    \midrule
    FW &
    \includegraphics[width=1.25cm, valign=c]{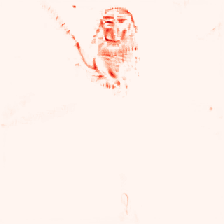} &
    \includegraphics[width=1.25cm, valign=c]{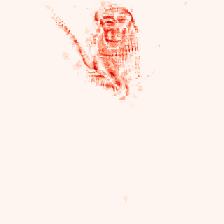} &
    \includegraphics[width=1.25cm, valign=c]{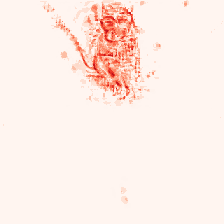} &
    \includegraphics[width=1.25cm, valign=c]{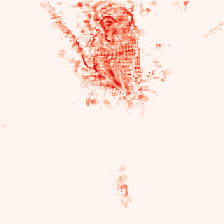} &
    \includegraphics[width=1.25cm, valign=c]{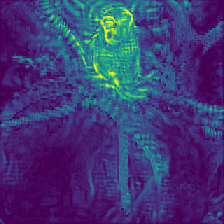} \\

    PGD &
    \includegraphics[width=1.25cm, valign=c]{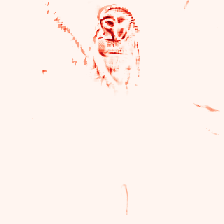} &
    \includegraphics[width=1.25cm, valign=c]{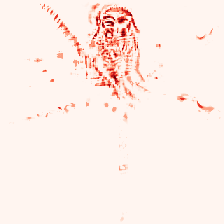} &
    \includegraphics[width=1.25cm, valign=c]{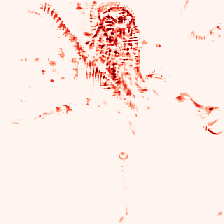} &
    \includegraphics[width=1.25cm, valign=c]{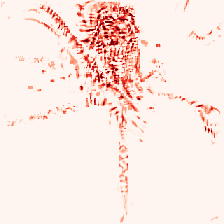} &
    \includegraphics[width=1.25cm, valign=c]{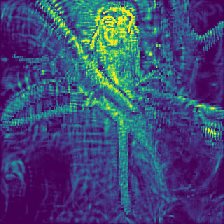} \\
    \bottomrule
\end{tabular}

%% file: img/stl10/stl10_examples4_short.tex
\begin{tabular}{c@{\;\;\;}c@{\;\;\;}c@{\;\;\;}c@{\;\;\;}c}
    Image & Sensitivity & LRP-$\alpha$-$\beta$ & SHAP & Guided Backprop \\
    \includegraphics[width=1.25cm, valign=c]{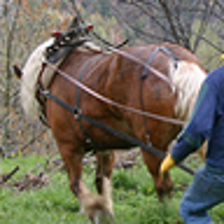} &
    \includegraphics[width=1.25cm, valign=c]{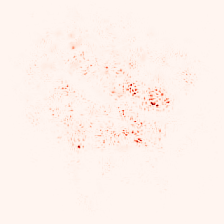} &
    \includegraphics[width=1.25cm, valign=c]{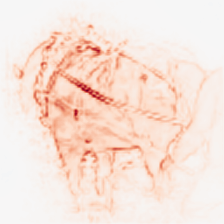} &
    \includegraphics[width=1.25cm, valign=c]{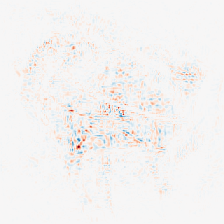} &
    \includegraphics[width=1.25cm, valign=c]{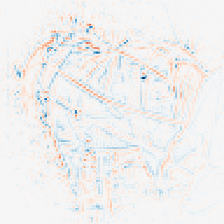} \\
    & SmoothGrad & DeepTaylor & LIME & L-RDE \\
    &
    \includegraphics[width=1.25cm, valign=c]{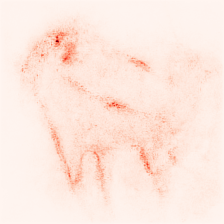} &
    \includegraphics[width=1.25cm, valign=c]{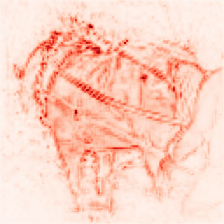} &
    \includegraphics[width=1.25cm, valign=c]{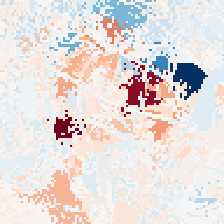} &
    \includegraphics[width=1.25cm, valign=c]{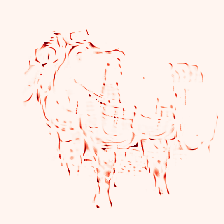}
\end{tabular}
\\[2ex]
\begin{tabular}{lc@{\;\;}c@{\;\;}c@{\;\;}c@{\;\;}c}
    \toprule
    & \multicolumn{4}{c}{RC-RDE} & MR-RDE \\
    & $k=2000$ & $k=4000$ & $k=6000$ & $k=8000$ & \\
    \midrule
    FW &
    \includegraphics[width=1.25cm, valign=c]{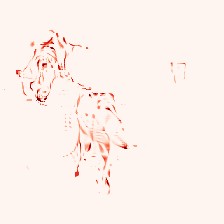} &
    \includegraphics[width=1.25cm, valign=c]{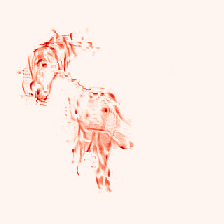} &
    \includegraphics[width=1.25cm, valign=c]{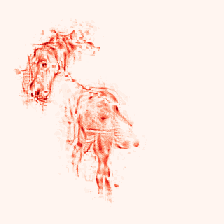} &
    \includegraphics[width=1.25cm, valign=c]{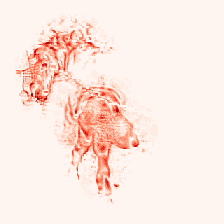} &
    \includegraphics[width=1.25cm, valign=c]{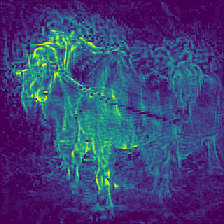} \\

    PGD &
    \includegraphics[width=1.25cm, valign=c]{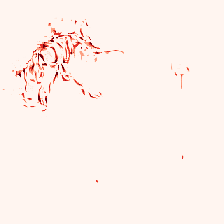} &
    \includegraphics[width=1.25cm, valign=c]{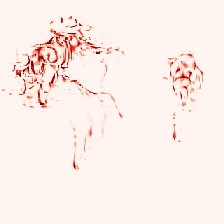} &
    \includegraphics[width=1.25cm, valign=c]{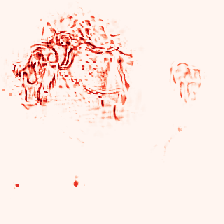} &
    \includegraphics[width=1.25cm, valign=c]{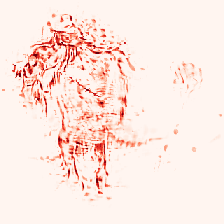} &
    \includegraphics[width=1.25cm, valign=c]{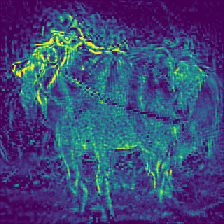} \\
    \bottomrule
\end{tabular}

%% file: img/mnist/mnist_examples_additional_fw.tex
%\begin{tabular}{c@{\;\;\;}c@{\;\;\;}c@{\;\;\;}c@{\;\;\;}c}
%   Image & Sensitivity & LRP-$\alpha$-$\beta$ & SHAP & Guided Backprop \\
%   \includegraphics[width=1.25cm, valign=c]{./img/mnist/mnist_testing_6425_input.png} &
%   \includegraphics[width=1.25cm, valign=c]{./img/mnist/mnist_testing_6425_sensitivity.png} &
%   \includegraphics[width=1.25cm, valign=c]{./img/mnist/mnist_testing_6425_lrp_ab.png} &
%    \includegraphics[width=1.25cm, valign=c]{./img/mnist/mnist_testing_6425_shap.png} &
%    \includegraphics[width=1.25cm, valign=c]{./img/mnist/mnist_testing_6425_guidedbackprop.png} \\
%    & SmoothGrad & DeepTaylor & LIME & L-RDE \\
%    &
%    \includegraphics[width=1.25cm, valign=c]{./img/mnist/mnist_testing_6425_smoothgrad.png} &
%    \includegraphics[width=1.25cm, valign=c]{./img/mnist/mnist_testing_6425_deeptaylor.png} &
%    \includegraphics[width=1.25cm, valign=c]{./img/mnist/mnist_testing_6425_lime.png} &
%    \includegraphics[width=1.25cm, valign=c]{./img/mnist/mnist_testing_6425_l_rde.png}
%\end{tabular}
%\\[2ex]
\begin{tabular}{@{}l@{\;}c@{\;}c@{\;}c@{\;}c@{\;}c@{\;}c@{}}
    \toprule
    & \multicolumn{4}{c}{RC-RDE} & MR-RDE & Ord-RDE \\
    & $k=50$ & $k=100$ & $k=150$ & $k=200$ & & \\
    \midrule
    %FW &
    %\includegraphics[width=1.25cm, valign=c]{./img/mnist/mnist_testing_6425_fw_rc_rde_50.png} &
    %\includegraphics[width=1.25cm, valign=c]{./img/mnist/mnist_testing_6425_fw_rc_rde_100.png} &
    %\includegraphics[width=1.25cm, valign=c]{./img/mnist/mnist_testing_6425_fw_rc_rde_150.png} &
    %\includegraphics[width=1.25cm, valign=c]{./img/mnist/mnist_testing_6425_fw_rc_rde_200.png} &
    %\includegraphics[width=1.25cm, valign=c]{./img/mnist/mnist_testing_6425_fw_rc_rde_multi_cmap.png} &
    %\includegraphics[width=1.25cm, valign=c]{./img/mnist/mnist_testing_6425_fw_ord_rde_cmap.png} \\

    AFW &
    \includegraphics[width=1.25cm, valign=c]{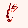} &
    \includegraphics[width=1.25cm, valign=c]{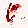} &
    \includegraphics[width=1.25cm, valign=c]{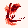} &
    \includegraphics[width=1.25cm, valign=c]{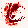} &
    \includegraphics[width=1.25cm, valign=c]{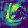} &
    \includegraphics[width=1.25cm, valign=c]{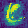} \\

    LCG &
    \includegraphics[width=1.25cm, valign=c]{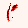} &
    \includegraphics[width=1.25cm, valign=c]{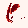} &
    \includegraphics[width=1.25cm, valign=c]{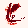} &
    \includegraphics[width=1.25cm, valign=c]{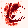} &
    \includegraphics[width=1.25cm, valign=c]{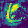} &
    \includegraphics[width=1.25cm, valign=c]{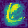} \\

    LAFW &
    \includegraphics[width=1.25cm, valign=c]{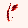} &
    \includegraphics[width=1.25cm, valign=c]{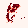} &
    \includegraphics[width=1.25cm, valign=c]{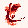} &
    \includegraphics[width=1.25cm, valign=c]{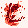} &
    \includegraphics[width=1.25cm, valign=c]{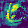} &
    \includegraphics[width=1.25cm, valign=c]{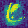} \\

    %SFW &
    %\includegraphics[width=1.25cm, valign=c]{./img/mnist/mnist_testing_6425_sfw_rc_rde_50.png} &
    %\includegraphics[width=1.25cm, valign=c]{./img/mnist/mnist_testing_6425_sfw_rc_rde_100.png} &
    %\includegraphics[width=1.25cm, valign=c]{./img/mnist/mnist_testing_6425_sfw_rc_rde_150.png} &
    %\includegraphics[width=1.25cm, valign=c]{./img/mnist/mnist_testing_6425_sfw_rc_rde_200.png} &
    %\includegraphics[width=1.25cm, valign=c]{./img/mnist/mnist_testing_6425_sfw_rc_rde_multi_cmap.png} &
    %\includegraphics[width=1.25cm, valign=c]{./img/mnist/mnist_testing_6425_sfw_ord_rde_cmap.png} \\

    %PGD &
    %\includegraphics[width=1.25cm, valign=c]{./img/mnist/mnist_testing_6425_pgd_rc_rde_50.png} &
    %\includegraphics[width=1.25cm, valign=c]{./img/mnist/mnist_testing_6425_pgd_rc_rde_100.png} &
    %\includegraphics[width=1.25cm, valign=c]{./img/mnist/mnist_testing_6425_pgd_rc_rde_150.png} &
    %\includegraphics[width=1.25cm, valign=c]{./img/mnist/mnist_testing_6425_pgd_rc_rde_200.png} &
    %\includegraphics[width=1.25cm, valign=c]{./img/mnist/mnist_testing_6425_pgd_rc_rde_multi_cmap.png} &
    %N/A \\
    \bottomrule
\end{tabular}

%% file: img/mnist/mnist_eval.tex
\begin{tabular}{c}
\begin{tikzpicture}
\begin{groupplot}[group style={group size=1 by 2, vertical sep=4ex}]

    % mnist runtimes
    \nextgroupplot[
        height=.6\figureheight,
        legend cell align={left},
        legend entries={
            {FW (RC-RDE)},
            {AFW (RC-RDE)},
            {LCG (RC-RDE)},
            {LAFW (RC-RDE)}
        },
        legend columns=2,
        legend style={anchor=north west, draw=none, line width=1.5, nodes={scale=0.8},
            /tikz/every even column/.append style={column sep=0.3cm}},
        legend to name=named-eval,
        tick align=outside,
        tick pos=left,
        width=\figurewidth,
        x grid style={white!69.01960784313725!black},
        xticklabels=\empty,
        xmin=50, xmax=400,
        y grid style={white!69.01960784313725!black},
        ylabel={runtime (seconds)},
        ymin=0.0, ymax=20.0,
        ytick={0,5,10,15,20},
        yticklabels={0,5,10,15,20}
    ]
    \addlegendimage{color0, mark=*, mark options={draw=black, fill=color0, thin}}
    \addlegendimage{color3, mark=diamond, mark options={draw=black, fill=color3, thin}}
    \addlegendimage{color8, mark=oplus*, mark options={draw=black, fill=color8, thin}}
    \addlegendimage{color1, mark=otimes*, mark options={draw=black, fill=color1, thin}}

    \path [fill=color0, fill opacity=0.25]
    (axis cs:50,14.740214119743)
    --(axis cs:50,13.452892920257)
    --(axis cs:100,13.7250201439355)
    --(axis cs:150,13.8372700008768)
    --(axis cs:200,13.947943075732)
    --(axis cs:250,14.0688219394082)
    --(axis cs:300,14.1588741384725)
    --(axis cs:350,14.314081028134)
    --(axis cs:400,14.3145172464922)
    --(axis cs:400,15.6217739935078)
    --(axis cs:400,15.6217739935078)
    --(axis cs:350,15.603016251866)
    --(axis cs:300,15.4435227415275)
    --(axis cs:250,15.1375629005918)
    --(axis cs:200,15.230678884268)
    --(axis cs:150,15.1143262791232)
    --(axis cs:100,15.0136930960645)
    --(axis cs:50,14.740214119743)
    --cycle;

    \path [fill=color3, fill opacity=0.25]
    (axis cs:50,7.51600507382122)
    --(axis cs:50,1.86470832617878)
    --(axis cs:100,4.50279339524534)
    --(axis cs:150,9.96680117512372)
    --(axis cs:200,13.6075959252207)
    --(axis cs:250,15.7779929018468)
    --(axis cs:300,15.9688514565692)
    --(axis cs:350,16.4448318880489)
    --(axis cs:400,17.5233554253286)
    --(axis cs:400,19.1203766946714)
    --(axis cs:400,19.1203766946714)
    --(axis cs:350,19.0695657519511)
    --(axis cs:300,19.3632159434308)
    --(axis cs:250,18.9719288581532)
    --(axis cs:200,18.3567766747793)
    --(axis cs:150,16.9488765448763)
    --(axis cs:100,11.8496186447547)
    --(axis cs:50,7.51600507382122)
    --cycle;

    \path [fill=color8, fill opacity=0.25]
    (axis cs:50,14.9601422921923)
    --(axis cs:50,13.7358210678077)
    --(axis cs:100,13.5409385995026)
    --(axis cs:150,13.6408229214705)
    --(axis cs:200,13.60329342534)
    --(axis cs:250,13.9141641860459)
    --(axis cs:300,13.4885787534918)
    --(axis cs:350,13.5399857176099)
    --(axis cs:400,13.6353740481851)
    --(axis cs:400,14.8692662718149)
    --(axis cs:400,14.8692662718149)
    --(axis cs:350,14.7669035623901)
    --(axis cs:300,14.7508282065082)
    --(axis cs:250,15.2324438139541)
    --(axis cs:200,14.88377789466)
    --(axis cs:150,14.8615390385295)
    --(axis cs:100,14.7929843204974)
    --(axis cs:50,14.9601422921923)
    --cycle;

    \path [fill=color1, fill opacity=0.25]
    (axis cs:50,17.7665900680442)
    --(axis cs:50,13.2616506919558)
    --(axis cs:100,13.9675767075409)
    --(axis cs:150,15.584163499653)
    --(axis cs:200,16.2663952905675)
    --(axis cs:250,16.2749971489637)
    --(axis cs:300,16.6320663300059)
    --(axis cs:350,16.149443494908)
    --(axis cs:400,16.4661992914082)
    --(axis cs:400,17.8278917885918)
    --(axis cs:400,17.8278917885918)
    --(axis cs:350,17.509237185092)
    --(axis cs:300,17.9640371099941)
    --(axis cs:250,17.6469884510363)
    --(axis cs:200,17.5313751094325)
    --(axis cs:150,18.133690140347)
    --(axis cs:100,17.9080520524591)
    --(axis cs:50,17.7665900680442)
    --cycle;

    \addplot [thick, color0, mark=*, mark size=1.5pt, mark options={draw=black, fill=color0, thin}]
    table {%
    50 14.09655352
    100 14.36935662
    150 14.47579814
    200 14.58931098
    250 14.60319242
    300 14.80119844
    350 14.95854864
    400 14.96814562
    };

    \addplot [thick, color3, mark=diamond, mark size=1.5pt, mark options={draw=black, fill=color3, thin}]
    table {%
    50 4.6903567
    100 8.17620602
    150 13.45783886
    200 15.9821863
    250 17.37496088
    300 17.6660337
    350 17.75719882
    400 18.32186606
    };

    \addplot [thick, color8, mark=oplus*, mark size=1.5pt, mark options={draw=black, fill=color8, thin}]
    table {%
    50 14.34798168
    100 14.16696146
    150 14.25118098
    200 14.24353566
    250 14.573304
    300 14.11970348
    350 14.15344464
    400 14.25232016
    };

    \addplot [thick, color1, mark=otimes*, mark size=1.5pt, mark options={draw=black, fill=color1, thin}]
    table {%
    50 15.51412038
    100 15.93781438
    150 16.85892682
    200 16.8988852
    250 16.9609928
    300 17.29805172
    350 16.82934034
    400 17.14704554
    };

    % mnist iterations
    \nextgroupplot[
        height=.6\figureheight,
        tick align=outside,
        tick pos=left,
        width=\figurewidth,
        xlabel={rate constraint $k$},
        x grid style={white!69.01960784313725!black},
        xtick={50,100,150,200,250,300,350,400},
        xticklabels={50,100,150,200,250,300,350,400},
        xmin=50, xmax=400,
        y grid style={white!69.01960784313725!black},
        ylabel={iterations},
        ymin=0.0, ymax=2500,
        ytick={0,500,1000,1500,2000,2500},
        yticklabels={0,500,1000,1500,2000,2500}
    ]

    \path [fill=color0, fill opacity=0.25]
    (axis cs:50,2000)
    --(axis cs:50,2000)
    --(axis cs:100,2000)
    --(axis cs:150,2000)
    --(axis cs:200,2000)
    --(axis cs:250,2000)
    --(axis cs:300,2000)
    --(axis cs:350,2000)
    --(axis cs:400,2000)
    --(axis cs:400,2000)
    --(axis cs:400,2000)
    --(axis cs:350,2000)
    --(axis cs:300,2000)
    --(axis cs:250,2000)
    --(axis cs:200,2000)
    --(axis cs:150,2000)
    --(axis cs:100,2000)
    --(axis cs:50,2000)
    --cycle;

    \path [fill=color3, fill opacity=0.25]
    (axis cs:50,636.91860617389)
    --(axis cs:50,-52.1186061738902)
    --(axis cs:100,226.996207412642)
    --(axis cs:150,961.56504538986)
    --(axis cs:200,1515.64642982204)
    --(axis cs:250,1736.34399678481)
    --(axis cs:300,1750.61867194174)
    --(axis cs:350,1832.69)
    --(axis cs:400,2000)
    --(axis cs:400,2000)
    --(axis cs:400,2000)
    --(axis cs:350,2092.95)
    --(axis cs:300,2104.82132805826)
    --(axis cs:250,2110.81600321519)
    --(axis cs:200,2096.99357017796)
    --(axis cs:150,1860.99495461014)
    --(axis cs:100,1163.56379258736)
    --(axis cs:50,636.91860617389)
    --cycle;

    \path [fill=color8, fill opacity=0.25]
    (axis cs:50,2000)
    --(axis cs:50,2000)
    --(axis cs:100,2000)
    --(axis cs:150,2000)
    --(axis cs:200,2000)
    --(axis cs:250,2000)
    --(axis cs:300,2000)
    --(axis cs:350,2000)
    --(axis cs:400,2000)
    --(axis cs:400,2000)
    --(axis cs:400,2000)
    --(axis cs:350,2000)
    --(axis cs:300,2000)
    --(axis cs:250,2000)
    --(axis cs:200,2000)
    --(axis cs:150,2000)
    --(axis cs:100,2000)
    --(axis cs:50,2000)
    --cycle;

    \path [fill=color1, fill opacity=0.25]
    (axis cs:50,1947.79547155742)
    --(axis cs:50,1104.68452844258)
    --(axis cs:100,1441.86032566052)
    --(axis cs:150,1823.24)
    --(axis cs:200,1729.62883119268)
    --(axis cs:250,1830.98)
    --(axis cs:300,1833.68)
    --(axis cs:350,2000)
    --(axis cs:400,2000)
    --(axis cs:400,2000)
    --(axis cs:400,2000)
    --(axis cs:350,2000)
    --(axis cs:300,2092.4)
    --(axis cs:250,2093.9)
    --(axis cs:200,2113.61116880732)
    --(axis cs:150,2098.2)
    --(axis cs:100,2083.93967433948)
    --(axis cs:50,1947.79547155742)
    --cycle;

    \addplot [thick, color0, mark=*, mark size=1.5pt, mark options={draw=black, fill=color0, thin}]
    table {%
    50 2000
    100 2000
    150 2000
    200 2000
    250 2000
    300 2000
    350 2000
    400 2000
    };

    \addplot [thick, color3, mark=diamond, mark size=1.5pt, mark options={draw=black, fill=color3, thin}]
    table {%
    50 292.4
    100 695.28
    150 1411.28
    200 1806.32
    250 1923.58
    300 1927.72
    350 1962.82
    400 2000
    };

    \addplot [thick, color8, mark=oplus*, mark size=1.5pt, mark options={draw=black, fill=color8, thin}]
    table {%
    50 2000
    100 2000
    150 2000
    200 2000
    250 2000
    300 2000
    350 2000
    400 2000
    };

    \addplot [thick, color1, mark=otimes*, mark size=1.5pt, mark options={draw=black, fill=color1, thin}]
    table {%
    50 1526.24
    100 1762.9
    150 1960.72
    200 1921.62
    250 1962.44
    300 1963.04
    350 2000
    400 2000
    };

\end{groupplot}
\end{tikzpicture}
\\
\ref*{named-eval}
\end{tabular}

%% file: img/mnist/mnist_examples_iterations.tex
\begin{tabular}{@{}l@{\;}c@{\;}c@{\;}c@{\;}c@{}}
    \toprule
    & $T=50$ & $T=100$ & $T=150$ & $T=200$ \\
    \midrule
    FW &
    \includegraphics[width=1.25cm, valign=c]{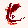} &
    \includegraphics[width=1.25cm, valign=c]{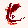} &
    \includegraphics[width=1.25cm, valign=c]{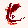} &
    \includegraphics[width=1.25cm, valign=c]{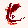} \\

    AFW &
    \includegraphics[width=1.25cm, valign=c]{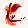} &
    \includegraphics[width=1.25cm, valign=c]{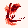} &
    \includegraphics[width=1.25cm, valign=c]{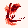} &
    \includegraphics[width=1.25cm, valign=c]{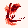} \\

    LCG &
    \includegraphics[width=1.25cm, valign=c]{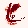} &
    \includegraphics[width=1.25cm, valign=c]{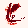} &
    \includegraphics[width=1.25cm, valign=c]{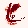} &
    \includegraphics[width=1.25cm, valign=c]{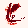} \\

    LAFW &
    \includegraphics[width=1.25cm, valign=c]{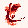} &
    \includegraphics[width=1.25cm, valign=c]{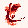} &
    \includegraphics[width=1.25cm, valign=c]{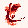} &
    \includegraphics[width=1.25cm, valign=c]{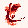} \\

    \bottomrule
\end{tabular}

%% file: img/mnist/mnist_curves_extra.tex
\begin{tikzpicture}[spy using outlines={rectangle, gray, magnification=3, connect spies}]
\begin{groupplot}[group style={group size=2 by 1, horizontal sep=6em}]

    % mnist rate distortion
    \nextgroupplot[
        height=.75\figureheight,
        legend cell align={left},
        legend entries={
            {MR-RDE (FW)},
            {MR-RDE (AFW)},
            {MR-RDE (PGD)},
            {MR-RDE (LCG)},
            {MR-RDE (LAFW)},
            {Ord-RDE (SFW)}
        },
        legend columns=3,
        legend style={anchor=north west, draw=none, line width=1.5, nodes={scale=0.8},
            /tikz/every even column/.append style={column sep=0.3cm}},
        legend to name=named-mnist-extra,
        tick align=outside,
        tick pos=left,
        width=\figurewidth,
        x grid style={white!69.01960784313725!black},
        xticklabels=\empty,
        xmin=0, xmax=100,
        y grid style={white!69.01960784313725!black},
        ylabel={distortion (squared distance)},
        ymin=0.0, ymax=1.0,
        ytick={0,0.2,0.4,0.6,0.8,1},
        yticklabels={0.0,0.2,0.4,0.6,0.8,1.0}
    ]
    \addlegendimage{color0, mark=*, mark options={draw=black, fill=color0, thin}}
    \addlegendimage{color3, mark=diamond, mark options={draw=black, fill=color3, thin}}
    \addlegendimage{color9, mark=square*, mark options={draw=black, fill=color9, thin}}
    \addlegendimage{color8, mark=oplus*, mark options={draw=black, fill=color8, thin}}
    \addlegendimage{color1, mark=otimes*, mark options={draw=black, fill=color1, thin}}
    \addlegendimage{color10, mark=triangle*, mark options={draw=black, fill=color10, thin}}

    % rc rde fw
    \input{./img/mnist/mnist_rate_distortion_fw_multi}
    % rc rde afw
    \input{./img/mnist/mnist_rate_distortion_afw_multi}
    % rc rde pgd
    \input{./img/mnist/mnist_rate_distortion_pgd_multi}
    % rc rde lcg
    \input{./img/mnist/mnist_rate_distortion_lcg_multi}
    % rc rde lafw
    \input{./img/mnist/mnist_rate_distortion_lafw_multi}
    % ord rde sfw
    \input{./img/mnist/mnist_rate_distortion_sfw_order_var_a}
        
    %\addplot [gray, dashed, forget plot]
    %table [row sep=\\]{%
    %0	0 \\
    %100	0 \\
    %};
    %\addplot [gray, dashed, forget plot]
    %table [row sep=\\]{%
    %0	0.915433638794865 \\
    %100	0.915433638794865 \\
    %};
    \coordinate (spyon1) at (axis cs:60,0.075);
    \coordinate (spyat1) at (axis cs:99,0.975);
    \spy [width=3.5cm, height=1.3cm] on (spyon1) in node[fill=white,anchor=north east] at (spyat1);

    % mnist rate accuracy
    \nextgroupplot[
        height=.75\figureheight,
        tick align=outside,
        tick pos=left,
        width=\figurewidth,
        xlabel={rate (non-randomized components)},
        x grid style={white!69.01960784313725!black},
        xticklabel={\pgfmathparse{\tick}\pgfmathprintnumber{\pgfmathresult}\%},
        xmin=0, xmax=100,
        y grid style={white!69.01960784313725!black},
        ylabel={accuracy},
        ymin=0.0, ymax=1.0,
        ytick={0,0.2,0.4,0.6,0.8,1},
        yticklabels={0.0,0.2,0.4,0.6,0.8,1.0}
    ]

    % rc rde fw
    \input{./img/mnist/mnist_rate_accuracy_fw_multi}
    % rc rde afw
    \input{./img/mnist/mnist_rate_accuracy_afw_multi}
    % rc rde pgd
    \input{./img/mnist/mnist_rate_accuracy_pgd_multi}
    % rc rde lcg
    \input{./img/mnist/mnist_rate_accuracy_lcg_multi}
    % rc rde lafw
    \input{./img/mnist/mnist_rate_accuracy_lafw_multi}
    % ord rde sfw
    \input{./img/mnist/mnist_rate_accuracy_sfw_order_var_a}
        
    %\addplot [gray, dashed, forget plot]
    %table [row sep=\\]{%
    %0	1 \\
    %100	1 \\
    %};
    %\addplot [gray, dashed, forget plot]
    %table [row sep=\\]{%
    %0	0.0800130208333333 \\
    %100	0.0800130208333333 \\
    %};
    \coordinate (spyon2) at (axis cs:60,0.92);
    \coordinate (spyat2) at (axis cs:99,0.025);
    \spy [width=3.5cm, height=1.3cm] on (spyon2) in node[fill=white,anchor=south east] at (spyat2);
   
\end{groupplot}
\end{tikzpicture}

\begin{center}
  \ref*{named-mnist-extra}
\end{center}
\vspace{-1.5em}

%% file: img/stl10/stl10_curves_extra.tex
\begin{tikzpicture}[spy using outlines={rectangle, gray, magnification=4, connect spies}]
\begin{groupplot}[group style={group size=2 by 1, horizontal sep=6em}]

    % stl10 rate distortion
    \nextgroupplot[
        height=.75\figureheight,
        legend cell align={left},
        legend entries={
            {MR-RDE (FW)},
            {MR-RDE (AFW)},
            {MR-RDE (PGD)},
            {MR-RDE (LCG)},
            {MR-RDE (LAFW)}
        },
        legend columns=3,
        legend style={anchor=north west, draw=none, line width=1.5, nodes={scale=0.8},
            /tikz/every even column/.append style={column sep=0.3cm}},
        legend to name=named-stl10-extra,
        tick align=outside,
        tick pos=left,
        width=\figurewidth,
        x grid style={white!69.01960784313725!black},
        xticklabels=\empty,        
        xmin=0, xmax=100,
        y grid style={white!69.01960784313725!black},
        ylabel={distortion (squared distance)},
        ymin=0.0, ymax=0.8,
        ytick={0,0.2,0.4,0.6,0.8,1},
        yticklabels={0.0,0.2,0.4,0.6,0.8,1.0}
    ]
    \addlegendimage{color0, mark=*, mark options={draw=black, fill=color0, thin}}
    \addlegendimage{color3, mark=diamond, mark options={draw=black, fill=color3, thin}}
    \addlegendimage{color9, mark=square*, mark options={draw=black, fill=color9, thin}}
    \addlegendimage{color8, mark=oplus*, mark options={draw=black, fill=color8, thin}}
    \addlegendimage{color1, mark=otimes*, mark options={draw=black, fill=color1, thin}}

    % rc rde fw
    \input{./img/stl10/stl10_rate_distortion_fw_multi}
    % rc rde afw
    \input{./img/stl10/stl10_rate_distortion_afw_multi}
    % rc rde pgd
    \input{./img/stl10/stl10_rate_distortion_pgd_multi}
    % rc rde lcg
    \input{./img/stl10/stl10_rate_distortion_lcg_multi}
    % rc rde lafw
    \input{./img/stl10/stl10_rate_distortion_lafw_multi}
        
    %\addplot [gray, dashed, forget plot]
    %table [row sep=\\]{%
    %0	0 \\
    %100	0 \\
    %};
    %\addplot [gray, dashed, forget plot]
    %table [row sep=\\]{%
    %0	0.718839381047421 \\
    %100	0.718839381047421 \\
    %};
    \coordinate (spyon1) at (axis cs:18,0.065);
    \coordinate (spyat1) at (axis cs:99,0.7795);
    \spy [width=3.5cm, height=1.3cm] on (spyon1) in node[fill=white,anchor=north east] at (spyat1);

    % stl10 rate accuracy
    \nextgroupplot[
        height=.75\figureheight,
        tick align=outside,
        tick pos=left,
        width=\figurewidth,
        xlabel={rate (non-randomized components)},
        x grid style={white!69.01960784313725!black},
        xticklabel={\pgfmathparse{\tick}\pgfmathprintnumber{\pgfmathresult}\%},
        xmin=0, xmax=100,
        y grid style={white!69.01960784313725!black},
        ylabel={accuracy},
        ymin=0.0, ymax=1.0,
        ytick={0,0.2,0.4,0.6,0.8,1},
        yticklabels={0.0,0.2,0.4,0.6,0.8,1.0}
    ]

    % rc rde fw
    \input{./img/stl10/stl10_rate_accuracy_fw_multi}
    % rc rde afw
    \input{./img/stl10/stl10_rate_accuracy_afw_multi}
    % rc rde pgd
    \input{./img/stl10/stl10_rate_accuracy_pgd_multi}
    % rc rde lcg
    \input{./img/stl10/stl10_rate_accuracy_lcg_multi}
    % rc rde lafw
    \input{./img/stl10/stl10_rate_accuracy_lafw_multi}
        
    %\addplot [gray, dashed, forget plot]
    %table [row sep=\\]{%
    %0	1 \\
    %100	1 \\
    %};
    %\addplot [gray, dashed, forget plot]
    %table [row sep=\\]{%
    %0	0.0765711805555555 \\
    %100	0.0765711805555555 \\
    %};
    \coordinate (spyon2) at (axis cs:18,0.89);
    \coordinate (spyat2) at (axis cs:99, 0.025);
    \spy [width=3.5cm, height=1.3cm] on (spyon2) in node[fill=white,anchor=south east] at (spyat2);
   
\end{groupplot}
\end{tikzpicture}

\begin{center}
  \ref*{named-stl10-extra}
\end{center}
\vspace{-1.5em}

%% file: img/mnist/mnist_curves_order.tex
\begin{tikzpicture}[spy using outlines={rectangle, gray, magnification=3, connect spies}]
\begin{groupplot}[group style={group size=2 by 1, horizontal sep=6em}]

    % mnist rate distortion
    \nextgroupplot[
        height=.75\figureheight,
        legend cell align={left},
        legend entries={
            {FW (Ord-RDE)},
            {AFW (Ord-RDE)},
            {SFW (Variant A) (Ord-RDE)},
            {LCG (Ord-RDE)},
            {LAFW (Ord-RDE)}
        },
        legend columns=3,
        legend style={anchor=north west, draw=none, line width=1.5,
            /tikz/every even column/.append style={column sep=0.3cm}},
        legend to name=named-mnist-order,
        tick align=outside,
        tick pos=left,
        width=\figurewidth,
        xlabel={rate (non-randomized components)},
        x grid style={white!69.01960784313725!black},
        xticklabel={\pgfmathparse{\tick}\pgfmathprintnumber{\pgfmathresult}\%},
        xmin=0, xmax=100,
        y grid style={white!69.01960784313725!black},
        ylabel={distortion (squared distance)},
        ymin=0.0, ymax=1.0,
        ytick={0,0.2,0.4,0.6,0.8,1},
        yticklabels={0.0,0.2,0.4,0.6,0.8,1.0}
    ]
    \addlegendimage{color0, mark=*, mark options={draw=black, fill=color0, thin}}
    \addlegendimage{color3, mark=diamond, mark options={draw=black, fill=color3, thin}}
    \addlegendimage{color10, mark=triangle*, mark options={draw=black, fill=color10, thin}}
    \addlegendimage{color8, mark=oplus*, mark options={draw=black, fill=color8, thin}}
    \addlegendimage{color1, mark=otimes*, mark options={draw=black, fill=color1, thin}}

    % ord rde fw
    \input{./img/mnist/mnist_rate_distortion_fw_order}
    % ord rde afw
    \input{./img/mnist/mnist_rate_distortion_afw_order}
    % ord rde sfw
    \input{./img/mnist/mnist_rate_distortion_sfw_order_var_a}
    % ord rde lcg
    \input{./img/mnist/mnist_rate_distortion_lcg_order}
    % ord rde lafw
    \input{./img/mnist/mnist_rate_distortion_lafw_order}

    \coordinate (spyon1) at (axis cs:60,0.095);
    \coordinate (spyat1) at (axis cs:99,0.975);
    \spy [width=3.5cm, height=1.3cm] on (spyon1) in node[fill=white,anchor=north east] at (spyat1);

    % mnist rate accuracy
    \nextgroupplot[
        height=.75\figureheight,
        tick align=outside,
        tick pos=left,
        width=\figurewidth,
        xlabel={rate (non-randomized components)},
        x grid style={white!69.01960784313725!black},
        xticklabel={\pgfmathparse{\tick}\pgfmathprintnumber{\pgfmathresult}\%},
        xmin=0, xmax=100,
        y grid style={white!69.01960784313725!black},
        ylabel={accuracy},
        ymin=0.0, ymax=1.0,
        ytick={0,0.2,0.4,0.6,0.8,1},
        yticklabels={0.0,0.2,0.4,0.6,0.8,1.0}
    ]

    % ord rde fw
    \input{./img/mnist/mnist_rate_accuracy_fw_order}
    % ord rde afw
    \input{./img/mnist/mnist_rate_accuracy_afw_order}
    % ord rde sfw
    \input{./img/mnist/mnist_rate_accuracy_sfw_order_var_a}
    % ord rde lcg
    \input{./img/mnist/mnist_rate_accuracy_lcg_order}
    % ord rde lafw
    \input{./img/mnist/mnist_rate_accuracy_lafw_order}
    
    \coordinate (spyon2) at (axis cs:60,0.9);
    \coordinate (spyat2) at (axis cs:99,0.025);
    \spy [width=3.5cm, height=1.3cm] on (spyon2) in node[fill=white,anchor=south east] at (spyat2);
   
\end{groupplot}
\end{tikzpicture}

\begin{center}
  \ref*{named-mnist-order}
\end{center}
\vspace{-1.5em}

%% file: img/mnist/mnist_curves_sfw.tex
\begin{tikzpicture}[spy using outlines={rectangle, gray, magnification=3, connect spies}]
\begin{groupplot}[group style={group size=2 by 1, horizontal sep=6em}]

    % mnist rate distortion
    \nextgroupplot[
        height=.75\figureheight,
        legend cell align={left},
        legend entries={
            {Variant A},
            {Variant B},
            {Variant C},
            {Variant D},
            {Variant E},
            {Variant F}
        },
        legend columns=3,
        legend style={anchor=north west, draw=none, line width=1.5,
            /tikz/every even column/.append style={column sep=0.3cm}},
        legend to name=named-mnist-sfw,
        tick align=outside,
        tick pos=left,
        width=\figurewidth,
        xlabel={rate (non-randomized components)},
        x grid style={white!69.01960784313725!black},
        xticklabel={\pgfmathparse{\tick}\pgfmathprintnumber{\pgfmathresult}\%},
        xmin=0, xmax=100,
        y grid style={white!69.01960784313725!black},
        ylabel={distortion (squared distance)},
        ymin=0.0, ymax=1.0,
        ytick={0,0.2,0.4,0.6,0.8,1},
        yticklabels={0.0,0.2,0.4,0.6,0.8,1.0}
    ]
    \addlegendimage{color10, mark=triangle*, mark options={draw=black, fill=color10, thin}}
    \addlegendimage{color0, mark=*, mark options={draw=black, fill=color0, thin}}
    \addlegendimage{color1, mark=otimes*, mark options={draw=black, fill=color1, thin}}
    \addlegendimage{color3, mark=diamond, mark options={draw=black, fill=color3, thin}}
    \addlegendimage{color4, mark=o, mark options={draw=black, fill=color4, thin}}
    \addlegendimage{color5, mark=x, mark options={draw=black, fill=color5, thin}}

    % ord rde sfw, variant a
    \input{./img/mnist/mnist_rate_distortion_sfw_order_var_a}
    % ord rde sfw, variant b
    \input{./img/mnist/mnist_rate_distortion_sfw_order_var_b}
    % ord rde sfw, variant c
    \input{./img/mnist/mnist_rate_distortion_sfw_order_var_c}
    % ord rde sfw, variant d
    \input{./img/mnist/mnist_rate_distortion_sfw_order_var_d}
    % ord rde sfw, variant e
    \input{./img/mnist/mnist_rate_distortion_sfw_order_var_e}
    % ord rde sfw, variant f
    \input{./img/mnist/mnist_rate_distortion_sfw_order_var_f}

    \coordinate (spyon1) at (axis cs:60,0.075);
    \coordinate (spyat1) at (axis cs:99,0.975);
    \spy [width=3.5cm, height=1.3cm] on (spyon1) in node[fill=white,anchor=north east] at (spyat1);

    % mnist rate accuracy
    \nextgroupplot[
        height=.75\figureheight,
        tick align=outside,
        tick pos=left,
        width=\figurewidth,
        xlabel={rate (non-randomized components)},
        x grid style={white!69.01960784313725!black},
        xticklabel={\pgfmathparse{\tick}\pgfmathprintnumber{\pgfmathresult}\%},
        xmin=0, xmax=100,
        y grid style={white!69.01960784313725!black},
        ylabel={accuracy},
        ymin=0.0, ymax=1.0,
        ytick={0,0.2,0.4,0.6,0.8,1},
        yticklabels={0.0,0.2,0.4,0.6,0.8,1.0}
    ]

    % ord rde sfw, variant a
    \input{./img/mnist/mnist_rate_accuracy_sfw_order_var_a}
    % ord rde sfw, variant b
    \input{./img/mnist/mnist_rate_accuracy_sfw_order_var_b}
    % ord rde sfw, variant c
    \input{./img/mnist/mnist_rate_accuracy_sfw_order_var_c}
    % ord rde sfw, variant d
    \input{./img/mnist/mnist_rate_accuracy_sfw_order_var_d}
    % ord rde sfw, variant e
    \input{./img/mnist/mnist_rate_accuracy_sfw_order_var_e}
    % ord rde sfw, variant f
    \input{./img/mnist/mnist_rate_accuracy_sfw_order_var_f}
    
    \coordinate (spyon2) at (axis cs:60,0.92);
    \coordinate (spyat2) at (axis cs:99,0.025);
    \spy [width=3.5cm, height=1.3cm] on (spyon2) in node[fill=white,anchor=south east] at (spyat2);
   
\end{groupplot}
\end{tikzpicture}

\begin{center}
  \ref*{named-mnist-sfw}
\end{center}
\vspace{-1.5em}

%% file: img/stl10/stl10_examples3.tex
%\begin{tabular}{c@{\;\;\;}c@{\;\;\;}c@{\;\;\;}c@{\;\;\;}c}
%    Image & Sensitivity & LRP-$\alpha$-$\beta$ & SHAP & Guided Backprop \\
%    \includegraphics[width=1.20cm, valign=c]{./img/stl10/0001/stl10_testing_0001_input.png} &
%    \includegraphics[width=1.20cm, valign=c]{./img/stl10/0001/stl10_testing_0001_sensitivity.png} &
%    \includegraphics[width=1.20cm, valign=c]{./img/stl10/0001/stl10_testing_0001_lrp_ab.png} &
%    \includegraphics[width=1.20cm, valign=c]{./img/stl10/0001/stl10_testing_0001_shap.png} &
%    \includegraphics[width=1.20cm, valign=c]{./img/stl10/0001/stl10_testing_0001_guidedbackprop.png} \\
%    & SmoothGrad & DeepTaylor & LIME & L-RDE \\
%    &
%    \includegraphics[width=1.20cm, valign=c]{./img/stl10/0001/stl10_testing_0001_smoothgrad.png} &
%    \includegraphics[width=1.20cm, valign=c]{./img/stl10/0001/stl10_testing_0001_deeptaylor.png} &
%    \includegraphics[width=1.20cm, valign=c]{./img/stl10/0001/stl10_testing_0001_lime.png} &
%    \includegraphics[width=1.20cm, valign=c]{./img/stl10/0001/stl10_testing_0001_l_rde.png}
%\end{tabular}
%\\[2ex]
\begin{tabular}{lc@{\;\;}c@{\;\;}c@{\;\;}c@{\;\;}c}
    \toprule
    & \multicolumn{4}{c}{RC-RDE} & MR-RDE \\
    & $k=2000$ & $k=4000$ & $k=6000$ & $k=8000$ & \\
    \midrule
%    FW &
%    \includegraphics[width=1.20cm, valign=c]{./img/stl10/0001/stl10_testing_0001_fw_rc_rde_2000.png} &
%    \includegraphics[width=1.20cm, valign=c]{./img/stl10/0001/stl10_testing_0001_fw_rc_rde_4000.png} &
%    \includegraphics[width=1.20cm, valign=c]{./img/stl10/0001/stl10_testing_0001_fw_rc_rde_6000.png} &
%    \includegraphics[width=1.20cm, valign=c]{./img/stl10/0001/stl10_testing_0001_fw_rc_rde_8000.png} &
%    \includegraphics[width=1.20cm, valign=c]{./img/stl10/0001/stl10_testing_0001_fw_rc_rde_multi_cmap.png} \\

    AFW &
    \includegraphics[width=1.20cm, valign=c]{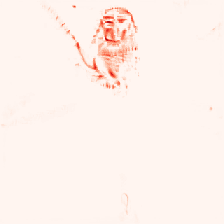} &
    \includegraphics[width=1.20cm, valign=c]{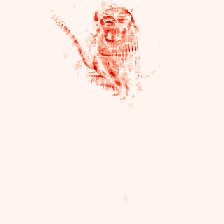} &
    \includegraphics[width=1.20cm, valign=c]{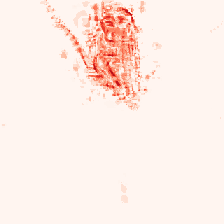} &
    \includegraphics[width=1.20cm, valign=c]{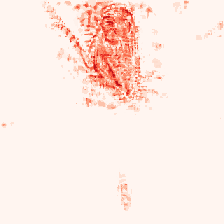} &
    \includegraphics[width=1.20cm, valign=c]{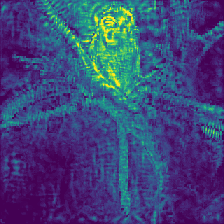} \\

    LCG &
    \includegraphics[width=1.20cm, valign=c]{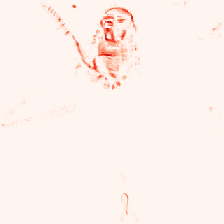} &
    \includegraphics[width=1.20cm, valign=c]{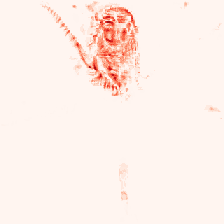} &
    \includegraphics[width=1.20cm, valign=c]{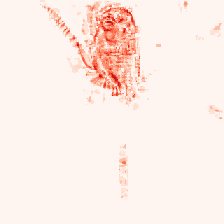} &
    \includegraphics[width=1.20cm, valign=c]{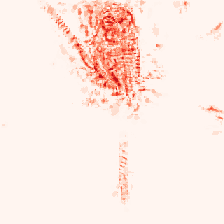} &
    \includegraphics[width=1.20cm, valign=c]{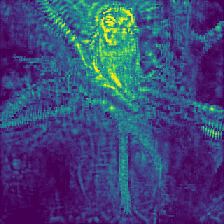} \\

    LAFW &
    \includegraphics[width=1.20cm, valign=c]{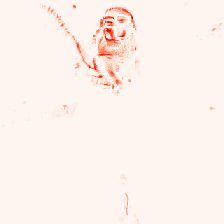} &
    \includegraphics[width=1.20cm, valign=c]{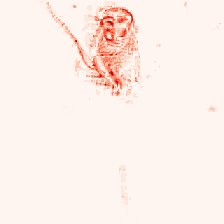} &
    \includegraphics[width=1.20cm, valign=c]{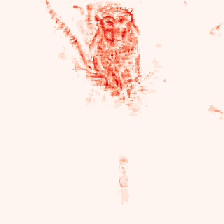} &
    \includegraphics[width=1.20cm, valign=c]{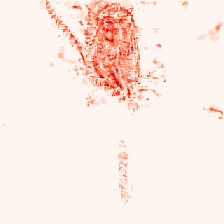} &
    \includegraphics[width=1.20cm, valign=c]{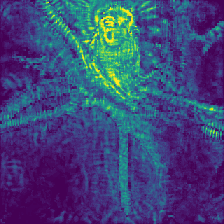} \\

    %SFW &
    %\includegraphics[width=1.20cm, valign=c]{./img/stl10/0001/stl10_testing_0001_sfw_rc_rde_2000.png} &
    %\includegraphics[width=1.20cm, valign=c]{./img/stl10/0001/stl10_testing_0001_sfw_rc_rde_4000.png} &
    %\includegraphics[width=1.20cm, valign=c]{./img/stl10/0001/stl10_testing_0001_sfw_rc_rde_6000.png} &
    %\includegraphics[width=1.20cm, valign=c]{./img/stl10/0001/stl10_testing_0001_sfw_rc_rde_8000.png} &
    %\includegraphics[width=1.20cm, valign=c]{./img/stl10/0001/stl10_testing_0001_sfw_rc_rde_multi_cmap.png} \\

%    PGD &
%    \includegraphics[width=1.20cm, valign=c]{./img/stl10/0001/stl10_testing_0001_pgd_rc_rde_2000.png} &
%    \includegraphics[width=1.20cm, valign=c]{./img/stl10/0001/stl10_testing_0001_pgd_rc_rde_4000.png} &
%    \includegraphics[width=1.20cm, valign=c]{./img/stl10/0001/stl10_testing_0001_pgd_rc_rde_6000.png} &
%    \includegraphics[width=1.20cm, valign=c]{./img/stl10/0001/stl10_testing_0001_pgd_rc_rde_8000.png} &
%    \includegraphics[width=1.20cm, valign=c]{./img/stl10/0001/stl10_testing_0001_pgd_rc_rde_multi_cmap.png} \\
    \bottomrule
\end{tabular}

%% file: img/stl10/stl10_examples4.tex
%\begin{tabular}{c@{\;\;\;}c@{\;\;\;}c@{\;\;\;}c@{\;\;\;}c}
%    Image & Sensitivity & LRP-$\alpha$-$\beta$ & SHAP & Guided Backprop \\
%    \includegraphics[width=1.20cm, valign=c]{./img/stl10/2485/stl10_testing_2485_input.png} &
%    \includegraphics[width=1.20cm, valign=c]{./img/stl10/2485/stl10_testing_2485_sensitivity.png} &
%    \includegraphics[width=1.20cm, valign=c]{./img/stl10/2485/stl10_testing_2485_lrp_ab.png} &
%    \includegraphics[width=1.20cm, valign=c]{./img/stl10/2485/stl10_testing_2485_shap.png} &
%    \includegraphics[width=1.20cm, valign=c]{./img/stl10/2485/stl10_testing_2485_guidedbackprop.png} \\
%    & SmoothGrad & DeepTaylor & LIME & L-RDE \\
%    &
%    \includegraphics[width=1.20cm, valign=c]{./img/stl10/2485/stl10_testing_2485_smoothgrad.png} &
%    \includegraphics[width=1.20cm, valign=c]{./img/stl10/2485/stl10_testing_2485_deeptaylor.png} &
%    \includegraphics[width=1.20cm, valign=c]{./img/stl10/2485/stl10_testing_2485_lime.png} &
%    \includegraphics[width=1.20cm, valign=c]{./img/stl10/2485/stl10_testing_2485_l_rde.png}
%\end{tabular}
%\\[2ex]
\begin{tabular}{lc@{\;\;}c@{\;\;}c@{\;\;}c@{\;\;}c}
    \toprule
    & \multicolumn{4}{c}{RC-RDE} & MR-RDE \\
    & $k=2000$ & $k=4000$ & $k=6000$ & $k=8000$ & \\
    \midrule
%    FW &
%    \includegraphics[width=1.20cm, valign=c]{./img/stl10/2485/stl10_testing_2485_fw_rc_rde_2000.png} &
%    \includegraphics[width=1.20cm, valign=c]{./img/stl10/2485/stl10_testing_2485_fw_rc_rde_4000.png} &
%    \includegraphics[width=1.20cm, valign=c]{./img/stl10/2485/stl10_testing_2485_fw_rc_rde_6000.png} &
%    \includegraphics[width=1.20cm, valign=c]{./img/stl10/2485/stl10_testing_2485_fw_rc_rde_8000.png} &
%    \includegraphics[width=1.20cm, valign=c]{./img/stl10/2485/stl10_testing_2485_fw_rc_rde_multi_cmap.png} \\

    AFW &
    \includegraphics[width=1.20cm, valign=c]{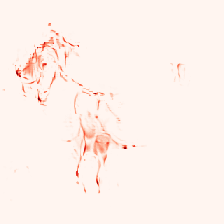} &
    \includegraphics[width=1.20cm, valign=c]{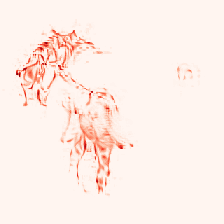} &
    \includegraphics[width=1.20cm, valign=c]{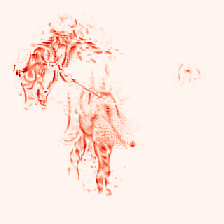} &
    \includegraphics[width=1.20cm, valign=c]{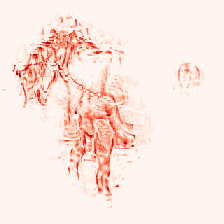} &
    \includegraphics[width=1.20cm, valign=c]{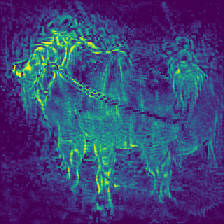} \\

    LCG &
    \includegraphics[width=1.20cm, valign=c]{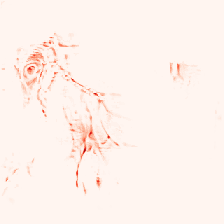} &
    \includegraphics[width=1.20cm, valign=c]{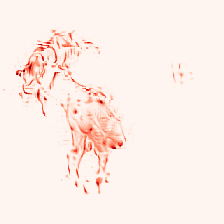} &
    \includegraphics[width=1.20cm, valign=c]{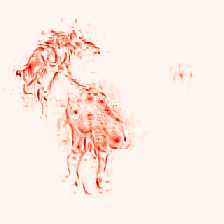} &
    \includegraphics[width=1.20cm, valign=c]{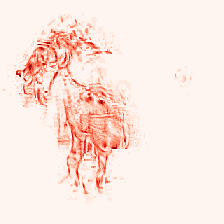} &
    \includegraphics[width=1.20cm, valign=c]{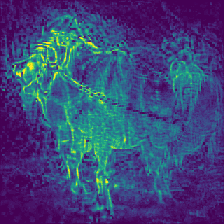} \\

    LAFW &
    \includegraphics[width=1.20cm, valign=c]{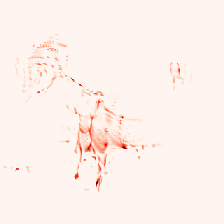} &
    \includegraphics[width=1.20cm, valign=c]{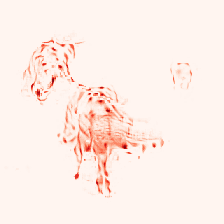} &
    \includegraphics[width=1.20cm, valign=c]{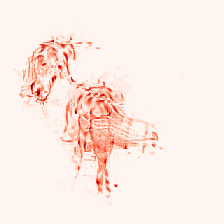} &
    \includegraphics[width=1.20cm, valign=c]{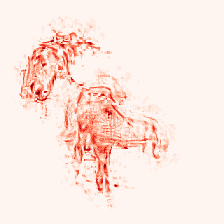} &
    \includegraphics[width=1.20cm, valign=c]{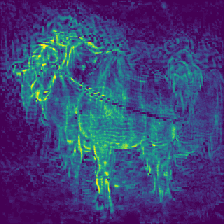} \\

    %SFW &
    %\includegraphics[width=1.20cm, valign=c]{./img/stl10/2485/stl10_testing_2485_sfw_rc_rde_2000.png} &
    %\includegraphics[width=1.20cm, valign=c]{./img/stl10/2485/stl10_testing_2485_sfw_rc_rde_4000.png} &
    %\includegraphics[width=1.20cm, valign=c]{./img/stl10/2485/stl10_testing_2485_sfw_rc_rde_6000.png} &
    %\includegraphics[width=1.20cm, valign=c]{./img/stl10/2485/stl10_testing_2485_sfw_rc_rde_8000.png} &
    %\includegraphics[width=1.20cm, valign=c]{./img/stl10/2485/stl10_testing_2485_sfw_rc_rde_multi_cmap.png} \\

%    PGD &
%    \includegraphics[width=1.20cm, valign=c]{./img/stl10/2485/stl10_testing_2485_pgd_rc_rde_2000.png} &
%    \includegraphics[width=1.20cm, valign=c]{./img/stl10/2485/stl10_testing_2485_pgd_rc_rde_4000.png} &
%    \includegraphics[width=1.20cm, valign=c]{./img/stl10/2485/stl10_testing_2485_pgd_rc_rde_6000.png} &
%    \includegraphics[width=1.20cm, valign=c]{./img/stl10/2485/stl10_testing_2485_pgd_rc_rde_8000.png} &
%    \includegraphics[width=1.20cm, valign=c]{./img/stl10/2485/stl10_testing_2485_pgd_rc_rde_multi_cmap.png} \\
    \bottomrule
\end{tabular}

%% file: img/stl10/stl10_examples1.tex
\begin{tabular}{c@{\;\;\;}c@{\;\;\;}c@{\;\;\;}c@{\;\;\;}c}
    Image & Sensitivity & LRP-$\alpha$-$\beta$ & SHAP & Guided Backprop \\
    \includegraphics[width=1.20cm, valign=c]{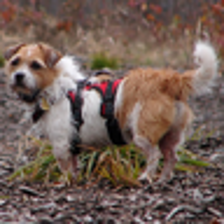} &
    \includegraphics[width=1.20cm, valign=c]{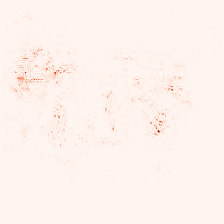} &
    \includegraphics[width=1.20cm, valign=c]{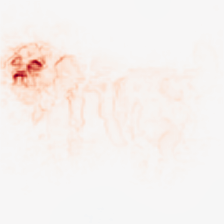} &
    \includegraphics[width=1.20cm, valign=c]{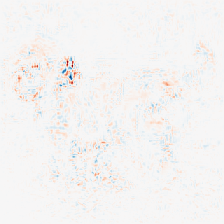} &
    \includegraphics[width=1.20cm, valign=c]{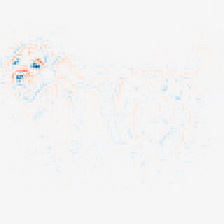} \\
    & SmoothGrad & DeepTaylor & LIME & L-RDE \\
    &
    \includegraphics[width=1.20cm, valign=c]{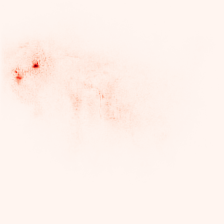} &
    \includegraphics[width=1.20cm, valign=c]{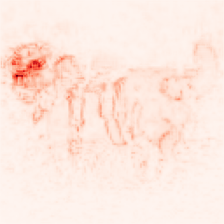} &
    \includegraphics[width=1.20cm, valign=c]{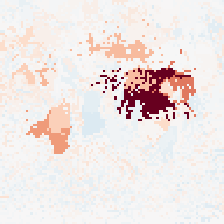} &
    \includegraphics[width=1.20cm, valign=c]{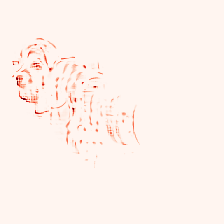}
\end{tabular}
\\[2ex]
\begin{tabular}{lc@{\;\;}c@{\;\;}c@{\;\;}c@{\;\;}c}
    \toprule
    & \multicolumn{4}{c}{RC-RDE} & MR-RDE \\
    & $k=2000$ & $k=4000$ & $k=6000$ & $k=8000$ & \\
    \midrule
    FW &
    \includegraphics[width=1.20cm, valign=c]{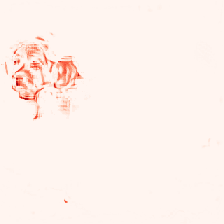} &
    \includegraphics[width=1.20cm, valign=c]{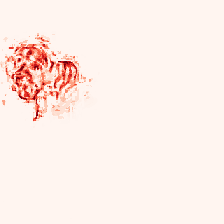} &
    \includegraphics[width=1.20cm, valign=c]{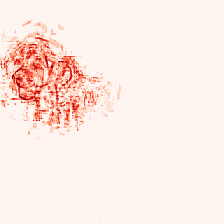} &
    \includegraphics[width=1.20cm, valign=c]{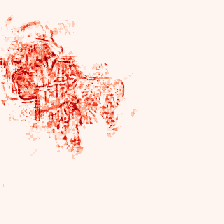} &
    \includegraphics[width=1.20cm, valign=c]{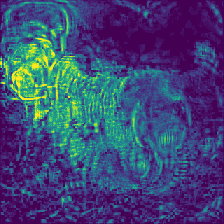} \\

    AFW &
    \includegraphics[width=1.20cm, valign=c]{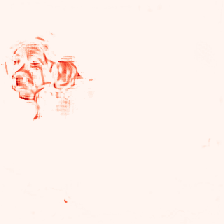} &
    \includegraphics[width=1.20cm, valign=c]{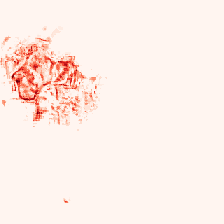} &
    \includegraphics[width=1.20cm, valign=c]{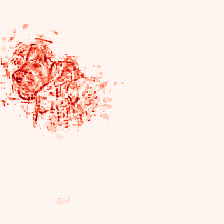} &
    \includegraphics[width=1.20cm, valign=c]{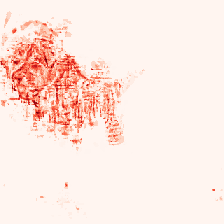} &
    \includegraphics[width=1.20cm, valign=c]{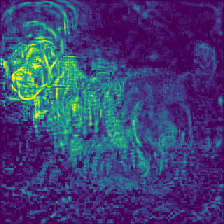} \\

    LCG &
    \includegraphics[width=1.20cm, valign=c]{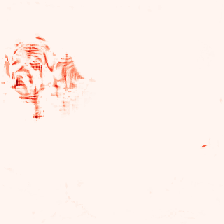} &
    \includegraphics[width=1.20cm, valign=c]{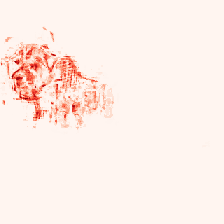} &
    \includegraphics[width=1.20cm, valign=c]{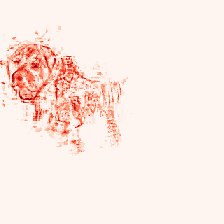} &
    \includegraphics[width=1.20cm, valign=c]{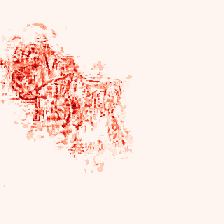} &
    \includegraphics[width=1.20cm, valign=c]{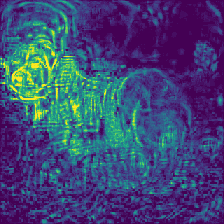} \\

    LAFW &
    \includegraphics[width=1.20cm, valign=c]{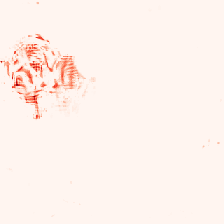} &
    \includegraphics[width=1.20cm, valign=c]{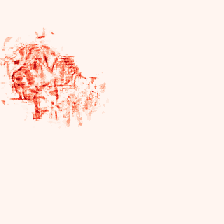} &
    \includegraphics[width=1.20cm, valign=c]{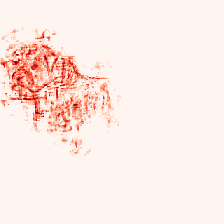} &
    \includegraphics[width=1.20cm, valign=c]{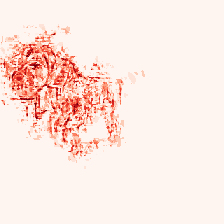} &
    \includegraphics[width=1.20cm, valign=c]{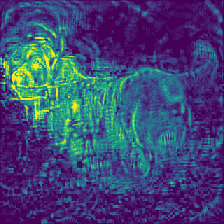} \\

    %SFW &
    %\includegraphics[width=1.20cm, valign=c]{./img/stl10/3892/stl10_testing_3892_sfw_rc_rde_2000.png} &
    %\includegraphics[width=1.20cm, valign=c]{./img/stl10/3892/stl10_testing_3892_sfw_rc_rde_4000.png} &
    %\includegraphics[width=1.20cm, valign=c]{./img/stl10/3892/stl10_testing_3892_sfw_rc_rde_6000.png} &
    %\includegraphics[width=1.20cm, valign=c]{./img/stl10/3892/stl10_testing_3892_sfw_rc_rde_8000.png} &
    %\includegraphics[width=1.20cm, valign=c]{./img/stl10/3892/stl10_testing_3892_sfw_rc_rde_multi_cmap.png} \\

    PGD &
    \includegraphics[width=1.20cm, valign=c]{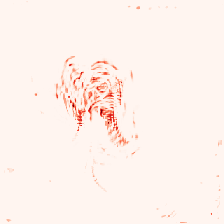} &
    \includegraphics[width=1.20cm, valign=c]{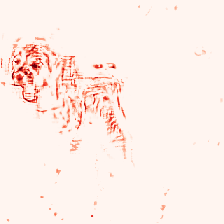} &
    \includegraphics[width=1.20cm, valign=c]{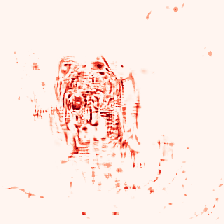} &
    \includegraphics[width=1.20cm, valign=c]{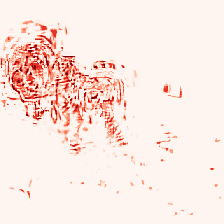} &
    \includegraphics[width=1.20cm, valign=c]{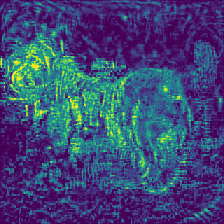} \\
    \bottomrule
\end{tabular}

%% file: img/stl10/stl10_examples2.tex
\begin{tabular}{c@{\;\;\;}c@{\;\;\;}c@{\;\;\;}c@{\;\;\;}c}
    Image & Sensitivity & LRP-$\alpha$-$\beta$ & SHAP & Guided Backprop \\
    \includegraphics[width=1.20cm, valign=c]{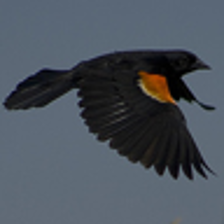} &
    \includegraphics[width=1.20cm, valign=c]{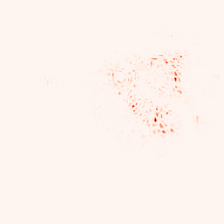} &
    \includegraphics[width=1.20cm, valign=c]{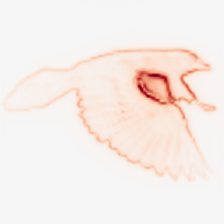} &
    \includegraphics[width=1.20cm, valign=c]{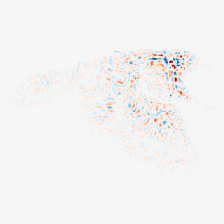} &
    \includegraphics[width=1.20cm, valign=c]{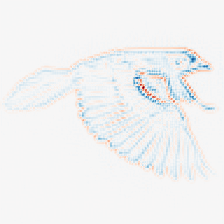} \\
    & SmoothGrad & DeepTaylor & LIME & L-RDE \\
    &
    \includegraphics[width=1.20cm, valign=c]{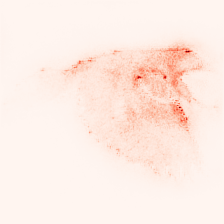} &
    \includegraphics[width=1.20cm, valign=c]{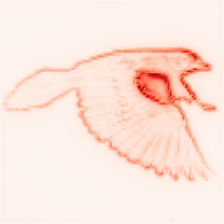} &
    \includegraphics[width=1.20cm, valign=c]{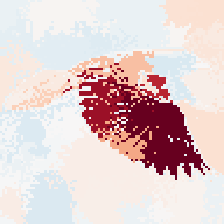} &
    \includegraphics[width=1.20cm, valign=c]{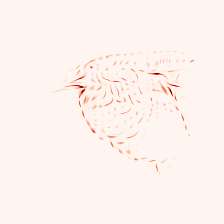}
\end{tabular}
\\[2ex]
\begin{tabular}{lc@{\;\;}c@{\;\;}c@{\;\;}c@{\;\;}c}
    \toprule
    & \multicolumn{4}{c}{RC-RDE} & MR-RDE \\
    & $k=2000$ & $k=4000$ & $k=6000$ & $k=8000$ & \\
    \midrule
    FW &
    \includegraphics[width=1.20cm, valign=c]{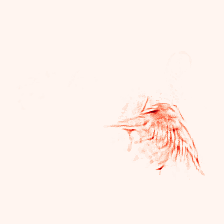} &
    \includegraphics[width=1.20cm, valign=c]{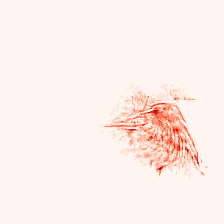} &
    \includegraphics[width=1.20cm, valign=c]{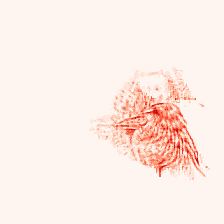} &
    \includegraphics[width=1.20cm, valign=c]{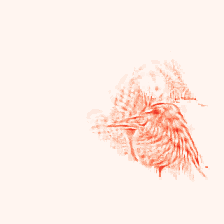} &
    \includegraphics[width=1.20cm, valign=c]{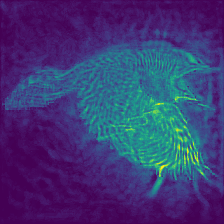} \\

    AFW &
    \includegraphics[width=1.20cm, valign=c]{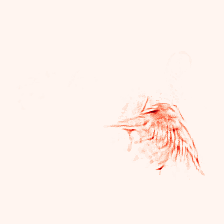} &
    \includegraphics[width=1.20cm, valign=c]{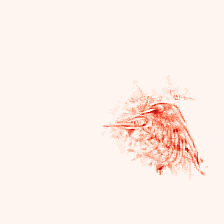} &
    \includegraphics[width=1.20cm, valign=c]{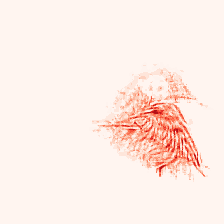} &
    \includegraphics[width=1.20cm, valign=c]{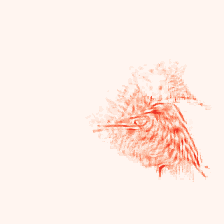} &
    \includegraphics[width=1.20cm, valign=c]{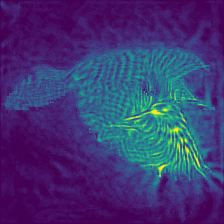} \\

    LCG &
    \includegraphics[width=1.20cm, valign=c]{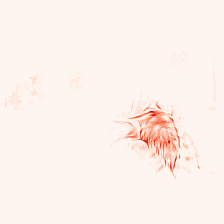} &
    \includegraphics[width=1.20cm, valign=c]{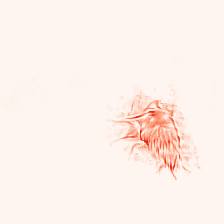} &
    \includegraphics[width=1.20cm, valign=c]{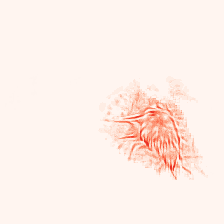} &
    \includegraphics[width=1.20cm, valign=c]{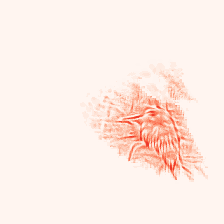} &
    \includegraphics[width=1.20cm, valign=c]{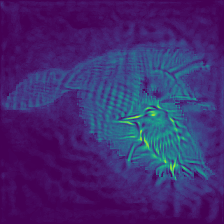} \\

    LAFW &
    \includegraphics[width=1.20cm, valign=c]{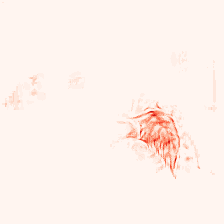} &
    \includegraphics[width=1.20cm, valign=c]{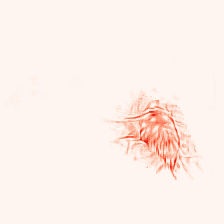} &
    \includegraphics[width=1.20cm, valign=c]{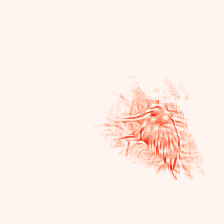} &
    \includegraphics[width=1.20cm, valign=c]{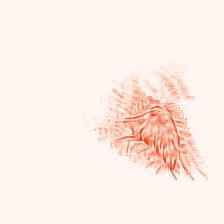} &
    \includegraphics[width=1.20cm, valign=c]{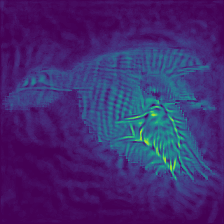} \\

    %SFW &
    %\includegraphics[width=1.20cm, valign=c]{./img/stl10/5253/stl10_testing_5253_sfw_rc_rde_2000.png} &
    %\includegraphics[width=1.20cm, valign=c]{./img/stl10/5253/stl10_testing_5253_sfw_rc_rde_4000.png} &
    %\includegraphics[width=1.20cm, valign=c]{./img/stl10/5253/stl10_testing_5253_sfw_rc_rde_6000.png} &
    %\includegraphics[width=1.20cm, valign=c]{./img/stl10/5253/stl10_testing_5253_sfw_rc_rde_8000.png} &
    %\includegraphics[width=1.20cm, valign=c]{./img/stl10/5253/stl10_testing_5253_sfw_rc_rde_multi_cmap.png} \\

    PGD &
    \includegraphics[width=1.20cm, valign=c]{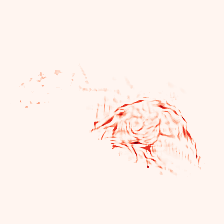} &
    \includegraphics[width=1.20cm, valign=c]{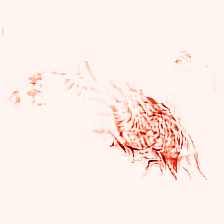} &
    \includegraphics[width=1.20cm, valign=c]{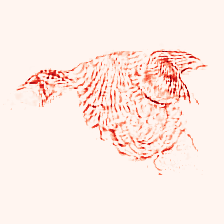} &
    \includegraphics[width=1.20cm, valign=c]{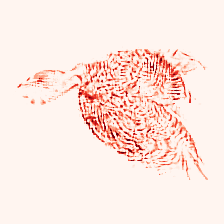} &
    \includegraphics[width=1.20cm, valign=c]{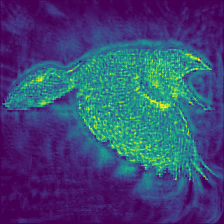} \\
    \bottomrule
\end{tabular}

%% file: img/stl10/stl10_examples5.tex
\begin{tabular}{c@{\;\;\;}c@{\;\;\;}c@{\;\;\;}c@{\;\;\;}c}
    Image & Sensitivity & LRP-$\alpha$-$\beta$ & SHAP & Guided Backprop \\
    \includegraphics[width=1.20cm, valign=c]{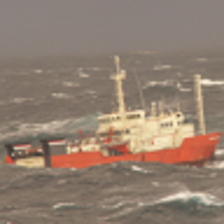} &
    \includegraphics[width=1.20cm, valign=c]{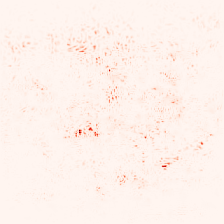} &
    \includegraphics[width=1.20cm, valign=c]{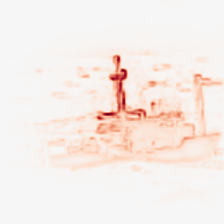} &
    \includegraphics[width=1.20cm, valign=c]{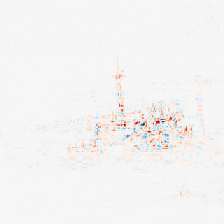} &
    \includegraphics[width=1.20cm, valign=c]{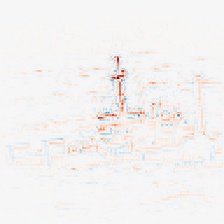} \\
    & SmoothGrad & DeepTaylor & LIME & L-RDE \\
    &
    \includegraphics[width=1.20cm, valign=c]{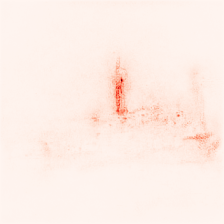} &
    \includegraphics[width=1.20cm, valign=c]{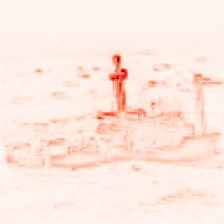} &
    \includegraphics[width=1.20cm, valign=c]{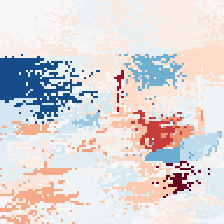} &
    \includegraphics[width=1.20cm, valign=c]{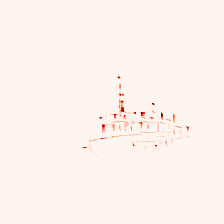}
\end{tabular}
\\[2ex]
\begin{tabular}{lc@{\;\;}c@{\;\;}c@{\;\;}c@{\;\;}c}
    \toprule
    & \multicolumn{4}{c}{RC-RDE} & MR-RDE \\
    & $k=2000$ & $k=4000$ & $k=6000$ & $k=8000$ & \\
    \midrule
    FW &
    \includegraphics[width=1.20cm, valign=c]{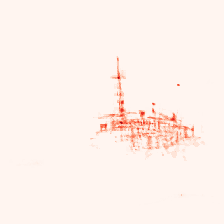} &
    \includegraphics[width=1.20cm, valign=c]{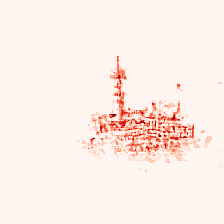} &
    \includegraphics[width=1.20cm, valign=c]{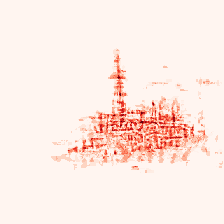} &
    \includegraphics[width=1.20cm, valign=c]{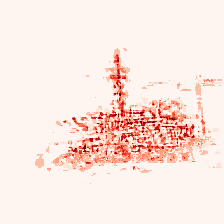} &
    \includegraphics[width=1.20cm, valign=c]{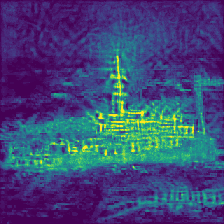} \\

    AFW &
    \includegraphics[width=1.20cm, valign=c]{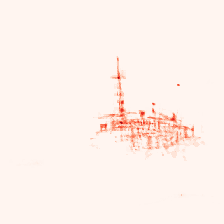} &
    \includegraphics[width=1.20cm, valign=c]{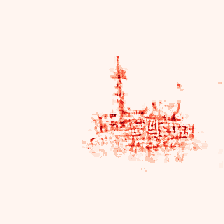} &
    \includegraphics[width=1.20cm, valign=c]{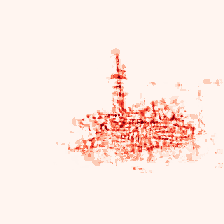} &
    \includegraphics[width=1.20cm, valign=c]{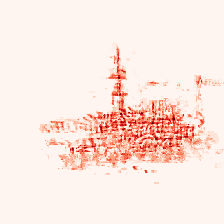} &
    \includegraphics[width=1.20cm, valign=c]{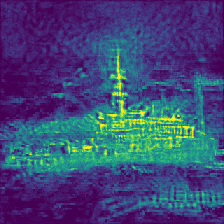} \\

    LCG &
    \includegraphics[width=1.20cm, valign=c]{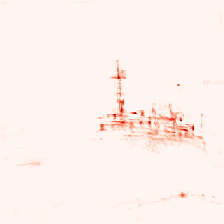} &
    \includegraphics[width=1.20cm, valign=c]{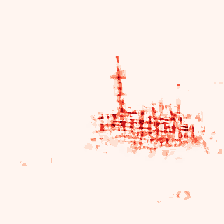} &
    \includegraphics[width=1.20cm, valign=c]{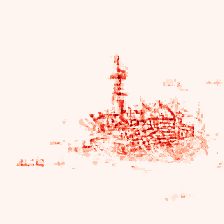} &
    \includegraphics[width=1.20cm, valign=c]{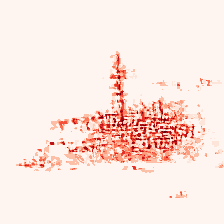} &
    \includegraphics[width=1.20cm, valign=c]{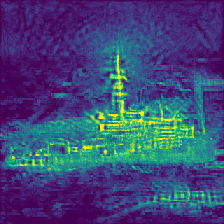} \\

    LAFW &
    \includegraphics[width=1.20cm, valign=c]{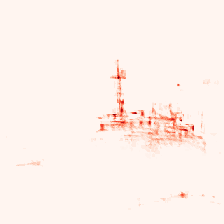} &
    \includegraphics[width=1.20cm, valign=c]{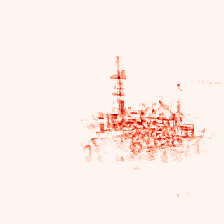} &
    \includegraphics[width=1.20cm, valign=c]{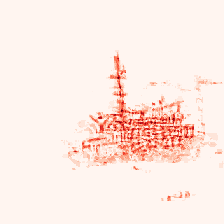} &
    \includegraphics[width=1.20cm, valign=c]{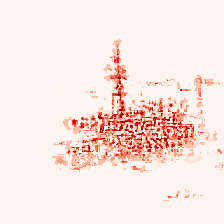} &
    \includegraphics[width=1.20cm, valign=c]{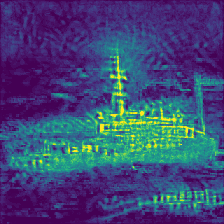} \\

    %SFW &
    %\includegraphics[width=1.20cm, valign=c]{./img/stl10/2394/stl10_testing_2394_sfw_rc_rde_2000.png} &
    %\includegraphics[width=1.20cm, valign=c]{./img/stl10/2394/stl10_testing_2394_sfw_rc_rde_4000.png} &
    %\includegraphics[width=1.20cm, valign=c]{./img/stl10/2394/stl10_testing_2394_sfw_rc_rde_6000.png} &
    %\includegraphics[width=1.20cm, valign=c]{./img/stl10/2394/stl10_testing_2394_sfw_rc_rde_8000.png} &
    %\includegraphics[width=1.20cm, valign=c]{./img/stl10/2394/stl10_testing_2394_sfw_rc_rde_multi_cmap.png} \\

    PGD &
    \includegraphics[width=1.20cm, valign=c]{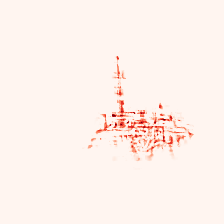} &
    \includegraphics[width=1.20cm, valign=c]{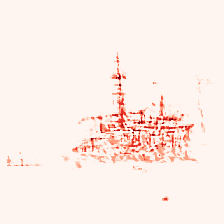} &
    \includegraphics[width=1.20cm, valign=c]{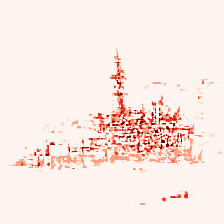} &
    \includegraphics[width=1.20cm, valign=c]{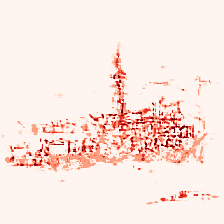} &
    \includegraphics[width=1.20cm, valign=c]{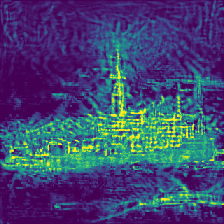} \\
    \bottomrule
\end{tabular}

%% file: img/stl10/stl10_examples7.tex
\begin{tabular}{c@{\;\;\;}c@{\;\;\;}c@{\;\;\;}c@{\;\;\;}c}
    Image & Sensitivity & LRP-$\alpha$-$\beta$ & SHAP & Guided Backprop \\
    \includegraphics[width=1.20cm, valign=c]{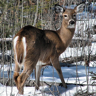} &
    \includegraphics[width=1.20cm, valign=c]{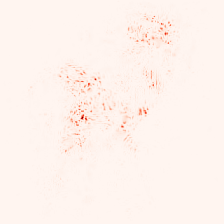} &
    \includegraphics[width=1.20cm, valign=c]{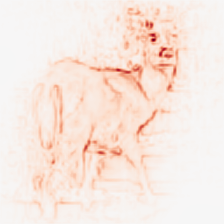} &
    \includegraphics[width=1.20cm, valign=c]{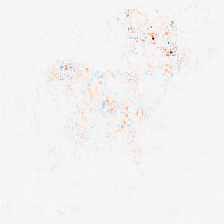} &
    \includegraphics[width=1.20cm, valign=c]{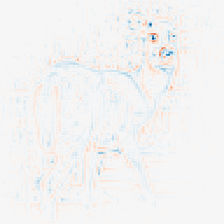} \\
    & SmoothGrad & DeepTaylor & LIME & L-RDE \\
    &
    \includegraphics[width=1.20cm, valign=c]{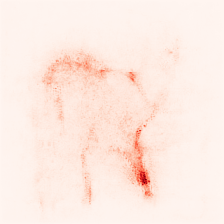} &
    \includegraphics[width=1.20cm, valign=c]{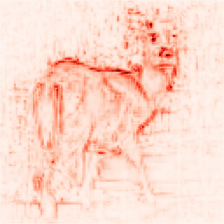} &
    \includegraphics[width=1.20cm, valign=c]{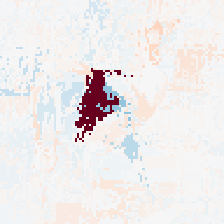} &
    \includegraphics[width=1.20cm, valign=c]{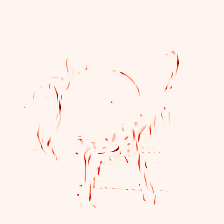}
\end{tabular}
\\[2ex]
\begin{tabular}{lc@{\;\;}c@{\;\;}c@{\;\;}c@{\;\;}c}
    \toprule
    & \multicolumn{4}{c}{RC-RDE} & MR-RDE \\
    & $k=2000$ & $k=4000$ & $k=6000$ & $k=8000$ & \\
    \midrule
    FW &
    \includegraphics[width=1.20cm, valign=c]{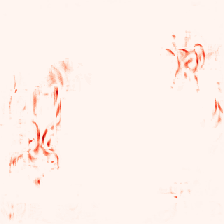} &
    \includegraphics[width=1.20cm, valign=c]{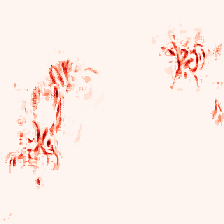} &
    \includegraphics[width=1.20cm, valign=c]{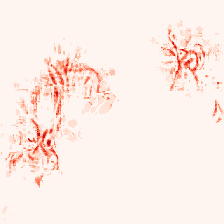} &
    \includegraphics[width=1.20cm, valign=c]{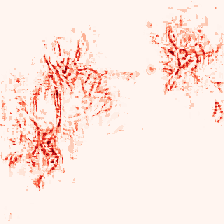} &
    \includegraphics[width=1.20cm, valign=c]{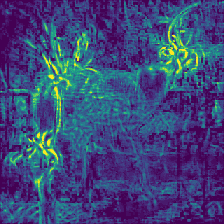} \\

    AFW &
    \includegraphics[width=1.20cm, valign=c]{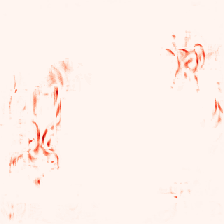} &
    \includegraphics[width=1.20cm, valign=c]{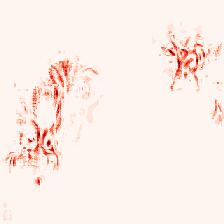} &
    \includegraphics[width=1.20cm, valign=c]{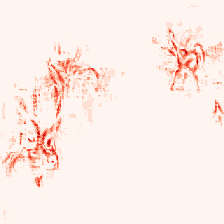} &
    \includegraphics[width=1.20cm, valign=c]{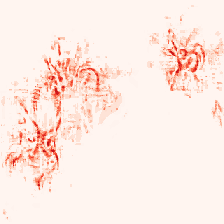} &
    \includegraphics[width=1.20cm, valign=c]{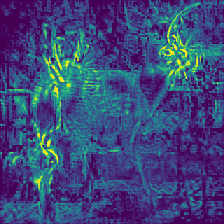} \\

    LCG &
    \includegraphics[width=1.20cm, valign=c]{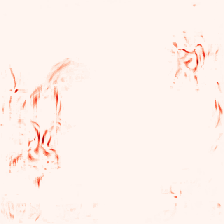} &
    \includegraphics[width=1.20cm, valign=c]{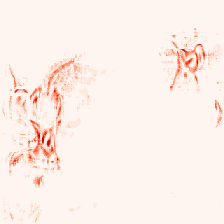} &
    \includegraphics[width=1.20cm, valign=c]{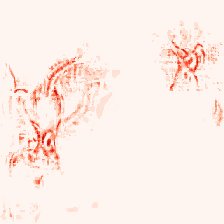} &
    \includegraphics[width=1.20cm, valign=c]{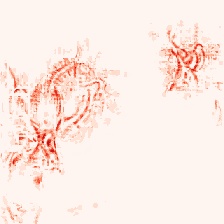} &
    \includegraphics[width=1.20cm, valign=c]{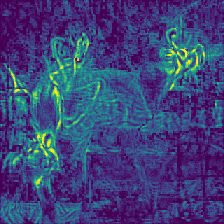} \\

    LAFW &
    \includegraphics[width=1.20cm, valign=c]{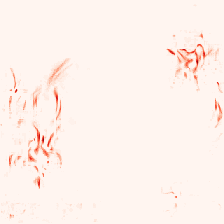} &
    \includegraphics[width=1.20cm, valign=c]{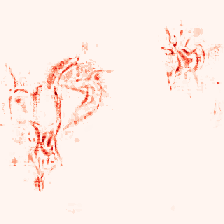} &
    \includegraphics[width=1.20cm, valign=c]{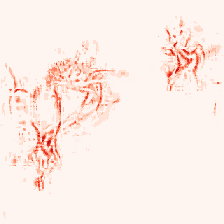} &
    \includegraphics[width=1.20cm, valign=c]{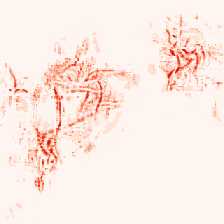} &
    \includegraphics[width=1.20cm, valign=c]{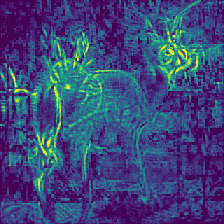} \\

    %SFW &
    %\includegraphics[width=1.20cm, valign=c]{./img/stl10/1011/stl10_testing_1011_sfw_rc_rde_2000.png} &
    %\includegraphics[width=1.20cm, valign=c]{./img/stl10/1011/stl10_testing_1011_sfw_rc_rde_4000.png} &
    %\includegraphics[width=1.20cm, valign=c]{./img/stl10/1011/stl10_testing_1011_sfw_rc_rde_6000.png} &
    %\includegraphics[width=1.20cm, valign=c]{./img/stl10/1011/stl10_testing_1011_sfw_rc_rde_8000.png} &
    %\includegraphics[width=1.20cm, valign=c]{./img/stl10/1011/stl10_testing_1011_sfw_rc_rde_multi_cmap.png} \\

    PGD &
    \includegraphics[width=1.20cm, valign=c]{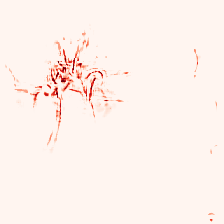} &
    \includegraphics[width=1.20cm, valign=c]{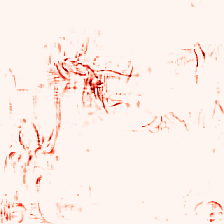} &
    \includegraphics[width=1.20cm, valign=c]{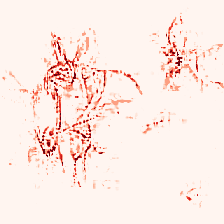} &
    \includegraphics[width=1.20cm, valign=c]{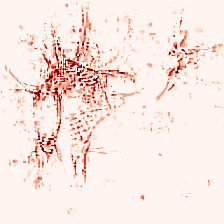} &
    \includegraphics[width=1.20cm, valign=c]{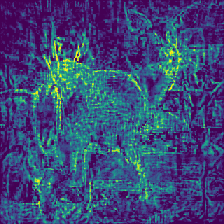} \\
    \bottomrule
\end{tabular}

%% file: img/stl10/stl10_examples6.tex
\begin{tabular}{c@{\;\;\;}c@{\;\;\;}c@{\;\;\;}c@{\;\;\;}c}
    Image & Sensitivity & LRP-$\alpha$-$\beta$ & SHAP & Guided Backprop \\
    \includegraphics[width=1.20cm, valign=c]{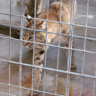} &
    \includegraphics[width=1.20cm, valign=c]{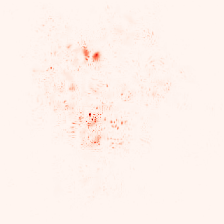} &
    \includegraphics[width=1.20cm, valign=c]{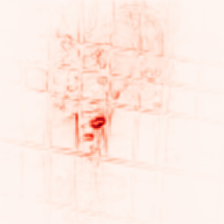} &
    \includegraphics[width=1.20cm, valign=c]{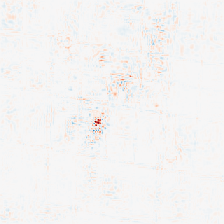} &
    \includegraphics[width=1.20cm, valign=c]{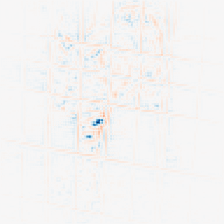} \\
    & SmoothGrad & DeepTaylor & LIME & L-RDE \\
    &
    \includegraphics[width=1.20cm, valign=c]{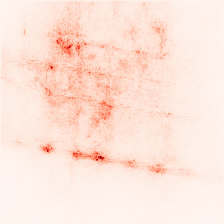} &
    \includegraphics[width=1.20cm, valign=c]{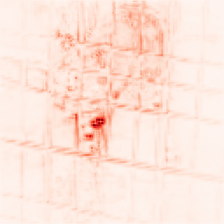} &
    \includegraphics[width=1.20cm, valign=c]{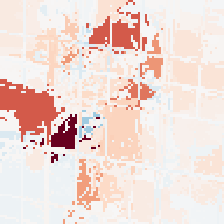} &
    \includegraphics[width=1.20cm, valign=c]{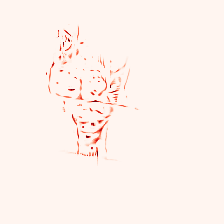}
\end{tabular}
\\[2ex]
\begin{tabular}{lc@{\;\;}c@{\;\;}c@{\;\;}c@{\;\;}c}
    \toprule
    & \multicolumn{4}{c}{RC-RDE} & MR-RDE \\
    & $k=2000$ & $k=4000$ & $k=6000$ & $k=8000$ & \\
    \midrule
    FW &
    \includegraphics[width=1.20cm, valign=c]{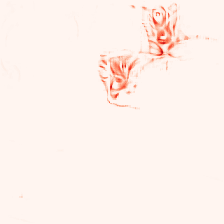} &
    \includegraphics[width=1.20cm, valign=c]{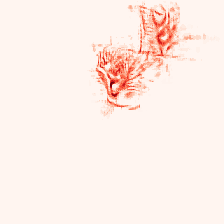} &
    \includegraphics[width=1.20cm, valign=c]{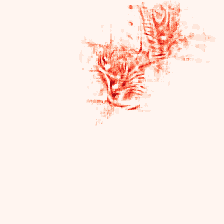} &
    \includegraphics[width=1.20cm, valign=c]{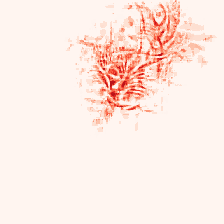} &
    \includegraphics[width=1.20cm, valign=c]{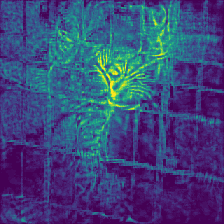} \\

    AFW &
    \includegraphics[width=1.20cm, valign=c]{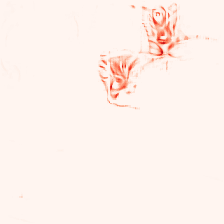} &
    \includegraphics[width=1.20cm, valign=c]{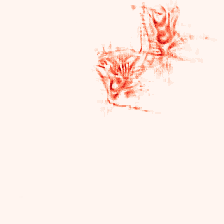} &
    \includegraphics[width=1.20cm, valign=c]{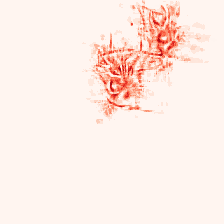} &
    \includegraphics[width=1.20cm, valign=c]{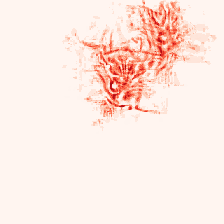} &
    \includegraphics[width=1.20cm, valign=c]{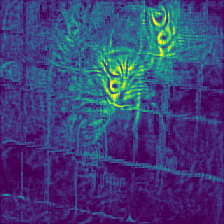} \\

    LCG &
    \includegraphics[width=1.20cm, valign=c]{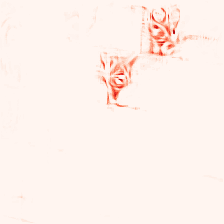} &
    \includegraphics[width=1.20cm, valign=c]{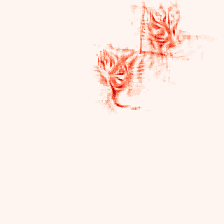} &
    \includegraphics[width=1.20cm, valign=c]{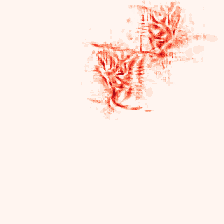} &
    \includegraphics[width=1.20cm, valign=c]{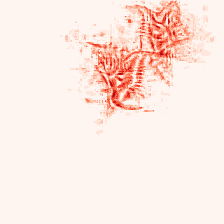} &
    \includegraphics[width=1.20cm, valign=c]{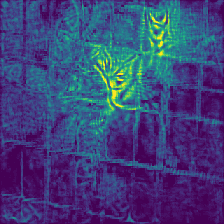} \\

    LAFW &
    \includegraphics[width=1.20cm, valign=c]{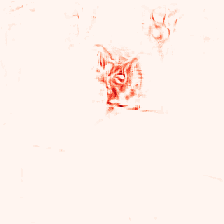} &
    \includegraphics[width=1.20cm, valign=c]{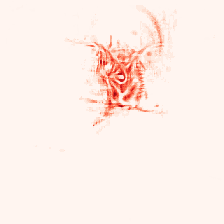} &
    \includegraphics[width=1.20cm, valign=c]{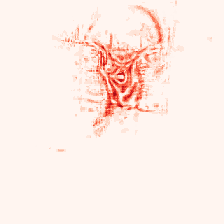} &
    \includegraphics[width=1.20cm, valign=c]{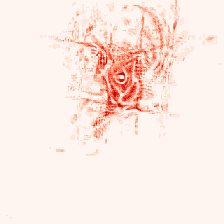} &
    \includegraphics[width=1.20cm, valign=c]{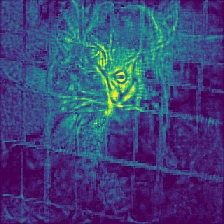} \\

    %SFW &
    %\includegraphics[width=1.20cm, valign=c]{./img/stl10/0101/stl10_testing_0101_sfw_rc_rde_2000.png} &
    %\includegraphics[width=1.20cm, valign=c]{./img/stl10/0101/stl10_testing_0101_sfw_rc_rde_4000.png} &
    %\includegraphics[width=1.20cm, valign=c]{./img/stl10/0101/stl10_testing_0101_sfw_rc_rde_6000.png} &
    %\includegraphics[width=1.20cm, valign=c]{./img/stl10/0101/stl10_testing_0101_sfw_rc_rde_8000.png} &
    %\includegraphics[width=1.20cm, valign=c]{./img/stl10/0101/stl10_testing_0101_sfw_rc_rde_multi_cmap.png} \\

    PGD &
    \includegraphics[width=1.20cm, valign=c]{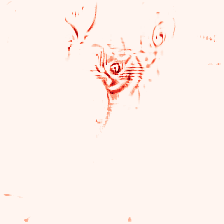} &
    \includegraphics[width=1.20cm, valign=c]{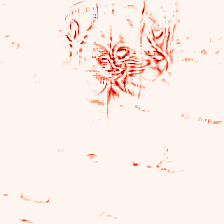} &
    \includegraphics[width=1.20cm, valign=c]{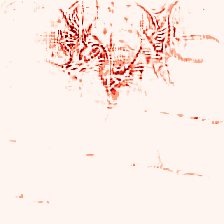} &
    \includegraphics[width=1.20cm, valign=c]{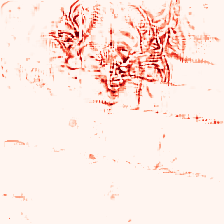} &
    \includegraphics[width=1.20cm, valign=c]{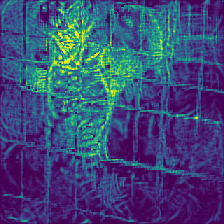} \\
    \bottomrule
\end{tabular}

%% file: img/stl10/stl10_examples8.tex
\begin{tabular}{c@{\;\;\;}c@{\;\;\;}c@{\;\;\;}c@{\;\;\;}c}
    Image & Sensitivity & LRP-$\alpha$-$\beta$ & SHAP & Guided Backprop \\
    \includegraphics[width=1.20cm, valign=c]{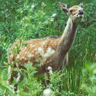} &
    \includegraphics[width=1.20cm, valign=c]{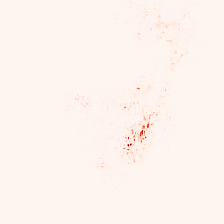} &
    \includegraphics[width=1.20cm, valign=c]{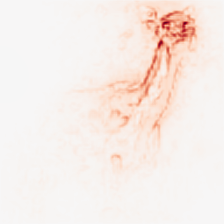} &
    \includegraphics[width=1.20cm, valign=c]{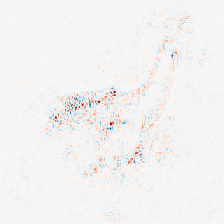} &
    \includegraphics[width=1.20cm, valign=c]{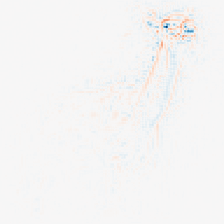} \\
    & SmoothGrad & DeepTaylor & LIME & L-RDE \\
    &
    \includegraphics[width=1.20cm, valign=c]{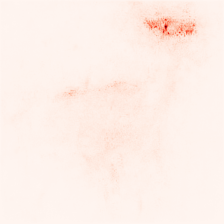} &
    \includegraphics[width=1.20cm, valign=c]{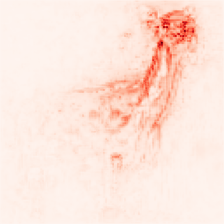} &
    \includegraphics[width=1.20cm, valign=c]{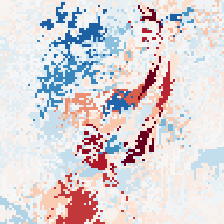} &
    \includegraphics[width=1.20cm, valign=c]{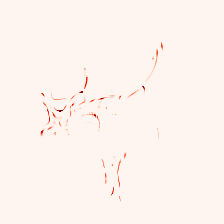}
\end{tabular}
\\[2ex]
\begin{tabular}{lc@{\;\;}c@{\;\;}c@{\;\;}c@{\;\;}c}
    \toprule
    & \multicolumn{4}{c}{RC-RDE} & MR-RDE \\
    & $k=2000$ & $k=4000$ & $k=6000$ & $k=8000$ & \\
    \midrule
    FW &
    \includegraphics[width=1.20cm, valign=c]{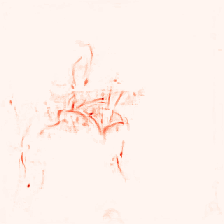} &
    \includegraphics[width=1.20cm, valign=c]{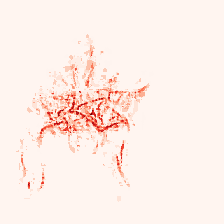} &
    \includegraphics[width=1.20cm, valign=c]{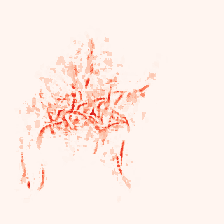} &
    \includegraphics[width=1.20cm, valign=c]{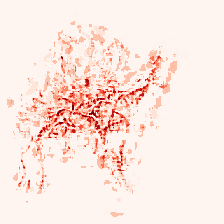} &
    \includegraphics[width=1.20cm, valign=c]{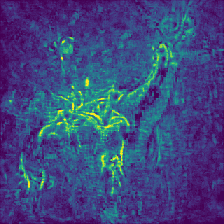} \\

    AFW &
    \includegraphics[width=1.20cm, valign=c]{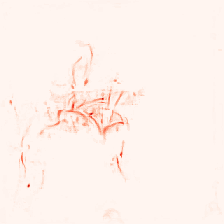} &
    \includegraphics[width=1.20cm, valign=c]{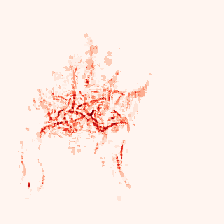} &
    \includegraphics[width=1.20cm, valign=c]{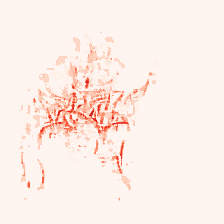} &
    \includegraphics[width=1.20cm, valign=c]{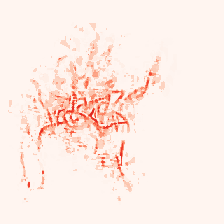} &
    \includegraphics[width=1.20cm, valign=c]{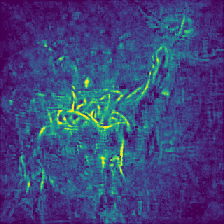} \\

    LCG &
    \includegraphics[width=1.20cm, valign=c]{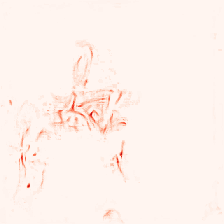} &
    \includegraphics[width=1.20cm, valign=c]{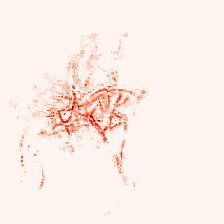} &
    \includegraphics[width=1.20cm, valign=c]{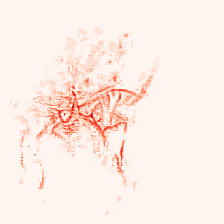} &
    \includegraphics[width=1.20cm, valign=c]{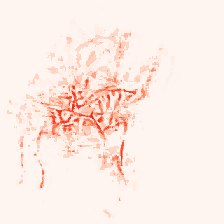} &
    \includegraphics[width=1.20cm, valign=c]{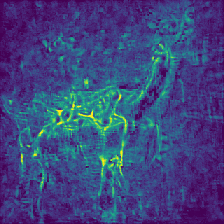} \\

    LAFW &
    \includegraphics[width=1.20cm, valign=c]{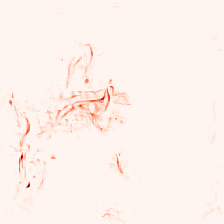} &
    \includegraphics[width=1.20cm, valign=c]{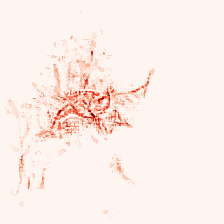} &
    \includegraphics[width=1.20cm, valign=c]{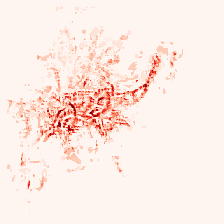} &
    \includegraphics[width=1.20cm, valign=c]{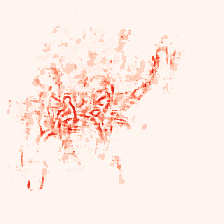} &
    \includegraphics[width=1.20cm, valign=c]{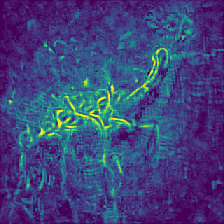} \\

    %SFW &
    %\includegraphics[width=1.20cm, valign=c]{./img/stl10/3891/stl10_testing_3891_sfw_rc_rde_2000.png} &
    %\includegraphics[width=1.20cm, valign=c]{./img/stl10/3891/stl10_testing_3891_sfw_rc_rde_4000.png} &
    %\includegraphics[width=1.20cm, valign=c]{./img/stl10/3891/stl10_testing_3891_sfw_rc_rde_6000.png} &
    %\includegraphics[width=1.20cm, valign=c]{./img/stl10/3891/stl10_testing_3891_sfw_rc_rde_8000.png} &
    %\includegraphics[width=1.20cm, valign=c]{./img/stl10/3891/stl10_testing_3891_sfw_rc_rde_multi_cmap.png} \\

    PGD &
    \includegraphics[width=1.20cm, valign=c]{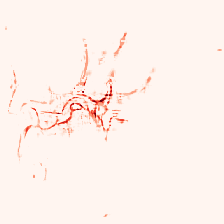} &
    \includegraphics[width=1.20cm, valign=c]{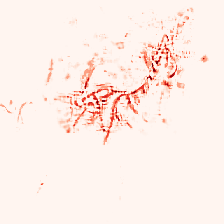} &
    \includegraphics[width=1.20cm, valign=c]{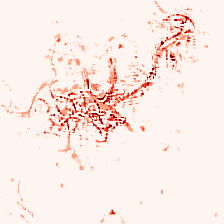} &
    \includegraphics[width=1.20cm, valign=c]{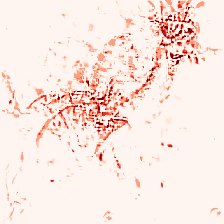} &
    \includegraphics[width=1.20cm, valign=c]{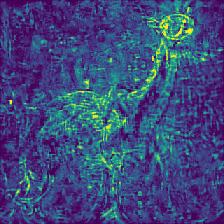} \\
    \bottomrule
\end{tabular}

%% file: img/stl10/stl10_examples9.tex
\begin{tabular}{c@{\;\;\;}c@{\;\;\;}c@{\;\;\;}c@{\;\;\;}c}
    Image & Sensitivity & LRP-$\alpha$-$\beta$ & SHAP & Guided Backprop \\
    \includegraphics[width=1.20cm, valign=c]{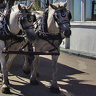} &
    \includegraphics[width=1.20cm, valign=c]{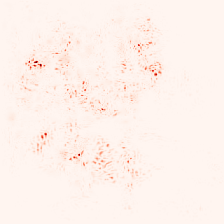} &
    \includegraphics[width=1.20cm, valign=c]{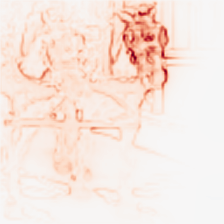} &
    \includegraphics[width=1.20cm, valign=c]{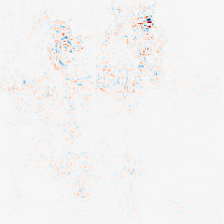} &
    \includegraphics[width=1.20cm, valign=c]{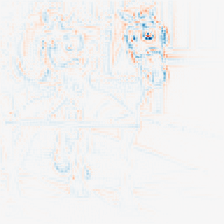} \\
    & SmoothGrad & DeepTaylor & LIME & L-RDE \\
    &
    \includegraphics[width=1.20cm, valign=c]{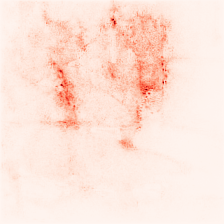} &
    \includegraphics[width=1.20cm, valign=c]{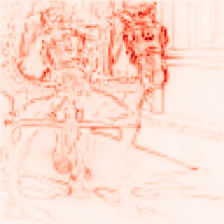} &
    \includegraphics[width=1.20cm, valign=c]{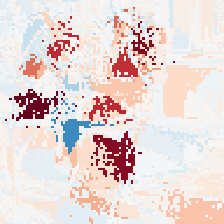} &
    \includegraphics[width=1.20cm, valign=c]{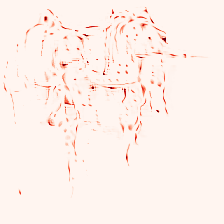}
\end{tabular}
\\[2ex]
\begin{tabular}{lc@{\;\;}c@{\;\;}c@{\;\;}c@{\;\;}c}
    \toprule
    & \multicolumn{4}{c}{RC-RDE} & MR-RDE \\
    & $k=2000$ & $k=4000$ & $k=6000$ & $k=8000$ & \\
    \midrule
    FW &
    \includegraphics[width=1.20cm, valign=c]{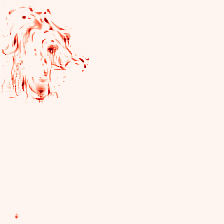} &
    \includegraphics[width=1.20cm, valign=c]{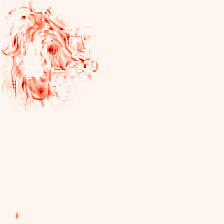} &
    \includegraphics[width=1.20cm, valign=c]{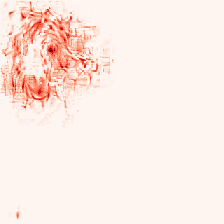} &
    \includegraphics[width=1.20cm, valign=c]{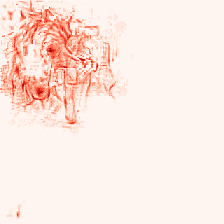} &
    \includegraphics[width=1.20cm, valign=c]{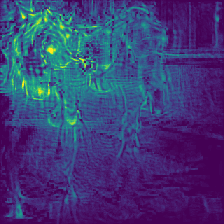} \\

    AFW &
    \includegraphics[width=1.20cm, valign=c]{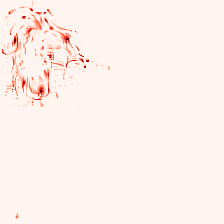} &
    \includegraphics[width=1.20cm, valign=c]{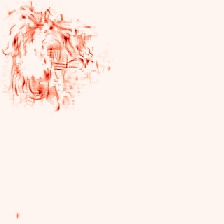} &
    \includegraphics[width=1.20cm, valign=c]{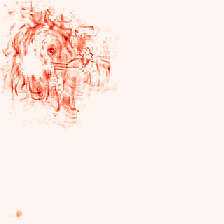} &
    \includegraphics[width=1.20cm, valign=c]{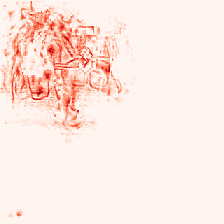} &
    \includegraphics[width=1.20cm, valign=c]{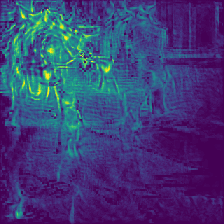} \\

    LCG &
    \includegraphics[width=1.20cm, valign=c]{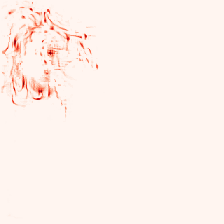} &
    \includegraphics[width=1.20cm, valign=c]{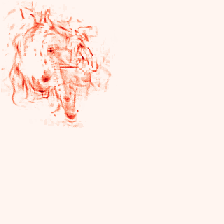} &
    \includegraphics[width=1.20cm, valign=c]{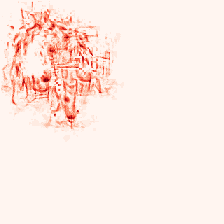} &
    \includegraphics[width=1.20cm, valign=c]{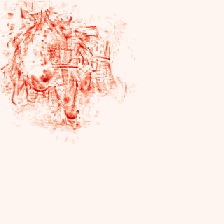} &
    \includegraphics[width=1.20cm, valign=c]{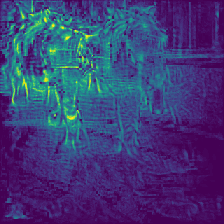} \\

    LAFW &
    \includegraphics[width=1.20cm, valign=c]{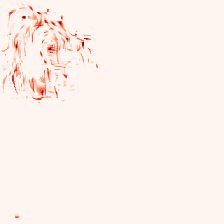} &
    \includegraphics[width=1.20cm, valign=c]{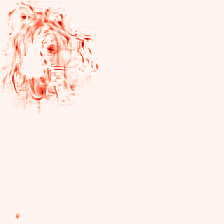} &
    \includegraphics[width=1.20cm, valign=c]{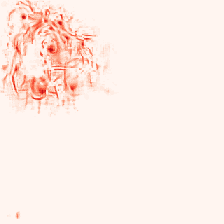} &
    \includegraphics[width=1.20cm, valign=c]{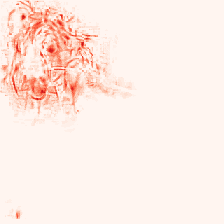} &
    \includegraphics[width=1.20cm, valign=c]{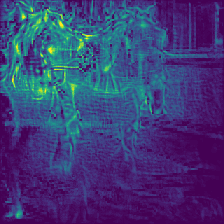} \\

    %SFW &
    %\includegraphics[width=1.20cm, valign=c]{./img/stl10/5254/stl10_testing_5254_sfw_rc_rde_2000.png} &
    %\includegraphics[width=1.20cm, valign=c]{./img/stl10/5254/stl10_testing_5254_sfw_rc_rde_4000.png} &
    %\includegraphics[width=1.20cm, valign=c]{./img/stl10/5254/stl10_testing_5254_sfw_rc_rde_6000.png} &
    %\includegraphics[width=1.20cm, valign=c]{./img/stl10/5254/stl10_testing_5254_sfw_rc_rde_8000.png} &
    %\includegraphics[width=1.20cm, valign=c]{./img/stl10/5254/stl10_testing_5254_sfw_rc_rde_multi_cmap.png} \\

    PGD &
    \includegraphics[width=1.20cm, valign=c]{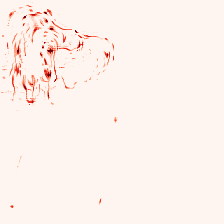} &
    \includegraphics[width=1.20cm, valign=c]{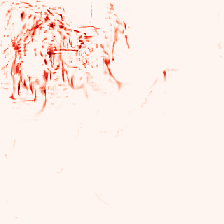} &
    \includegraphics[width=1.20cm, valign=c]{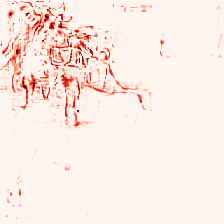} &
    \includegraphics[width=1.20cm, valign=c]{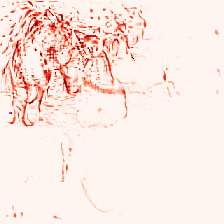} &
    \includegraphics[width=1.20cm, valign=c]{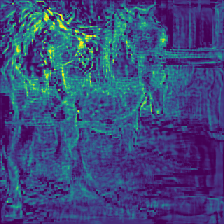} \\
    \bottomrule
\end{tabular}

%% file: img/stl10/stl10_examples10.tex
\begin{tabular}{c@{\;\;\;}c@{\;\;\;}c@{\;\;\;}c@{\;\;\;}c}
    Image & Sensitivity & LRP-$\alpha$-$\beta$ & SHAP & Guided Backprop \\
    \includegraphics[width=1.20cm, valign=c]{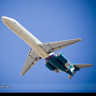} &
    \includegraphics[width=1.20cm, valign=c]{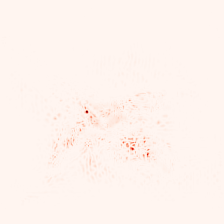} &
    \includegraphics[width=1.20cm, valign=c]{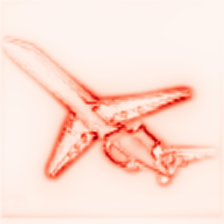} &
    \includegraphics[width=1.20cm, valign=c]{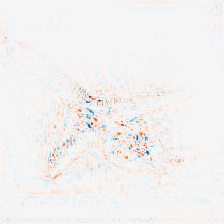} &
    \includegraphics[width=1.20cm, valign=c]{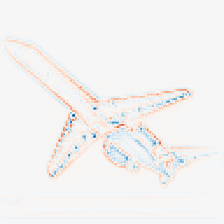} \\
    & SmoothGrad & DeepTaylor & LIME & L-RDE \\
    &
    \includegraphics[width=1.20cm, valign=c]{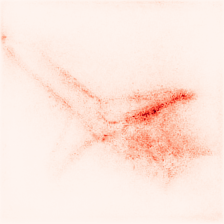} &
    \includegraphics[width=1.20cm, valign=c]{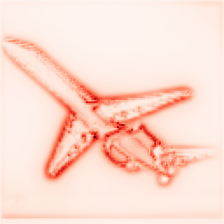} &
    \includegraphics[width=1.20cm, valign=c]{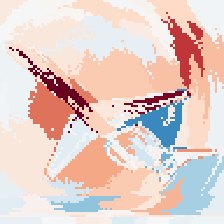} &
    \includegraphics[width=1.20cm, valign=c]{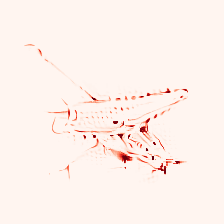}
\end{tabular}
\\[2ex]
\begin{tabular}{lc@{\;\;}c@{\;\;}c@{\;\;}c@{\;\;}c}
    \toprule
    & \multicolumn{4}{c}{RC-RDE} & MR-RDE \\
    & $k=2000$ & $k=4000$ & $k=6000$ & $k=8000$ & \\
    \midrule
    FW &
    \includegraphics[width=1.20cm, valign=c]{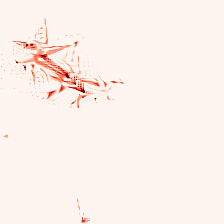} &
    \includegraphics[width=1.20cm, valign=c]{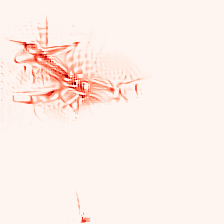} &
    \includegraphics[width=1.20cm, valign=c]{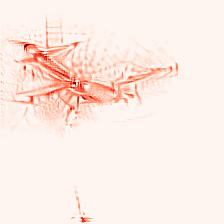} &
    \includegraphics[width=1.20cm, valign=c]{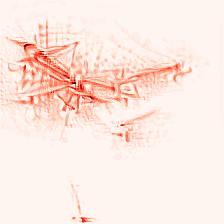} &
    \includegraphics[width=1.20cm, valign=c]{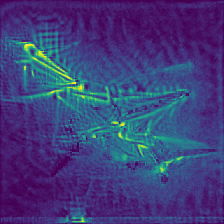} \\

    AFW &
    \includegraphics[width=1.20cm, valign=c]{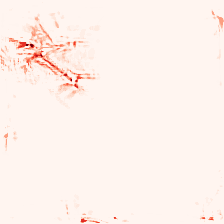} &
    \includegraphics[width=1.20cm, valign=c]{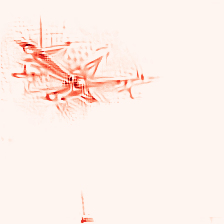} &
    \includegraphics[width=1.20cm, valign=c]{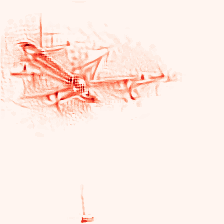} &
    \includegraphics[width=1.20cm, valign=c]{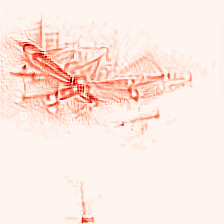} &
    \includegraphics[width=1.20cm, valign=c]{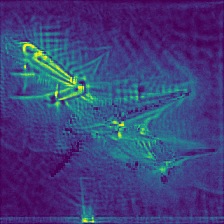} \\

    LCG &
    \includegraphics[width=1.20cm, valign=c]{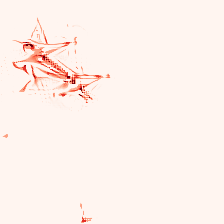} &
    \includegraphics[width=1.20cm, valign=c]{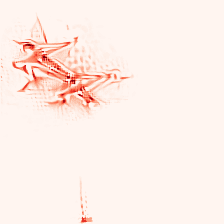} &
    \includegraphics[width=1.20cm, valign=c]{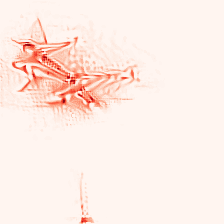} &
    \includegraphics[width=1.20cm, valign=c]{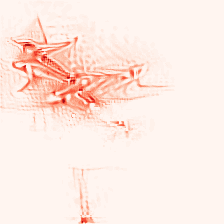} &
    \includegraphics[width=1.20cm, valign=c]{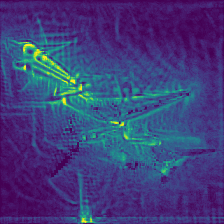} \\

    LAFW &
    \includegraphics[width=1.20cm, valign=c]{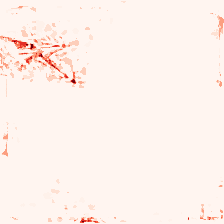} &
    \includegraphics[width=1.20cm, valign=c]{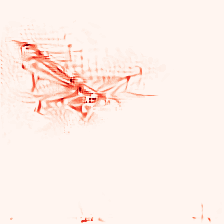} &
    \includegraphics[width=1.20cm, valign=c]{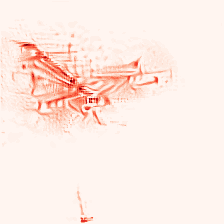} &
    \includegraphics[width=1.20cm, valign=c]{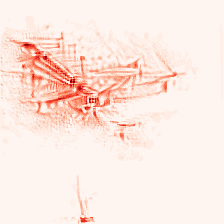} &
    \includegraphics[width=1.20cm, valign=c]{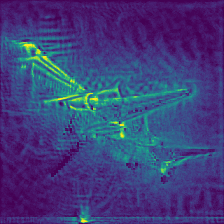} \\

    %SFW &
    %\includegraphics[width=1.20cm, valign=c]{./img/stl10/6819/stl10_testing_6819_sfw_rc_rde_2000.png} &
    %\includegraphics[width=1.20cm, valign=c]{./img/stl10/6819/stl10_testing_6819_sfw_rc_rde_4000.png} &
    %\includegraphics[width=1.20cm, valign=c]{./img/stl10/6819/stl10_testing_6819_sfw_rc_rde_6000.png} &
    %\includegraphics[width=1.20cm, valign=c]{./img/stl10/6819/stl10_testing_6819_sfw_rc_rde_8000.png} &
    %\includegraphics[width=1.20cm, valign=c]{./img/stl10/6819/stl10_testing_6819_sfw_rc_rde_multi_cmap.png} \\

    PGD &
    \includegraphics[width=1.20cm, valign=c]{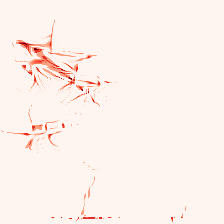} &
    \includegraphics[width=1.20cm, valign=c]{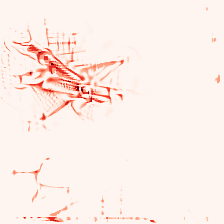} &
    \includegraphics[width=1.20cm, valign=c]{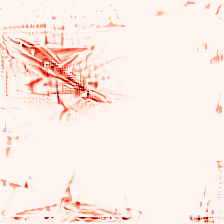} &
    \includegraphics[width=1.20cm, valign=c]{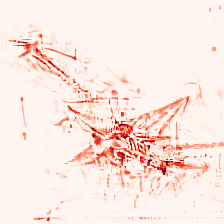} &
    \includegraphics[width=1.20cm, valign=c]{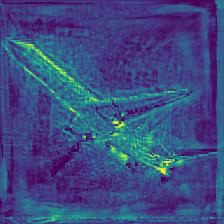} \\
    \bottomrule
\end{tabular}